\documentclass{article}

     \usepackage[preprint,nonatbib]{neurips_2021}

\usepackage[utf8]{inputenc} %
\usepackage[T1]{fontenc}    %
\usepackage{refcount}
\usepackage{xstring}
\usepackage{hyperref}       %
\usepackage{url}            %
\usepackage{booktabs}       %
\usepackage{amsfonts}       %
\usepackage{nicefrac}       %
\usepackage{microtype}      %
\usepackage[pdftex]{graphicx}
\usepackage[dvipsnames]{xcolor}
\usepackage{colortbl}
\usepackage{pgfplots}
\usepackage{caption}
\usepackage{subcaption}
\usepackage{xcolor}
\usepackage{amsmath}
\usepackage{mathtools}
\usepackage{MnSymbol}
\usepackage{wasysym}
\usepackage[ruled,vlined]{algorithm2e}
\usepackage[capitalise,noabbrev]{cleveref}
\usepackage{xcolor}
\usepackage{sidecap}
\usepackage{rotating}
\usepackage{pdflscape}
\usepackage{floatrow}
\newfloatcommand{ttabboxsideright}{table}[\capbeside\thisfloatsetup{capbesideposition={right,top}}][\FBwidth]
\DeclareUnicodeCharacter{2212}{−}
\usepgfplotslibrary{groupplots,dateplot}
\usetikzlibrary{patterns,shapes.arrows}
\pgfplotsset{compat=1.9}
\usepackage{tikzscale}

\makeatletter
\newcommand{\printfnsymbol}[1]{%
  \textsuperscript{\@fnsymbol{#1}}%
}
\makeatother

\definecolor{cerise}{rgb}{0.871, 0.192, 0.388}

\definecolor{carmine}{rgb}{0.59, 0.0, 0.09}

\definecolor{naturalcolor}{rgb}{0.835,0.369,0.0}
\definecolor{specializedcolor}{rgb}{0.800,0.475,0.655}
\definecolor{structuredcolor}{rgb}{0.941,0.894,0.259}

\newcommand{\densesym}{{\protect\scalebox{1.4}{$\bullet$}}}

\newcommand{\lastsym}{{\protect\scalebox{1.4}{$\blacktriangleright$}}}

\definecolor{s32color}{rgb}{0.122,0.467,0.706}
\definecolor{b32color}{rgb}{1.000,0.498,0.055}
\definecolor{b16color}{rgb}{0.173,0.627,0.173}
\definecolor{l32color}{rgb}{0.839,0.153,0.157}
\definecolor{l16color}{rgb}{0.580,0.404,0.741}
\definecolor{h14color}{rgb}{0.549,0.337,0.294}
\definecolor{g14color}{rgb}{0.890,0.467,0.761}

\newcommand{\name}{Vision MoE}
\newcommand{\abbv}{{V-MoE}}
\newcommand{\maxrouting}{Batch Prioritized Routing}

\title{Scaling Vision with Sparse Mixture of Experts}

\author{ {\bf Carlos Riquelme} \thanks{These authors contributed equally. Correspondence to \{ rikel, jpuigcerver, basilm \}@google.com} \\
Google Brain\\
\And
{\bf Joan Puigcerver} \printfnsymbol{1} \\
Google Brain\\
\And
{\bf Basil Mustafa} \printfnsymbol{1}  \\
Google Brain\\
\And
{\bf Maxim Neumann}   \\
Google Brain\\
\AND
{\bf Rodolphe Jenatton}   \\
Google Brain\\
\And
{\bf Andr\'e Susano Pinto}   \\
Google Brain\\
\And
{\bf Daniel Keysers}   \\
Google Brain\\ 
\And
{\bf Neil Houlsby}   \\
Google Brain\\
}

\begin{document}

\maketitle
\begin{abstract}
Sparsely-gated Mixture of Experts networks (MoEs) have demonstrated excellent scalability in Natural Language Processing.
In Computer Vision, however, almost all performant networks are ``dense'', that is, every input is processed by every parameter.
We present a \name{} (\abbv{}), a sparse version of the Vision Transformer, that is scalable and competitive with the largest dense networks.
When applied to image recognition, \abbv{} matches the performance of state-of-the-art networks, while requiring as little as \emph{half} of the compute at inference time.
Further, we propose an extension to the routing algorithm that can prioritize subsets of each input across the entire batch, leading to adaptive per-image compute.
This allows \abbv{} to trade-off performance and compute smoothly at test-time.
Finally, we demonstrate the potential of \abbv{} to scale vision models, and train a 15B parameter model that attains $90.35\%$ on ImageNet.

\end{abstract}

\section{Introduction}

Deep learning historically shows that increasing network capacity and dataset size generally improves performance.
In computer vision, large models pre-trained on large datasets often achieve the state of the art~\cite{sun2017revisiting,raffel2019exploring,kolesnikov2019big,dosovitskiy2020image,arnab2021vivit}.
This approach has had even more success in Natural Language Processing (NLP), where large pre-trained models are ubiquitous, and perform very well on many tasks~\cite{peters2018elmo,devlin2019bert}.
Text Transformers~\cite{vaswani2017attention} are the largest models to date, some with over 100B parameters~\cite{brown2020language}.
However, training and serving such models is expensive~\cite{strubell2019energy,patterson2021carbon}.
This is partially because these deep networks are typically ``dense''-- every example is processed using every parameter --thus, scale comes at high computational cost.
In contrast, conditional computation~\cite{bengio2013deep} aims to increase model capacity while keeping the training and inference cost roughly constant by applying only a subset of parameters to each example.
In NLP, sparse Mixture of Experts (MoEs) are gaining popularity~\cite{shazeer2017outrageously,lepikhin2020gshard,fedus2021switch}, enabling training and inference with fewer resources while unlocking trillion parameter models.

In this work, we explore conditional computation for vision at scale.
We introduce the \name{} (\abbv{}), a sparse variant of the recent Vision Transformer (ViT) architecture~\cite{dosovitskiy2020image} for image classification.
The \abbv{} replaces a subset of the dense feedforward layers in ViT with sparse MoE layers, where each image patch is ``routed'' to a subset of ``experts'' (MLPs).
Due to unique failure modes and non-differentiability, routing in deep sparse models is challenging.
We explore various design choices, and present an effective recipe for the pre-training and transfer of \abbv{}, notably outperforming their dense counterparts.
We further show that \abbv{} models are remarkably flexible. The performance vs.~inference-cost trade-off of \textit{already trained} models can be smoothly adjusted during inference by modulating the sparsity level with respect to the input and/or the model weights.

With \abbv{}, we can scale to model sizes of 15B parameters, the largest vision models to date.
We match the performance of state-of-the-art dense models, while requiring fewer time to train. 
Alternatively, \abbv{} can match the cost of ViT while achieving better performance.
To help control this tradeoff, we propose \maxrouting{}, a routing algorithm that repurposes model sparsity to skip the computation of some patches, reducing compute on uninformative image regions. 

We summarize our main contributions as follows:\\
\textbf{Vision models at scale.}
We present the Vision Mixture of Experts, a distributed sparsely-activated Transformer model for vision. We train models with up to 24 MoE layers, 32 experts per layer, and almost 15B parameters.
We show that these models can be stably trained, seamlessly used for transfer, and successfully fine-tuned with as few as 1\,000 datapoints. Moreover, our largest model achieves 90.35\% test accuracy on ImageNet when fine-tuned. \\
\textbf{Performance and inference.}
We show that \abbv{}s strongly outperform their dense counterparts on upstream, few-shot and full fine-tuning metrics in absolute terms.
Moreover, at inference time, the V-MoE models can be adjusted to either (i) match the performance of the largest dense model while using as little as half of the amount of compute, or actual runtime, or (ii) significantly outperform it at the same cost. \\
\textbf{\maxrouting{}.}
We propose a new priority-based routing algorithm that allows V-MoEs to discard the least useful patches.
Thus, we devote less compute to each image.
In particular, we show V-MoEs match the performance of the dense models while saving 20\% of the training FLOPs. \\
\textbf{Analysis.}
We provide some visualization of the routing decisions, revealing patterns and conclusions which helped motivate design decisions and may further improve understanding in the field.

\section{The Vision Mixture of Experts}
\label{sec:model}

\begin{figure}
\centering
\includegraphics[width=\linewidth]{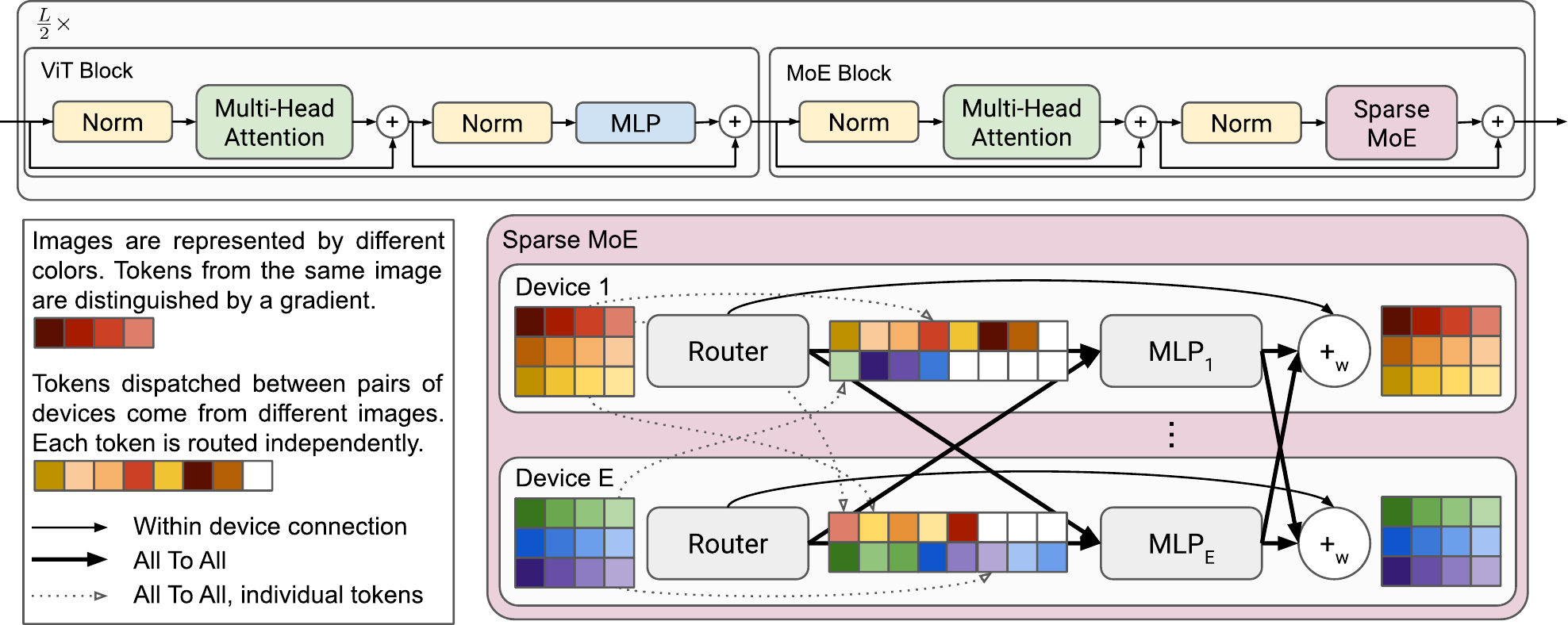}
\caption{\textbf{Overview of the architecture.} \abbv{} is composed of $L$ ViT blocks. In some, we replace the MLP with a sparsely activated \textit{mixture} of MLPs. Each MLP (the expert) is stored on a separate device, and processes a fixed number of tokens. %
The communication of these tokens between devices is shown in this example, which depicts the case when $k = 1$ expert is selected per token. 
Here each expert uses a capacity ratio $C = \frac{4}{3}$: the sparse MoE layer receives 12 tokens per device, but each expert has capacity for 16 ($\frac{16 \cdot 1}{12} = \frac{4}{3}$; see \cref{sec:expert_capacity}).
Non-expert components of \abbv{} such as routers, attention layers and normal MLP blocks are replicated identically across devices.}
\label{im:v_moe_architecture}
\end{figure}

We first describe MoEs and sparse MoEs. We then present how we apply this methodology to vision, 
before explaining our design choices for the routing algorithm and the implementation of \abbv{}s.

\subsection{Conditional Computation with MoEs}
Conditional computation aims at activating different subsets of a network for different inputs~\cite{bengio2013deep}.
A mixture-of-experts model is a specific instantiation whereby different model ``experts'' are responsible for different regions of the input space~\cite{jacobs1991adaptive}.

We follow the setting of~\cite{shazeer2017outrageously}, who present for deep learning a mixture of experts layer with $E$ experts as $\mathrm{MoE}(\mathbf{x})= \sum_{i=1}^E g(\mathbf{x})_i \ e_i(\mathbf{x})$ where $\mathbf{x}\in\mathbb{R}^D$ is the input to the layer, 
$e_i: \mathbb{R}^D\mapsto\mathbb{R}^D$ is the function computed by expert $i$, 
and $g: \mathbb{R}^D\mapsto\mathbb{R}^E$ is the ``routing'' function which prescribes the input-conditioned weight for the experts.
Both $e_i$ and $g$ are parameterized by neural networks.
As defined, this is still a dense network.  
However, if $g$ is sparse, i.e., restricted to assign only $k \ll E$ non-zero weights, then unused experts need not be computed.
This unlocks super-linear scaling of the number of model parameters with respect to inference and training compute.

\subsection{MoEs for Vision}
We explore the application of sparsity to vision in the context of the Vision Transformer (ViT)~\cite{dosovitskiy2020image}.
ViT has been shown to scale well in the transfer learning setting, attaining better accuracies than CNNs with less pre-training compute.
ViT processes images as a sequence of patches.
An input image is first divided into a grid of equal-sized patches.
These are linearly projected to the Transformer's~\cite{vaswani2017attention} hidden size.
After adding positional embeddings, the patch embeddings (tokens) are processed by a Transformer, which consists predominately of alternating self-attention and MLP layers.

The MLPs have two layers and a GeLU~\cite{hendrycks2016gaussian} non-linearity:
$\mathrm{MLP}(\mathbf{x})= \mathbf{W}_2 \ \sigma_\text{gelu}(\mathbf{W}_1 \mathbf{x})$.
For \name{}, we replace a subset of these with MoE layers, where each expert is an MLP; see \cref{im:v_moe_architecture}. The experts have the same architecture $e_i(\mathbf{x})=\mathrm{MLP}_{\theta_i}(\mathbf{x})$ but with different weights 
$\theta_i = (\mathbf{W}_1^i, \mathbf{W}_2^i)$. This follows a similar design pattern as the M4 machine translation model~\cite{lepikhin2020gshard}.

\subsection{Routing}
\label{sec:routing}
For each MoE layer in \abbv{}, we use the routing function $g(\mathbf{x}) = \text{TOP}_k \left ( \text{softmax} \left( \mathbf{W}\mathbf{x} + \mathbf{\epsilon} \right) \right )$, where $\text{TOP}_k$ is an operation that sets all elements of the vector to zero except the elements with the largest $k$ values, and $\epsilon$ is sampled independently $\epsilon \sim \mathcal{N}(0, \frac{1}{E^2})$ entry-wise.
In practice, we use $k=1$ or $k=2$.
In the context of the Vision Transformer, $\mathbf{x}$ is a representation of an image token at some layer of the network.
Therefore, \abbv{} routes patch representations, not entire images.

The difference between previous formulations~\cite{shazeer2017outrageously} is that we apply $\text{TOP}_k$ \textit{after} the softmax over experts weights, instead of \textit{before}.
This allows us to train with $k=1$ (otherwise gradients with respect to routings are zero almost everywhere) and also performs better for $k > 1$ (see \cref{sec:app_moe_model}).

Finally, we add a small amount of noise with standard deviation $\frac{1}{E}$ to the activations $\mathbf{W}\mathbf{x}$, which we find improves performance.
We empirically found this performed well but that the setup was robust to this parameter.
The noise typically altered routing decisions $\sim$15\% of the time in earlier layers, and $\sim$2--3\% in deeper layers.

\subsection{Expert's Buffer Capacity}
\label{sec:expert_capacity}
During training, sparse models may favor only a small set of experts~\cite{hansen1999combining,rosenbaum2019routing}.
This common failure mode can cause two problems.
First, statistical inefficiency: in the limit of collapse to a single expert, the model is no more powerful than a dense model.
Second, computational inefficiency: imbalanced assignment of items to experts may lead to a poor hardware utilization.

\looseness=-1
To combat imbalance and simplify our implementation, we fix the \emph{buffer capacity} of each expert 
(i.e. the number of tokens that each expert processes), 
and train our model with auxiliary losses that encourage load balancing. This is essentially the same approach as followed by
\cite{shazeer2017outrageously,lepikhin2020gshard,fedus2021switch}. In our case, we use slight variants of two of the auxiliary 
losses proposed in \cite{shazeer2017outrageously}, as described in \cref{sec:app_moe_model}. 

We define the buffer capacity of an expert ($B_e$) as a function of the number of images in the batch ($N$), the number of tokens 
per image ($P$),  the number of selected experts per token ($k$), the total number of experts ($E$), and the \emph{capacity ratio} ($C$):
$B_e = \text{round}\left(\frac{k N P C}{E}\right)$.

If the router assigns more than $B_e$ tokens to a given expert, only $B_e$ of them are processed.
The remaining tokens are not entirely `lost' as their information is preserved by residual connections (the 
top diagram of \cref{im:v_moe_architecture}).
Also, if $k > 1$, several experts try to process each token.
Tokens are never fully discarded.
If an expert is assigned fewer than $B_e$ tokens, the rest of its buffer is zero-padded. 

We use the \emph{capacity ratio} to adjust the capacity of the experts. With $C > 1$, a \emph{slack} capacity is added to account for a potential 
routing imbalance. This is typically useful for fine-tuning when the new data might come from a very different distribution than during upstream training. 
With $C < 1$, the router is forced to ignore some assignments. In \cref{sec:skip_patch} we propose a new algorithm that takes 
advantage of setting $C \ll 1$ to discard the least useful tokens and save compute during inference.

\section{Transfer Learning}
In this section, we first present training different variants of \abbv{} on a large dataset 
(\cref{sect:data}) in order to
be used for Transfer Learning afterwards. The ability to easily adapt our massive models to new tasks,
using a small amount of data from the new task, is extremely valuable: it allows to amortize the cost 
of pre-training across multiple tasks. 
We consider two different approaches to Transfer Learning:
linear few-shot learning on fixed representations and full fine-tuning of the model.

\subsection{Models}
We build V-MoE on different variants of ViT~\cite{dosovitskiy2020image}:
ViT-S(mall), ViT-B(ase), ViT-L(arge) and ViT-H(uge), the hyperparameters of which are described in \cref{sec:model_table}.
There are three additional major design decisions that affect the cost (and potentially the quality) of our model:

\textbf{Number of MoE layers.}
Following \cite{lepikhin2020gshard}, we place the MoEs on every other layer (we refer to these as \abbv{} \emph{Every-2}).
In addition, we experimented with using fewer MoE layers, by placing them on the last-$n$ \emph{even} blocks 
(thus we dub these \abbv{} \emph{Last-n}). In \cref{app_analysis_value_routers} we observe that, although using fewer MoE 
layers decreases the number of parameters of the model, it has typically little impact on quality and can speed-up the models significantly,
since less communication overhead is incurred.

\textbf{Number of selected experts} $k$:
The cost of our model does not depend on the total number of experts but the number of \emph{selected} ones per token. 
Concurrent works in NLP fix $k = 1$~\cite{fedus2021switch} or $k = 2$~\cite{shazeer2017outrageously,lepikhin2020gshard}.
In our case, we use by default $k = 2$ (see \cref{im:performance_increasing_k} in \cref{sec:experiment_details} for the exploration of different values of $k$), while we found the total number of experts $E = 32$ to be the sweet spot in our setting.

\textbf{Buffer capacity} $C$:
As mentioned in \cref{sec:expert_capacity}, we use a fixed buffer capacity. While this is typically regarded as a downside or engineering difficulty to implement these models, we can adjust the \emph{capacity ratio} to control different trade-offs. We can intentionally
set it to a low ratio to save compute, using Batch Prioritized Routing (see \cref{sec:skip_patch}). 
During upstream training, we set $C = 1.05$ by default to give a small amount of slack without increasing the cost noticeably.

Note that for a given trained model, the latter two---$k$ and $C$---can be adjusted without further training, whereas the positioning and quantity of expert layers is effectively fixed to match pre-training.

\subsection{Data}\label{sect:data}
We pre-train our models on JFT-300M~\cite{sun2017revisiting}, a semi-automatically noisy-labeled dataset. 
It has $\sim$~305M training and 50\,000 validation images,
organised in a hierarchy of 18\,291 classes (average 1.89 labels per image). We deduplicate it with respect to all our validation/test sets as in previous efforts~\cite{kolesnikov2019big}.\footnote{We also checked the effect of deduplication with respect to the ImageNet \textit{training} set, showing negligible (within noise) impact on few-shot results (only 1-shot worsened, see \cref{tab:dedup_models}).}

Our few-shot experiments on ImageNet (i.e.\ ILSVRC2012) use only 1, 5, or 10 shots per class to adapt the upstream model, evaluating the resulting model on the validation set.

We also fine-tuned the pre-trained models on the full training set (ca.\ 1M images).
We report performance in a similar regime for four other datasets in
\cref{sec:model_table}. Lastly, we explore the ability to fine-tune our large models in the low-data regime by evaluating
them on the Visual Task Adaptation Benchmark (VTAB)~\cite{zhai2019largescale}, 
a diverse suite of 19 tasks with only 1\,000 data points per task.
As well as natural image classification, VTAB includes specialized tasks (e.g.\ medical or satellite imagery) and structured tasks 
(e.g.\ counting or assessing rotation/distance).

\subsection{Upstream results}
\begin{figure}
\centering
\begin{subfigure}{.47\textwidth}
  \centering
  \includegraphics[width=\linewidth]{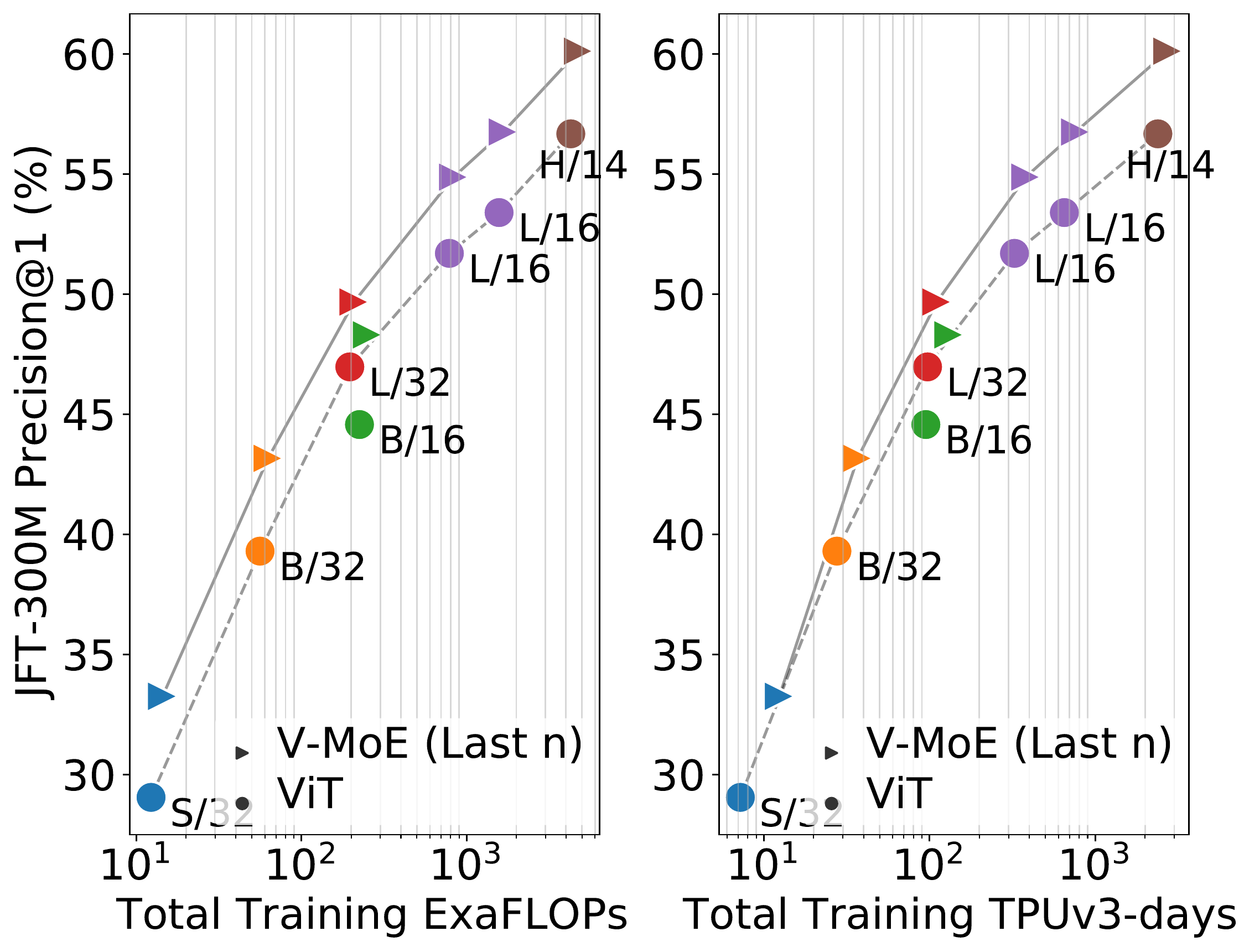}
  \caption{JFT-300M}
  \label{im:upstream_vs_exaflops_and_days}
\end{subfigure}%
~~~
\begin{subfigure}{.47\textwidth}
  \centering
  \includegraphics[width=\linewidth]{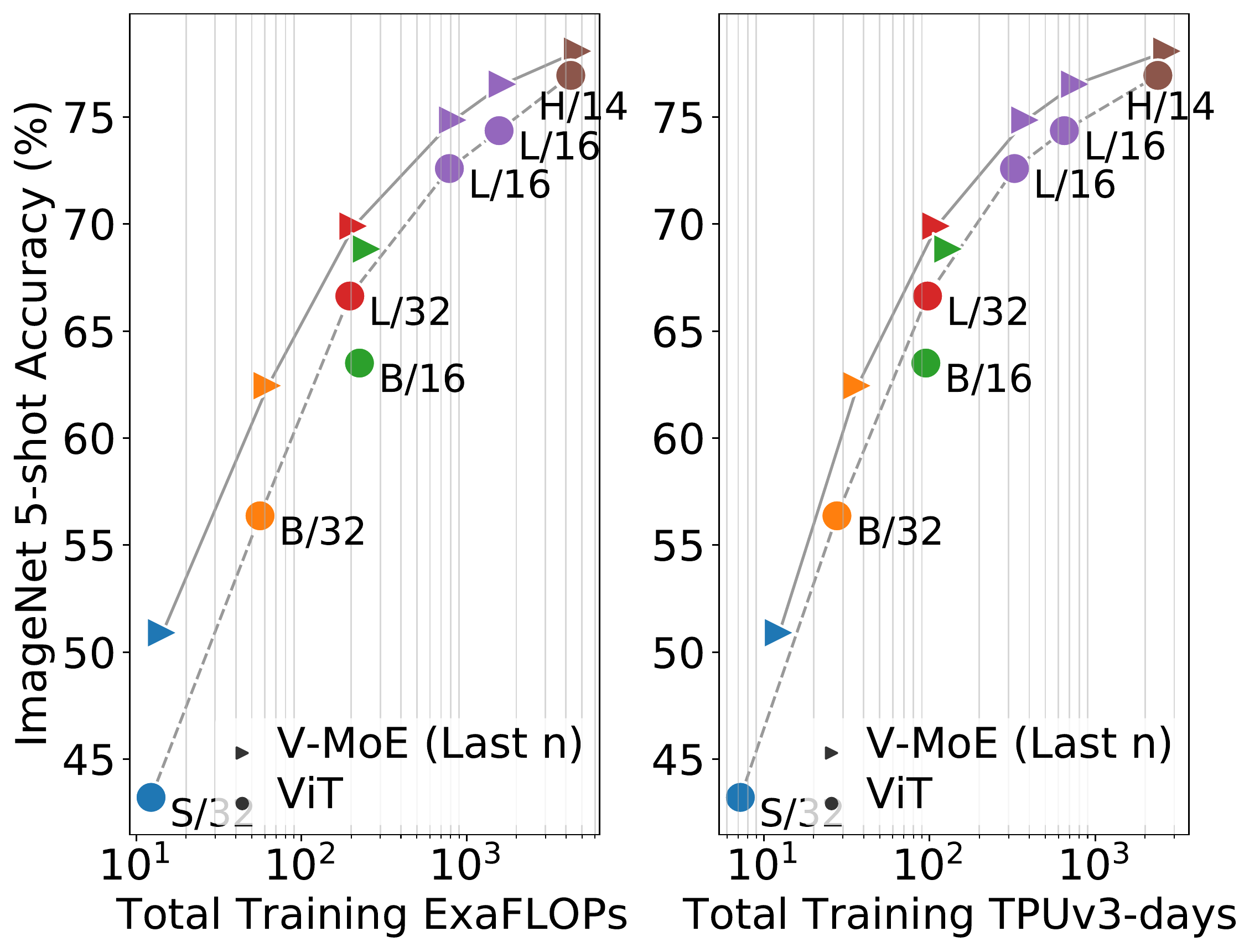}
  \caption{ImageNet 5-shot}
  \label{im:imagenet5shot_vs_exaflops_and_days}
\end{subfigure}
\caption{%
\textbf{JFT-300M Precision@1 and ImageNet 5-shot accuracy.}
Colors represent different ViT variants, markers represent either standard $\densesym{}$ViT or $\lastsym{}$V-MoEs on the last $n$ even blocks.
The lines represent the Pareto frontier of ViT (dashed) and V-MoE (solid) variants.
}
\label{im:performance_vs_exaflops_and_days}
\end{figure}

JFT is a multilabel dataset, so we measure model performance via precision@1 (see \cref{sec:prec_at_1_jft} for details). Note that as in previous works~\cite{dosovitskiy2020image}, hyperparameters were tuned for transfer performance, and JFT precision could be improved at the expense of downstream tasks e.g. by reducing weight decay.
\Cref{im:upstream_vs_exaflops_and_days} shows the quality of different \abbv{} and ViT variants with respect to total training compute and time.
It shows models that select $k=2$ experts and place MoEs in the last $n$ even blocks 
($n=5$ for \abbv{}-H, $n=2$ otherwise), but the best results are achieved by \abbv{}-H/14 \emph{Every-2} (see \cref{tab:main_models}, 14 is the patch size). 
See \cref{sec:model_table} for results of all models.

Expert models provide notable gains across all model sizes, for only a mild increase in FLOPs, 
establishing a new Pareto frontier (gray lines). Alternatively, we can match or improve performance 
of ViT models at lower cost (e.g.\ \abbv{}-L/16 improves upon ViT-H/14).
Similar conclusions hold for training time, which includes communication overhead of dispatching data across devices.

\subsection{Linear few-shot results}
We evaluate the quality of the representations learned using few-shot linear transfer.
Given training examples from the new dataset $\{ (X, Y)_i \}$, we use the pre-trained model $\mathcal{M}$ to extract a fixed representation $\mathcal{M}(x_i)$ of each image. We fit a linear regression model mapping  $\mathcal{M}(x_i)$ to the one-hot encoding of the target labels $Y_i$, following \cite{dosovitskiy2020image} (see \cite[Chapter~5]{hastie2017elements} for background).

\Cref{im:imagenet5shot_vs_exaflops_and_days} shows that the upstream gains are preserved under $5$-shot ImageNet evaluation,
considering both compute and time; in other words, the quality of the representations learned by \abbv{} also outperforms
ViT models when looking at a new task. \cref{tab:main_models} further shows the results on $\{1, 10\}$-shot for some selected models,
and the full detailed results are available in \cref{sec:model_table}.

\begin{figure}[tb]\RawFloats\TopFloatBoxes
\begin{floatrow}[2] %
\floatbox[{\capbeside\thisfloatsetup{capbesideposition={right,top}}}]{figure}[\FBwidth]{}%
{\includegraphics[width=0.48\textwidth]{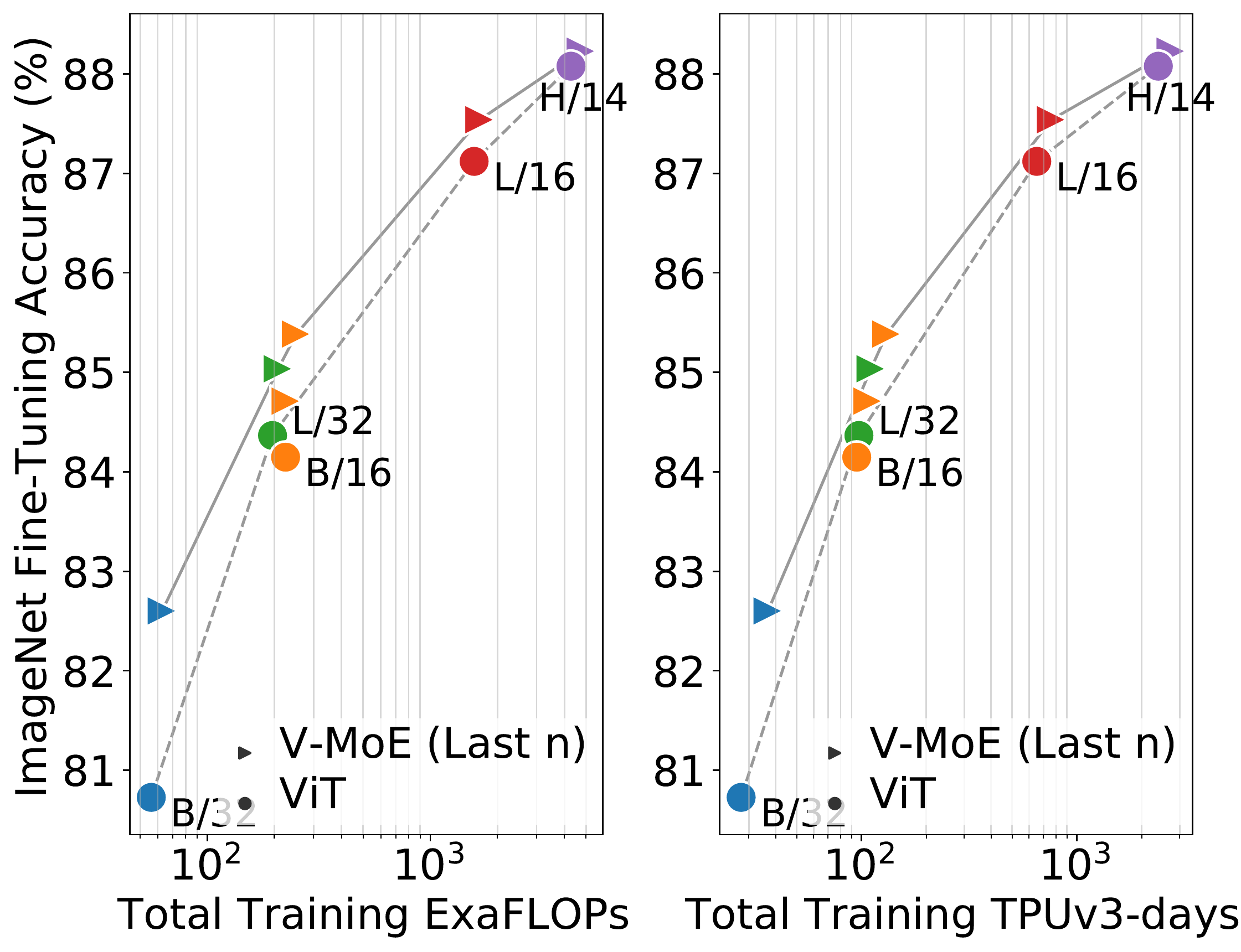}}%
\ttabbox[\Xhsize]{}{%
\begin{floatrow}[1]
\ffigbox[\Xhsize]{}{%
\caption{\textbf{ImageNet Fine-Tuning Accuracy}. Colors represent different VIT variants, markers represent either standard $\densesym{}$ViT or $\lastsym{}$V-MoEs on the last $n$ even blocks. %
Lines show the Pareto frontier of VIT (dashed) and V-MoE (solid).\label{fig:imagenet2012_test_acc_flops_days}}}
\end{floatrow}%
\vspace{3ex}
\begin{floatrow}[1]
\ttabboxsideright[\FBwidth]%
{\begin{tabular}{@{}l@{\hspace{1ex}}rr@{}}
\toprule
& ViT & \abbv{} \\ 
\midrule
L/16 & 76.3$_{\pm 0.5}$ & 77.2$_{\pm 0.4}$ \\
H/14 & 77.6$_{\pm 0.2}$ & 77.8$_{\pm 0.4}$ \\
\bottomrule
\end{tabular}}%
{\caption{\textbf{VTAB.} Scores and 95\% confidence intervals for ViT and \abbv{}.\label{tab:vtab}}}%
\end{floatrow}%
}%
\end{floatrow}
\end{figure}

\subsection{Full fine-tuning results}
The typically most performant approach for Transfer Learning \cite{dhillon2019baseline} consists of replacing the
upstream classification head with a new task-specific one and fine-tuning the whole model.
Though one may expect that massive models like V-MoEs require special handling for fine-tuning, we broadly follow the standard fine-tuning protocol for Vision Transformers. 
We use the auxiliary loss during fine-tuning as well, although we observe that it is often not needed in this step, as the router is already well trained.
We explore the two sets of tasks considered therein:

\textbf{Full data.}
We follow the setup of \cite{dosovitskiy2020image}, except that we apply a dropout rate of 0.1 on the expert MLPs (as done in \cite{fedus2021switch}), 
and we halve the number of fine-tuning steps for all datasets other than ImageNet.
\Cref{fig:imagenet2012_test_acc_flops_days} shows the results on ImageNet (averaged over three runs).
Here, \abbv{} also performs better than dense counterparts, though we suspect the fine-tuning protocol could be further improved and 
tailored to the sparse models. See \cref{tab:model_table} for all details, including results on other datasets.

\textbf{Low-data regime.}
On the VTAB benchmark, we use a similar setup and hyperparameter budget as \cite{dosovitskiy2020image} 
(but fine-tune with half the schedule length). \Cref{tab:vtab} shows that, while performance is similar for \abbv{}-H/14, experts
provide significant gains at the ViT-L/16 level,
indicating that despite the large size of these models, they can still be fine-tuned with small amounts of data and no further tricks.

\begin{table}[btp]
\caption{\label{tab:main_models} Main \abbv{} \& VIT models;
\cref{tab:model_table} shows results for additional models and datasets.}
\resizebox{\textwidth}{!}{\addtolength{\tabcolsep}{-2pt}
\begin{tabular}{@{}lrrrrrrrrr@{}} 
\toprule
 Model & Params & JFT prec@1 & IN/1shot & IN/5shot & IN/10shot & IN/Fine-t. & ExaFLOPs & TPUv3-days \\
\midrule
VIT-H/14 & 656M & 56.68 & 62.34 & 76.95 & 79.02 & 88.08 & 4.27k & 2.38k \\
\abbv{}-L/16, Every-2 & 3.4B & 57.65 & 62.41 & 77.10 & 79.01 & 87.41 & 2.17k & 1.20k  \\
\abbv{}-H/14, Last-5 & 2.7B & 60.12 & 62.95 & 78.08 & 80.10 & 88.23 & 4.75k & 2.73k  \\
\abbv{}-H/14, Every-2 & 7.2B & 60.62 & 63.38 & 78.21 & 80.33 & 88.36 & 5.79k & 3.47k  \\
\midrule
\abbv{}-15B, Every-2 & 14.7B & --- & 68.66 & 82.78 & 84.29 & 90.35 & 33.9k & 16.8k  \\
NFNet-F4+ \cite{brock2021high} & 527M & --- & --- & --- & --- & 89.20 & --- & 1.86k \\
MPL \cite{pham2020meta} & 480M & --- & --- & --- & --- & 90.20 & --- & 22.5k \\
\bottomrule
\end{tabular}%
}
\end{table}

\subsection{Scaling up V-MoE}
Finally, we test how well \abbv{} can scale vision models to a very large number of parameters, while continuing to improve performance.
For this, we increase the size of the model and use a larger pre-training dataset:
JFT-3B is a larger version of JFT-300M, it contains almost 3B images and is noisily annotated with 30k classes.
Inspired by \cite{zhai2021scaling}, we apply the changes detailed in \cref{app:vitg_tricks}, and train a 48-block \abbv{} model, with every-2 expert placement (32 experts and $k=2$), resulting in a model with 14.7B parameters, which we denote by \abbv{}-15B.

We successfully train \abbv{}-15B, which is, as far as we are aware, the largest vision model to date. It has an impressive accuracy of 82.78\% on 5-shot ImageNet and 90.35\% when fully fine-tuned, as shown
in \cref{sec:model_table}, which also includes more details about the model.
Training this model required 16.8k TPUv3-core-days.
To contextualize this result, the current state of the art on ImageNet is Meta Pseudo-Labelling (MPL)~\cite{pham2020meta}.
MPL trains an EfficientNet-based model on unlabelled JFT-300M using ImageNet pseudo-labelling, achieving 90.2\% while requiring 22.5k TPUv3-core-days.

\section{Skipping Tokens with Batch Prioritized Routing}
\label{sec:skip_patch}

\begin{figure}[tb]
\centering
\includegraphics[width=1.0\textwidth]{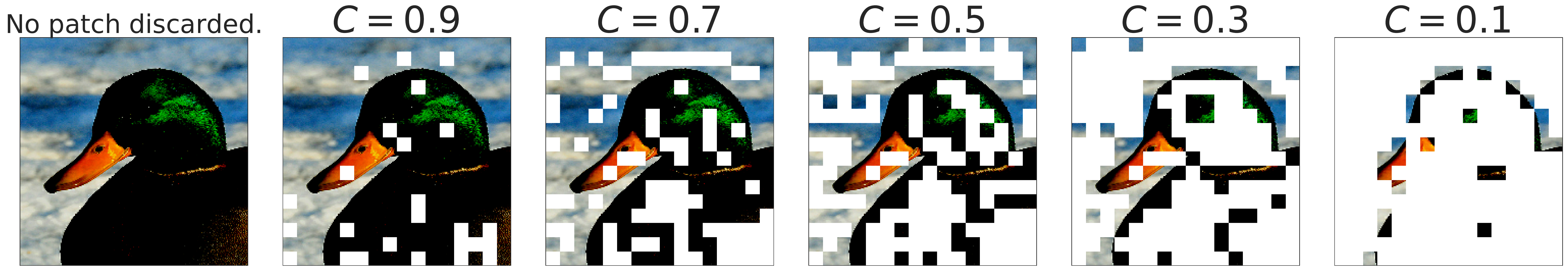}
\caption{White patches are discarded tokens in the first layer of experts, for different capacities,
using \maxrouting{} (\cref{sec:max_routing}) with a \abbv{}-H/14. See \cref{app_skip_patch_images} for more examples.}
\label{im:patch_discard}
\end{figure}

We present a new routing algorithm that allows the model to prioritize important tokens (corresp. patches).
By simultaneously reducing the capacity of each expert, we can discard the least useful tokens. 
Intuitively, not every patch is equally important to classify a given image, e.g., most background patches can be dropped to let the model only focus on the ones with the relevant entities.

\subsection{From Vanilla Routing to \maxrouting{}}\label{sec:max_routing}
With the notation from~\cref{sec:model},
the routing function $g$ is applied row-wise to a batch of inputs $\mathbf{X} \in \mathbb{R}^{N\cdot P \times D}$.
A batch contains $N$ images composed of $P$ tokens each;
each row of $\mathbf{X}$ corresponds to the $D$-dimensional representation of a particular token of an image.
Accordingly, $g(\mathbf{X})_{t,i} \in \mathbb{R}$ denotes the routing weight for the $t$-th token and the $i$-th expert. 

In all routing algorithms considered, for $i < j$, every TOP-$i$ assignment has priority over any TOP-$j$ assignment. The router first tries to dispatch \textit{all} $i^\text{th}$ expert choices before assigning \textit{any} $j^\text{th}$ choice%
\footnote{A token may however successfully assign all its TOP-$k$ choices while another may not allocate a single one. This can happen for instance if the latter selects very popular experts that run out of capacity.}.

Given the TOP-$i$ position, the default---or \textit{vanilla}---routing, as used in \cite{shazeer2017outrageously,lepikhin2020gshard,fedus2021switch}, 
assigns tokens to experts as follows. It sequentially goes over the rows of $g(\mathbf{X})$ and assigns each token to its TOP-$i$ expert \emph{when} %
the expert's buffer is not full.
As a result, priority is given to tokens depending on the rank of their corresponding row. While images in a batch are randomly ordered, tokens within an image follow a pre-defined \emph{fixed} order. %
The algorithm is detailed in \cref{algo:default_patch_assignment} of \cref{app_skip_patch_training}.

\textbf{\maxrouting{}} (BPR).
To favour the ``most important'' tokens, we propose to compute a \emph{priority score} $s(\mathbf{x})$ on each token, and sort $g(\mathbf{X})$ accordingly before proceeding with the allocation.
We sort tokens based on their maximum routing weight, formally $s(\mathbf{X})_t = \max_i g(\mathbf{X})_{t, i}$.
The sum of TOP-$k$ weights, i.e. $s(\mathbf{X})_t = \sum_i g(\mathbf{X})_{t, i}$, worked equally well. 
These two simple approaches outperformed other options we explored, e.g., directly parameterising and learning the function $s$.

We reuse the router outputs as a proxy for the priority of allocation. Our experiments show this preserves the performant predictive behaviour of the model, even though the router outputs primarily encode how well tokens and experts can be paired, not the token's ``importance'' for the final classification task.
\Cref{im:patch_discard} visualizes token prioritisation with \maxrouting{} for increasingly small capacities.
Since all tokens across all images in the batch $\mathbf{X}$ compete with each other, different images may receive different amounts of compute. We summarize BPR in \cref{algo:max_weight_patch_assignment}, in \cref{app_skip_patch_training}.

\begin{figure}\RawFloats\TopFloatBoxes
\centering
\begin{floatrow}[2] %
\ffigbox[\FBwidth]{\includegraphics[width=.48\textwidth]%
{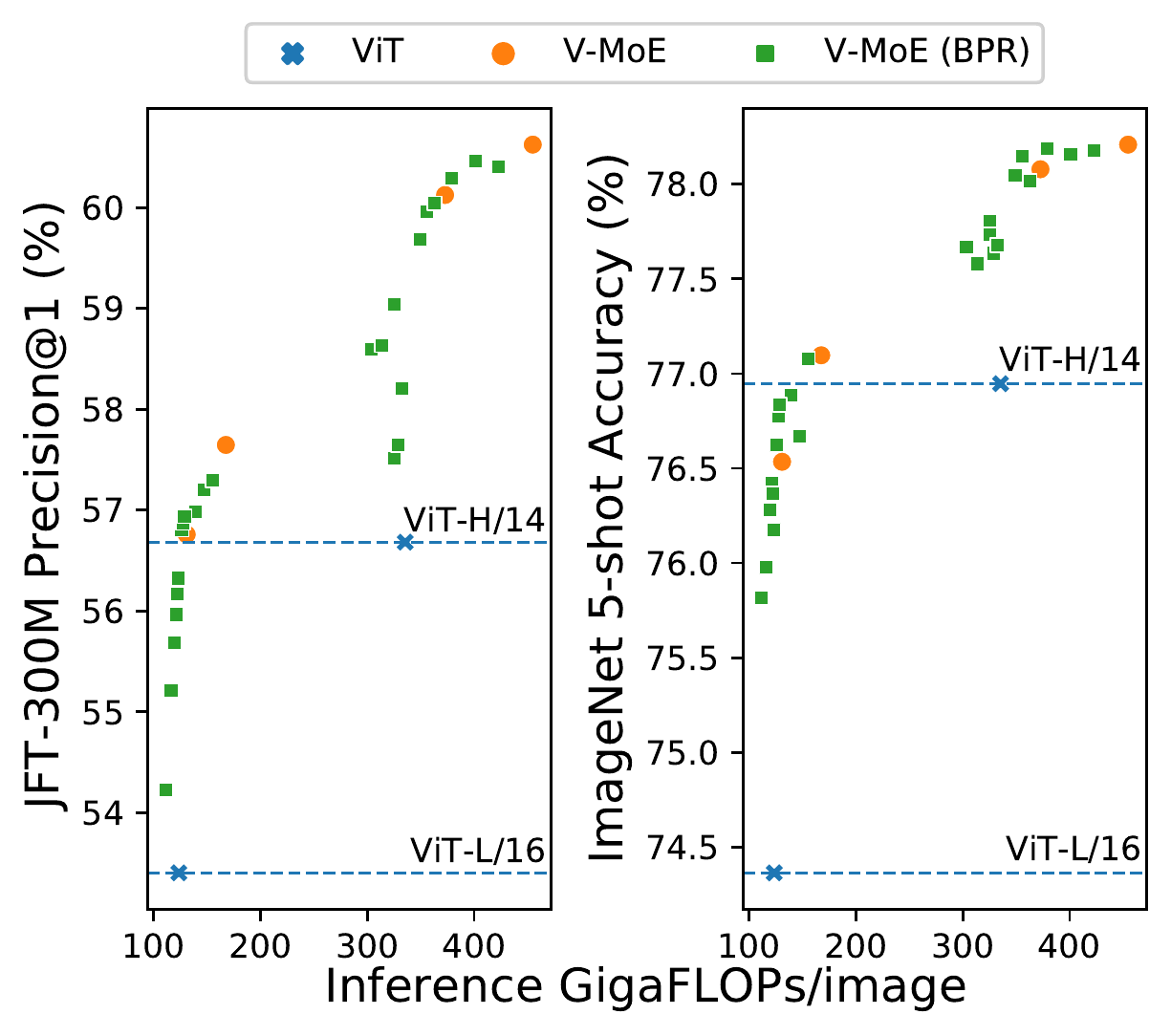}}{\caption{%
\textbf{Reducing compute with priority routing.} Performance vs.\ \emph{inference} FLOPs for large models.
\abbv{}s with the original vanilla routing are represented by $\bullet$, while $\blacksquare$ shows \abbv{}s where BPR and a mix of ${C \in \{0.6, 0.7, 0.8\}}$ and ${k \in \{1, 2\}}$ are used to reduce compute. 
ViT models shown as $\mathbf{x}$.%
\label{im:upstream_5shot_at_inference}%
}%
}%
\ffigbox[\Xhsize]{\includegraphics[width=.52\textwidth]%
{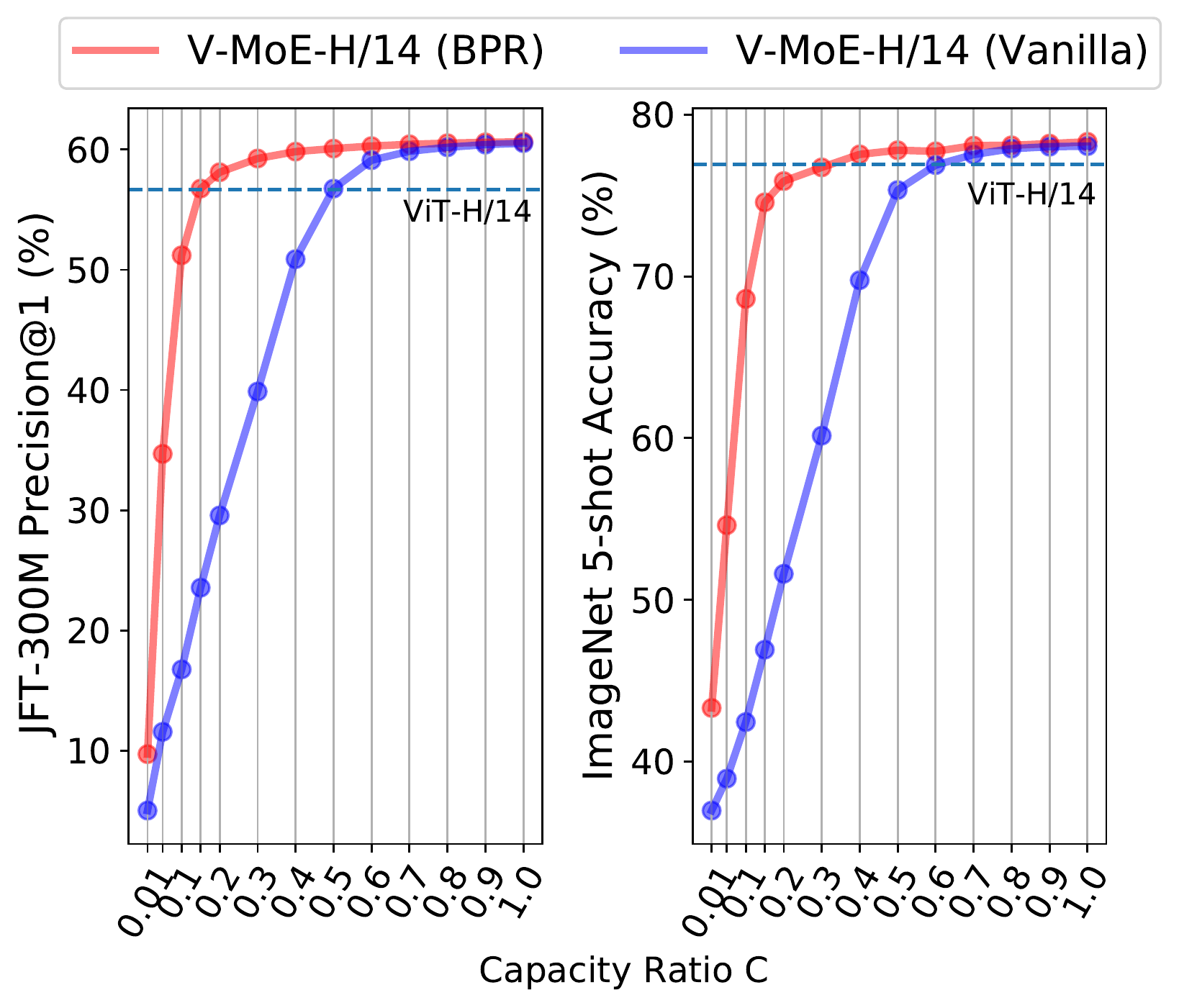}}{\caption{%
\textbf{Priority routing works where vanilla fails.}
Performance vs. \emph{inference} capacity ratio for a \abbv{}-H/14 model with $k=2$.
Even for large $C$'s BPR outperforms vanilla; at low $C$ the difference is stark. BPR is competitive with dense by processing only {15-30\%} of the tokens.%
\label{im:vit_h_max_routing}%
}%
}%
\end{floatrow}
\vspace{-0.5cm}
\end{figure}

\subsection{Skip tokens with low capacity \texorpdfstring{$C$}{C}}
\maxrouting{} opens the door to reducing the buffer size by smartly selecting which tokens to favor. This can have a dramatic impact in the computational cost of the overall sparse model.
We discuss now inference and training results with $C$ defined in \cref{sec:expert_capacity} in the regime  $C \ll 1$. 

\textbf{At inference time.}
Prioritized routing is agnostic to how the model was originally trained.
\cref{im:vit_h_max_routing} shows the effect of reducing compute at inference time by using BPR versus vanilla routing, on a \abbv{}-H/14 model \emph{trained} using vanilla routing.
The difference in performance between both methods is remarkable ---especially for $C \le 0.5$, where the model truly starts fully dropping tokens, as $k=2$.
Also, BPR allows the model to be competitive with the dense one even at quite low capacities.
As shown in \cref{im:upstream_5shot_at_inference} for \abbv{}-L/16 and \abbv{}-H/14, \maxrouting{} and low $C$ allow \abbv{} to smoothly trade-off performance and FLOPS at inference time, quite a unique model feature.
More concretely, \cref{tab:matching_performance_max_routing} shows \abbv{} models can beat the dense VIT-H performance by using less than half the FLOPs and less than 60\% of the runtime.
Conversely, we can match the inference FLOPs cost and preserve a one-point accuracy gain in ImageNet/5shot and almost three-point in JFT precision at one (\cref{tab:matching_cost_max_routing}).
Dense models generally require less runtime for the same amount of FLOPs due to the data transfer involved in the \abbv{} implementation.

\textbf{At training time.}
\maxrouting{} can also be leveraged during training.
In \cref{app_skip_patch_training} we show how expert models with max-weight  routing can match the dense performance while saving around 20\% of the total training FLOPs, and strongly outperform vanilla with a similar FLOP budget.
\section{Model Analysis}
\label{sec:model_analysis}

Although large-scale sparse MoEs have led to strong performance~\cite{fedus2021switch, lepikhin2020gshard, shazeer2017outrageously},
little is known and understood about how the internals of those complex models work.
We argue that such exploratory experiments can inform the design of new algorithms. In this section, we provide the first such analysis at this scale, which guided the development of the algorithms presented in the paper.

\textbf{Specialized experts.}
Intuitively, routers should learn to distribute images across experts based on their similarity.
For instance, if the model had three experts, and the task mainly involved three categories---say animals, cars, and buildings---one would expect an expert to specialize in each of those.
We test this intuition, with some obvious caveats: (a) experts are placed at several network depths, (b) $k$ experts are combined, and (c) routing happens at the token rather than the image level.

\begin{figure}[tb]
\centering
\includegraphics[width=1.0\textwidth]{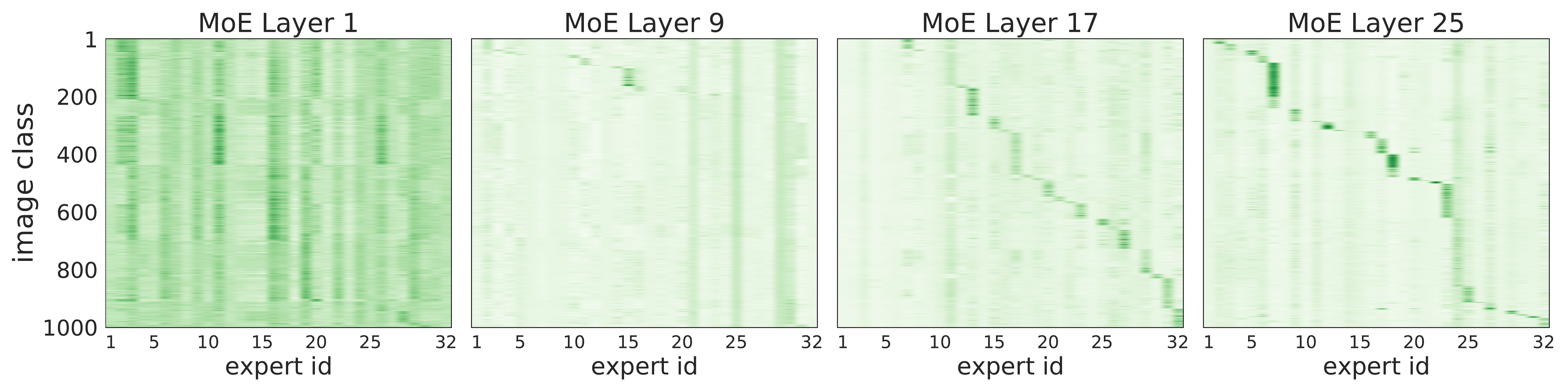}
\caption{\textbf{Deeper routing decisions correlate with image classes}.
We show 4 MoE layers of a \abbv{}-H/14. The $x$-axis corresponds to the 32 experts in a layer. The $y$-axis are the 1\,000 ImageNet classes; orderings for both axes are different across plots.
For each pair (expert $e$, class $c$) we show the average routing weight for the tokens corresponding to all images with class $c$ for that particular expert $e$. \Cref{im:routes_class_experts_all} includes all the remaining layers; see \Cref{app_analysis_specialized_experts} for details.}
\label{im:routes_class_experts}
\end{figure}

\Cref{im:routes_class_experts} illustrates how many images of a given ImageNet class use each expert. The plots were produced by running a fine-tuned \abbv{}-H \emph{Every-2} model. Interestingly, we saw similar
patterns with the upstream model without fine-tuning.
Experts specialize in discriminating between small sets of classes (those primarily routed through the expert).
In earlier MoE layers we do not observe this. Experts may instead focus on aspects common to all classes (background, basic shapes, colours) - for example, \cref{im:routes_patch_experts} (\cref{app_analysis}) shows correlations with patch location in earlier layers.

\textbf{The value of routers.}
After training a sparse MoE, it is natural to study the usefulness of the learned routers, in the light of several pitfalls.
For example, the routers may just act as a load balancer if experts end up learning very similar functions, or the routers may simply choose poor assignments.
In~\cref{app_analysis_value_routers}, we replace, after training, one router at a time with a uniformly random router.
The models are robust to early routing changes while more sensitive to the decisions in the last layers.

\textbf{Routing weights distributions.}
We analyse the router outputs in~\cref{app_analysis_routing_weights_distribution}, and observe the distribution of selected weights varies wildly across different mixture of experts layers.

\textbf{Changing $k$ at inference time.}
We have observed expert models are remarkably flexible.
Somewhat surprisingly, sparse models are fairly robust to mismatches between their training and inference configurations.
In~\cref{app_analysis_routing_changing_k}, we explore the effect of training with some original value of $k$ while applying the model at inference time with a different $k^\prime \neq k$.
This can be handy to control (decrease or increase) the amount of FLOPs per input in a particular production system.
\section{Related work}

\textbf{Conditional Computation.} To grow the number of model parameters without proportionally increasing the computational cost, conditional computation~\cite{bengio2013deep,davis2013low,cho2014exponentially} only activates some relevant parts of the model in an \textit{input-dependent} fashion, like in decision trees~\cite{breiman1984classification}.
In deep learning,
the activation of portions of the model can use stochastic neurons~\cite{bengio2013estimating} or reinforcement learning~\cite{bengio2015conditional, denoyer2014deep, rosenbaum2017routing}.

\textbf{Mixture of Experts.} 
MoEs~\cite{jacobs1991adaptive, jordan1994hierarchical, chen1999improved, yuksel2012twenty} combine the outputs of sub-models known as \textit{experts} via a \textit{router} in an input-dependent way. MoEs have successfully used this form of conditional computation in a range of applications~\cite{gavrila2007multi, hu1997patient, tani1999learning, sminchisescu2004learning, zeevi1997time}. An input can select either all experts~\cite{eigen2013learning} or only a sparse mixture thereof as in recent massive language models~\cite{shazeer2017outrageously, lepikhin2020gshard, fedus2021switch}.

\textbf{MoEs for Language.}
MoEs have recently scaled language models up to trillions of parameters. 
Our approach is inspired by~\cite{shazeer2017outrageously} who proposed a top-$k$ gating in LSTMs, with auxiliary losses ensuring the expert balance~\cite{hansen1999combining}. \cite{lepikhin2020gshard} further scaled up this approach for transformers, showing strong gains for neural machine translation. With over one trillion parameters and one expert per input, \cite{fedus2021switch} sped up pre-training compared to a dense baseline~\cite{raffel2019exploring} while showing gains thanks to transfer and distillation. \cite{lewis2021base} alternatively enforced a balanced routing by solving a linear assignment problem.

\textbf{MoEs for Vision.} For computer vision, previous work on MoEs~\cite{eigen2013learning, ahmed2016network, gross2017hard, abbas2020biased, wang2020deep, pavlitskaya2020using, yang2019condconv} focused on architectures whose scale is considerably smaller than that of both language models and our model. In DeepMoE~\cite{wang2020deep}, the ``experts'' are the channels of convolutional layers that are adaptively selected by a multi-headed sparse gate. This is similar to~\cite{yang2019condconv} where the kernels of convolutional layers are activated on a per-example basis.
Other approaches use shallow MoEs, learning a \textit{single router}, either disjointly~\cite{gross2017hard} or jointly~\cite{ahmed2016network}, together with CNNs playing the role of experts. \cite{abbas2020biased} further have a cost-aware procedure to bias the assignments of inputs across the experts. 
Unlike shallow MoEs, we operate with up to several tens of routing decisions \emph{per token} along the depth of the model. Scaling up routing depth was marked as a major challenge in~\cite{ramachandran2018diversity}, which we successfully tackle in our work.

\section{Conclusions}
\label{sec:conclusions}
We have employed sparse conditional computation to train some of the largest vision models to date, showing significant improvements in representation learning and transfer learning.
Alongside \abbv{}, we have proposed \maxrouting{}, which allows successful repurposing of \textit{model} sparsity to introduce sparsity \textit{with respect to the inputs}. This can be done without further adapting the model, allowing the re-use of trained models with sparse conditional computation.

This has interesting connotations for recent work in NLP using sparse models; recent analysis shows model sparsity is the most promising way to reduce model CO\textsubscript{2} emissions~\cite{patterson2021carbon} and that 90\% of the footprint stems from inference costs --- we present an algorithm which takes the most efficient models and makes them even \textit{more} efficient without any further model adaptation.

This is just the beginning of conditional computation at scale for vision; extensions include
scaling up the expert count, reducing dependance on data and improving transfer of the representations produced by sparse models. Directions relating to heteregonous expert architectures and conditional variable-length routes should also be fruitful. We expect increasing importance of sparse model scaling, especially in data rich domains such as large scale multimodal or video modelling.

\clearpage

\begin{ack}
We thank Alex Kolesnikov, Lucas Beyer and Xiaohua Zhai for providing continuous help and details about scaling ViT models;
Alexey Dosovitskiy, who provided some of the pre-trained ViT models;
Ilya Tolstikhin, who suggested placing experts only in the last layers;
Josip Djolonga for his early review of the manuscript;
Dmitry Lepikhin for providing details about the original GShard implementation;
Barret Zoph and Liam Fedus for insightful comments and feedback;
James Bradbury, Blake Hechtman and the rest of JAX and TPU team who helped us running our models efficiently,
and many others from Google Brain for their support.
\end{ack}

\bibliographystyle{abbrv}
\bibliography{main}

\clearpage
\appendix
\clearpage
\section{Further details about the Vision Mixture of Experts}
\label{sec:app_moe_model}

In this section, we provide additional details about the definition of \abbv{}.

\subsection{Ablation on the modification of the routing function}

Our formulation is similar to that in~\cite{shazeer2017outrageously}, except that we apply the ``top $k$'' operation \textit{after} normalization of the experts weights, i.e. $\text{TOP}_k$ and softmax are applied in reverse order.

We choose this ordering because the original formulation from~\cite{shazeer2017outrageously} cannot be trained easily in the case of $k=1$; it would lead to zero gradient with respect to $\mathbf{x}$ and $W$ almost everywhere.
Moreover, even for $k > 1$, we found our alternative formulation to perform better (see \cref{tab:routing_comparison}). 

\begin{table}
\begin{center}
\resizebox{\textwidth}{!}{
\begin{tabular}{ c|c|c|c|c|c|c|c } 
 Model & Routing Function & Proposed in & K & prec@1 & ImageNet/1shot & ImageNet/5shot & ImageNet/10shot  \\
 \hline \hline
 VIT-S/32 & TOP-K(softmax) & This work & 2 & 34.15 & 38.42 & 53.11 & 56.06 \\
  VIT-S/32 & softmax(TOP-K) & \cite{shazeer2017outrageously} & 2 & 33.75 & 35.59 & 50.21 & 53.63 \\
\end{tabular}
}
\vspace*{3mm}
\caption{\label{tab:routing_comparison} Comparison of routing functions.}
\end{center}
\end{table}

\subsection{Description of the load balancing losses}

We describe below the regularizers that we use to enforce a balanced usage of the experts.
Those regularizers present slight modifications with respect to their original definitions in~\cite{shazeer2017outrageously}. 

\begin{table}
\begin{center}
\begin{tabular}{ c|c|c|c|c } 

 Token & Expert 1 & Expert 2 & Expert 3 & Selected Expert \\
  & $w_1$ & $w_2$ & $w_3$ &  \\
 \hline
 $x_1$ & 0.9 & 0.5 & 0.1 & Expert 1 \\ 
 $x_2$ & 0.1 & 0.5 & 0.9 & Expert 3 \\
 $x_3$ & 0.9 & 0.5 & 0.1 & Expert 1 \\ 
 $x_4$ & 0.1 & 0.5 & 0.9 & Expert 3 \\
 $\cdots$ & $\cdots$ & $\cdots$ & $\cdots$ & $\cdots$ \\
\end{tabular}
\vspace*{3mm}
\caption{\label{tab:load_loss_required} Simple example ($k=1$) where average weights are balanced, but Expert 2 is never selected.}
\end{center}
\end{table}

\textbf{Importance Loss.}
We incentivize a balanced usage of experts via an importance loss.
The importance of expert $i$ for a batch of images $\mathbf{X}$ is defined as the normalized routing weight corresponding to expert $i$ summed over images:
\begin{equation}\label{eq:importance_i_definition}
    \text{Imp}_i(\mathbf{X}) := \sum_{\mathbf{x} \in \mathbf{X}} \text{softmax} (W\mathbf{x})_i,
\end{equation}
where $W$ is the layer-specific weight matrix for the router. 
We use the squared coefficient of variation of the importance distribution over experts, $\text{Imp}(\mathbf{X}) := \{ \text{Imp}_i(\mathbf{X}) \}_{i=1}^E$:
\begin{equation}\label{eq:importance_loss}
    \mathcal{L}_{\text{Imp}}(\mathbf{X}) = \left( \frac{\text{std}(\text{Imp}(\mathbf{X}))}{\text{mean}(\text{Imp}(\mathbf{X}))} \right)^2 \propto \text{var}(\text{Imp}(\mathbf{X})).
\end{equation}
\cite{shazeer2017outrageously} proposed a similar loss, while in their case token $\mathbf{x}$ contributed to the importance of expert $i$ in \cref{eq:importance_i_definition} \emph{only} if $i$ was indeed selected for $\mathbf{x}$.
We observed some modest empirical benefits thanks to \cref{eq:importance_loss}.

\textbf{Load Loss.}
The importance loss seeks to guarantee that all experts have on average similar output routing weights.
Unfortunately, it is not difficult to think of routing configurations where these weights are balanced overall, but, still, some small subset of experts get all the assignments (see \cref{tab:load_loss_required}).

Ideally, we would like to also explicitly balance the number of assignments.
This quantity is discrete; therefore it is not differentiable, and we need to rely on a proxy.
Following the proposal in \cite{shazeer2017outrageously}, for each expert $i$ and token $\mathbf{x}$, we compute the probability of $i$ being selected ---i.e., being among the top-$k$--- for $\mathbf{x}$ if we were to re-sample \emph{only} the noise for expert $i$.
For simplicity, we slightly modify the definition in \cite{shazeer2017outrageously}.
For each token $\mathbf{x}$, we define the score threshold above which experts were selected; this is simply the $k$-th maximum score:
\begin{equation}
    \text{threshold}_k(\mathbf{x}) := \text{max}_{k\text{-th}} \left( W\mathbf{x} + \epsilon \right),
\end{equation}
where $\epsilon$ was the noise vector originally sampled during the forward pass.
Then, for each expert $i$ we compute the probability of $i$ being above the threshold if we were to only re-sample its noise:
\begin{equation}
    p_{i}(\mathbf{x}) := \mathbf{P}((W\mathbf{x})_i + \epsilon_{\text{new}} \ge \text{threshold}_k(\mathbf{x})) = \mathbf{P}( \epsilon_{\text{new}} \ge \text{threshold}_k(\mathbf{x}) - (W\mathbf{x})_i).
\end{equation}
The probability is defined over $\epsilon_{\text{new}} \sim \mathcal{N}(0, \sigma^2)$, with $\sigma = 1/E$.
The load for expert $i$ over batch $\mathbf{X}$ is:
\begin{equation}
    \text{load}_i(\mathbf{X}) = \sum_{\mathbf{x} \in \mathbf{X}} p_i(\mathbf{x}).
\end{equation}
Finally, the load loss corresponds to the squared coefficient of variation of the load distribution:
\begin{equation}\label{eq:load_loss}
    \mathcal{L}_{\text{load}}(\mathbf{X}) = \left( \frac{\text{std}(\text{load}(\mathbf{X}))}{\text{mean}(\text{load}(\mathbf{X}))} \right)^2,  \qquad \text{load}(\mathbf{X}) := \{ \text{load}_i(\mathbf{X}) \}_{i=1}^E.
\end{equation}

\textbf{Final Auxiliary Loss.}
The final auxiliary loss is just the average over both:
\begin{equation}
    \mathcal{L}_{\text{aux}}(X) =  \frac{1}{2} \ \mathcal{L}_{\text{imp}}(X) + \frac{1}{2} \ \mathcal{L}_{\text{load}}(X).
\end{equation}
The overall loss is: $\mathcal{L}(X) = \mathcal{L}_{\text{classification}}(X) + \lambda \ \mathcal{L}_{\text{aux}}(X)$, for some hyperparameter $\lambda > 0$.
We set $\lambda = 0.01$ in all our experiments, observing that this choice was robust and not sensitive.
 
\clearpage
\section{Transfer Experiment Details}
\label{sec:experiment_details}

\subsection{Additional fine-tuning datasets}\label{sec:additional_datasets}
Alongside finetuning on ImageNet (ILSVRC2012\cite{deng2009imagenet}), we also train on four other datasets shown in \cref{tab:finetuning_datasets}.
\begin{table}[]
\caption{\textbf{Finetuning datasets.}
\label{tab:finetuning_datasets}}
\begin{tabular}{@{}lll@{}}
\toprule
Dataset                                        & Num examples & Num classes \\ \midrule
CIFAR10 \cite{krizhevsky2009learning}           & 50\,000       & 10          \\
CIFAR100 \cite{krizhevsky2009learning}          & 50\,000       & 100         \\
Oxford Flowers 102 \cite{nilsback2008automated} & 1\,020        & 102         \\
Oxford-IIT Pet \cite{parkhi2012cats}            & 3\,680        & 37          \\
ImageNet (ILSVRC2012 \cite{deng2009imagenet})   & 1\,281\,167   & 1\,000      \\
\bottomrule
\end{tabular}
\end{table}
For the Visual Task Adaptation Benchmark (VTAB\cite{zhai2019large}), we finetune on 19 datasets with 1\,000 datapoints per class. We refer interested readers to the original work by Zhai et al.~\cite{zhai2019large} for more details, but in brief, the benchmark consists of 3 task categories:

\begin{itemize}
    \item \textbf{Natural tasks}
    \small{\textcolor{gray}{CalTech101 \cite{caltech101} $\cdot$ CIFAR100 \cite{krizhevsky2009learning} $\cdot$ Street View House Numbers (SVHN - \cite{svhn}) $\cdot$ Describable Textures (DTD - \cite{dtd}) $\cdot$ Oxford Flowers \cite{nilsback2008automated} $\cdot$ Oxford Pets \cite{parkhi2012cats}}}
    \normalsize These tasks contain `classical' natural real-world images obtained with a camera.
    \item \textbf{Specialised tasks}
    \small{\textcolor{gray}{EuroSAT \cite{eurosat} $\cdot$ Diabetic Retinopothy \cite{retino} PatchCamelyon \cite{camelyon} $\cdot$ Remote Sensing Image Scene Classification (RESISC - \cite{resisc})
    }}
    \normalsize These are datasets of images which were captured with specialised (medical, satellite etc) photographic equipment.
    \item \textbf{Structured datasets}
    \small{\textcolor{gray}{
    DeepMind Lab (Object distance prediction - \cite{zhai2019largescale}) $\cdot$ SmallNOrb (Azimuth \& Elevation prediction - \cite{smallnorb} CLEVR (Counting \& Distance prediction \cite{clevr} $\cdot$ Kitti (Vehicle distance prediction \cite{kitti}) $\cdot$ dSprites (pixel location \& orientation prediction - \cite{dsprites})
    }}
    \normalsize These assess understanding of scene structure in some way, predominately from synthetic environments. Example tasks include 3D depth estimation and counting.
\end{itemize}

\subsection{Upstream hyperparameters}\label{sec:upstream_hparams}

We present the architectural details for the upstream models in \cref{tab:model_table} (embedding size---equivalently referred to as hidden size, MLP dimension, number of Transformer blocks, etc.).
\cref{tab:upstream_hparams} shows the training hyper-parameters for our main models.
We use the original setup for each ViT model \cite{dosovitskiy2020image}.
However, ViT-S was not formally introduced in \cite{dosovitskiy2020image}, and our parameters for ViT-S (dense and sparse) do not match DeiT-Small introduced in \cite{touvron2020deit}.

\begin{table}[tb]
\centering
\caption{
\textbf{Hyper-parameter values for upstream training on JFT}.
Weight decay of 0.1 indicates that this value is applied to all model parameters (including biases), 
while (0.03, 3) indicates that 0.03 is used for the kernels and 3 for the classification head.
\label{tab:upstream_hparams}}
\begin{tabular}{lrrrrr}
\toprule
Variant   & JFT-300M Epochs & Optimizer & Base LR & LR decay & Weight Decay\\
\midrule
S/32      & 5        & Adam      & $1 \cdot 10^{-3}$ & linear & 0.1\\
B/{16,32} & 7        & Adam      & $8 \cdot 10^{-4}$ & linear & 0.1\\
L/32      & 7        & Adam      & $6 \cdot 10^{-4}$ & linear & 0.1\\  
L/16      & \{7,14\} & Adam      & $4 \cdot 10^{-4}$ & linear & 0.1\\
H/14      & 14       & Adam      & $3 \cdot 10^{-4}$ & linear & 0.1\\
V-MoE-15B & ---      & Adafactor & $8 \cdot 10^{-4}$ & rsqrt\footnote{A linear learning rate cooldown is applied at the end of training.} & (0.03, 3)\\
\bottomrule     
\end{tabular}
\end{table}

\subsection{Model modifications for scaling to \abbv{}-15B}
\label{app:vitg_tricks}
There are many changes to typical dense models which can be applied alongside model sparsity in order to scale models up. In order to scale the base architecture to which we add sparse mixture of expert layers, we make the following changes based on \cite{zhai2021scaling}:

\begin{itemize}
    \item \textbf{Low precision}: We use \texttt{bfloat16} instead of \texttt{float32} to store the gradient moving average.
    \item \textbf{Learning-rate decay:} We replace the linear schedule with an inverse square root schedule (\texttt{rsqrt}).
    \item \textbf{Weight decay:} We apply weight decay to the kernel weights in the model with value 0.03 (while biases are not regularized), except for the head kernel where we apply a stronger regularization of 3.0.
    \item \textbf{Model head:} We replace the token head~\cite{dosovitskiy2020image}---where the first token is selected---with a new self-attention based head that also includes an additional MLP \cite{zhai2021scaling}.
\end{itemize}

\subsection{Fine-tuning hyperparameters}\label{sec:finetune_hparams}
\Cref{tab:finetune_hparams} shows the hyperparameters used for finetuning. As discussed, they are broadly identical to those used in the Vision Transformer \cite{dosovitskiy2020image}, though with half the schedule length. We also apply expert dropout of 0.1 on the expert MLPs (as suggested in \cite{fedus2021switch}); this did not make a significant difference, typically marginally reducing or improving performance.
\begin{table}[tb]
\centering
\caption{
\label{tab:finetune_hparams} \textbf{Hyper-parameter values for fine-tuning on different datasets.}}
\begin{tabular}{lrrr}
\toprule
Dataset  & Steps & Base LR & Expert Dropout \\
\midrule
ImageNet & 10\,000 & \{0.003, 0.01, 0.03, 0.06\} & 0.1\\
CIFAR10  & 2\,500  & \{0.001, 0.003, 0.01, 0.03\} & 0.1\\
CIFAR100 & 2\,500  & \{0.001, 0.003, 0.01, 0.03\} & 0.1\\
Oxford-IIIT Pets & 250 & \{0.001, 0.003, 0.01, 0.03\} & 0.1\\
Oxford Flowers-102 & 250 & \{0.001, 0.003, 0.01, 0.03\} & 0.1\\
VTAB (19 tasks) & 1\,250 & 0.001 & 0.1\\
\bottomrule     
\end{tabular}
\end{table}

We finetuned the \abbv{}-15B model on ImageNet at resolution 560x560 for 30\,000 steps (i.e., about 6 epochs) with base learning rate 0.006. We used debiased Polyak averaging similar to \cite{dosovitskiy2020image} with momentum 0.999999.

\newpage 
\begin{landscape}
\subsection{Results and details for all models}\label{sec:model_table}%
\begin{table}[H]
\centering
\caption{%
\label{tab:model_table}
\textbf{Upstream, few-shot and downstream performance for dense and sparse models. Architectural details and training costs also provided.}
All \abbv{} models have $E=32$ experts and were trained with $C=1.05$. 
We specify 
the number of selected experts per token ($k$),
the number of JFT-300M epochs, 
the number of Transformer blocks ($L$),
the number of attention heads ($H$),
the patch embedding size ($D$), 
the hidden size of the MLP, 
the total number of parameters,
the JFT-300M Precision@1 (\%),
the ImageNet 1, 5 and 10-shot accuracy (\%),
the fine-tuning accuracy (\%) on ImageNet (INet/Ft.),
CIFAR10, CIFAR100, Oxford-IIIT Pets, and Oxford Flowers-102;
the total training time on a single core of a TPUv3,
and the total training compute (in exaFLOPs).}
\resizebox{\textwidth}{!}{{\setlength{\tabcolsep}{1ex}\begin{tabular}{llrrrrrrrrrrrrrrrrr}
\toprule
                Name &    $k$ &  Epochs &  Blocks &  Heads &  Embed. &   MLP &   Params & JFT-300M & INet/1s & INet/5s & INet/10s & INet/Ft. & CIFAR10 & CIFAR100 &    Pets & Flowers & TPUv3-days & ExaFLOPs \\
\midrule
            ViT-S/32 &  --- &       5 &       8 &      8 &     512 &  2048 &    36.5M &    29.05 &   29.37 &   43.21 &    46.38 &    73.73 &   97.95 &    87.20 &   91.03 &   96.78 &       7.22 &    12.27 \\
  V-MoE-S/32, Last 2 &    1 &       5 &       8 &      8 &     512 &  2048 &   166.7M &    30.93 &   30.65 &   46.06 &    49.47 &    76.32 &   98.05 &    87.93 &   92.62 &   95.88 &      10.83 &    12.50 \\
  V-MoE-S/32, Last 2 &    2 &       5 &       8 &      8 &     512 &  2048 &   166.7M &    33.26 &   35.49 &   50.90 &    54.16 &    77.10 &   98.19 &    88.86 &   93.20 &   96.50 &      12.40 &    14.40 \\
 V-MoE-S/32, Every 2 &    2 &       5 &       8 &      8 &     512 &  2048 &   296.9M &    34.00 &   37.53 &   51.75 &    54.97 &    77.08 &   98.23 &    88.50 &   94.02 &   97.86 &      17.60 &    16.53 \\
  V-MoE-S/32, Last 2 &    5 &       5 &       8 &      8 &     512 &  2048 &   166.7M &    35.49 &   38.77 &   53.60 &    56.94 &    77.59 &   98.25 &    89.25 &   93.26 &   97.31 &      18.49 &    20.44 \\
            ViT-B/32 &  --- &       7 &      12 &     12 &     768 &  3072 &   102.1M &    39.31 &   40.58 &   56.37 &    59.63 &    80.73 &   98.61 &    90.49 &   93.40 &   99.27 &      27.62 &    56.08 \\
  V-MoE-B/32, Last 2 &    1 &       7 &      12 &     12 &     768 &  3072 &   395.0M &    41.41 &   44.49 &   60.14 &    63.63 &    81.70 &   98.88 &    91.28 &   94.85 &   99.21 &      30.59 &    56.41 \\
  V-MoE-B/32, Last 2 &    2 &       7 &      12 &     12 &     768 &  3072 &   395.0M &    43.17 &   48.04 &   62.45 &    65.72 &    82.60 &   98.67 &    91.47 &   95.25 &   99.21 &      36.80 &    62.75 \\
 V-MoE-B/32, Every 2 &    2 &       7 &      12 &     12 &     768 &  3072 &   980.6M &    43.37 &   47.57 &   62.88 &    65.94 &    82.21 &   98.89 &    91.73 &   95.39 &   99.60 &      54.88 &    76.09 \\
  V-MoE-B/32, Last 2 &    5 &       7 &      12 &     12 &     768 &  3072 &   395.0M &    43.94 &   49.07 &   63.33 &    66.68 &    82.72 &   98.87 &    91.46 &   95.07 &   99.24 &      49.11 &    81.75 \\
            ViT-L/32 &  --- &       7 &      24 &     16 &    1024 &  4096 &   325.3M &    46.98 &   50.95 &   66.64 &    69.77 &    84.37 &   99.19 &    92.52 &   95.83 &   99.45 &      97.30 &   196.13 \\
  V-MoE-L/32, Last 2 &    2 &       7 &      24 &     16 &    1024 &  4096 &   845.8M &    49.68 &   54.52 &   69.90 &    72.80 &    85.04 &   99.24 &    92.50 &   96.34 &   99.08 &     110.65 &   207.94 \\
            ViT-B/16 &  --- &       7 &      12 &     12 &     768 &  3072 &   100.5M &    44.58 &   48.21 &   63.50 &    66.94 &    84.15 &   99.00 &    91.87 &   95.80 &   99.56 &      95.04 &   224.45 \\
  V-MoE-B/16, Last 2 &    1 &       7 &      12 &     12 &     768 &  3072 &   393.3M &    47.21 &   51.98 &   67.94 &    70.93 &    84.71 &   99.09 &    92.37 &   96.40 &   99.57 &     106.95 &   225.78 \\
  V-MoE-B/16, Last 2 &    2 &       7 &      12 &     12 &     768 &  3072 &   393.3M &    48.31 &   54.92 &   68.84 &    71.81 &    85.39 &   99.21 &    92.78 &   96.56 &   99.63 &     130.86 &   250.70 \\
 V-MoE-L/32, Every 2 &    2 &       7 &      24 &     16 &    1024 &  4096 &  3448.2M &    49.31 &   53.61 &   69.21 &    72.02 &    84.81 &   99.18 &    93.02 &   96.32 &   99.33 &     165.51 &   267.10 \\
 V-MoE-B/16, Every 2 &    2 &       7 &      12 &     12 &     768 &  3072 &   979.0M &    49.31 &   55.45 &   69.60 &    72.50 &    85.26 &   99.16 &    92.76 &   96.74 &   99.20 &     201.40 &   303.24 \\
            ViT-L/16 &  --- &      14 &      24 &     16 &    1024 &  4096 &   323.1M &    53.40 &   60.25 &   74.36 &    76.62 &    87.12 &   99.33 &    93.93 &   97.12 &   99.63 &     651.26 &  1572.92 \\
  V-MoE-L/16, Last 2 &    1 &      14 &      24 &     16 &    1024 &  4096 &   843.6M &    55.80 &   60.53 &   75.81 &    78.00 &    87.47 &   99.39 &    94.39 &   97.09 &   99.39 &     698.14 &  1577.40 \\
  V-MoE-L/16, Last 2 &    2 &      14 &      24 &     16 &    1024 &  4096 &   843.6M &    56.76 &   61.46 &   76.53 &    78.64 &    87.54 &   99.29 &    94.19 &   97.37 &   99.58 &     761.27 &  1666.10 \\
 V-MoE-L/16, Every 2 &    2 &      14 &      24 &     16 &    1024 &  4096 &  3446.0M &    57.65 &   62.41 &   77.10 &    79.01 &    87.41 &   99.48 &    94.64 &   97.55 &   99.38 &    1205.99 &  2177.14 \\
            ViT-H/14 &  --- &      14 &      32 &     16 &    1280 &  5120 &   655.8M &    56.68 &   62.34 &   76.95 &    79.02 &    88.08 &   99.50 &    94.71 &   97.11 &   99.71 &    2387.99 &  4276.42 \\
  V-MoE-H/14, Last 5 &    2 &      14 &      32 &     16 &    1280 &  5120 &  2688.6M &    60.12 &   62.95 &   78.08 &    80.10 &    88.23 &   99.53 &    94.86 &   97.17 &   99.67 &    2735.70 &  4750.73 \\
 V-MoE-H/14, Every 2 &    2 &      14 &      32 &     16 &    1280 &  5120 &  7160.8M &    60.62 &   63.38 &   78.21 &    80.33 &    88.36 &   99.58 &    94.91 &   97.45 &   99.68 &    3477.18 &  5795.35 \\
           V-MoE-15B &    2 &      --- &      48 &     16 &    1408 &  6400 & 14705.1M &  --- &   68.66 &   82.78 &    84.29 &    90.35 & --- &  --- & --- & --- &   16775.50 & 33943.30 \\
\bottomrule
\end{tabular}
}}
\end{table}
\end{landscape}
\newpage

\begin{figure}
\centering
\begin{subfigure}{.3\textwidth}
  \centering
  \includegraphics[width=\linewidth]{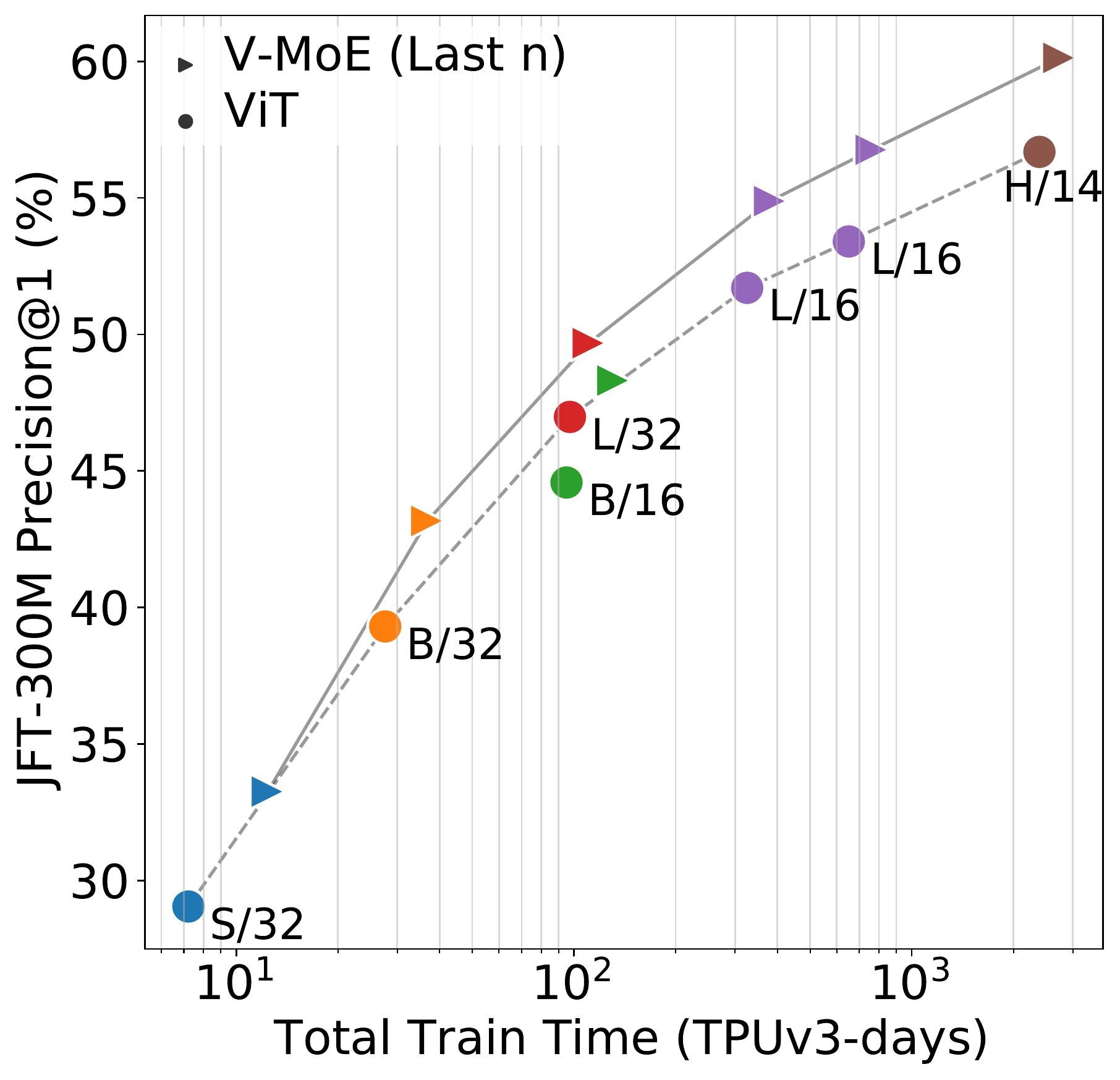}
  \caption{}
  \label{im:upstream_vs_days}
\end{subfigure}%
\begin{subfigure}{.3\textwidth}
  \centering
  \includegraphics[width=\linewidth]{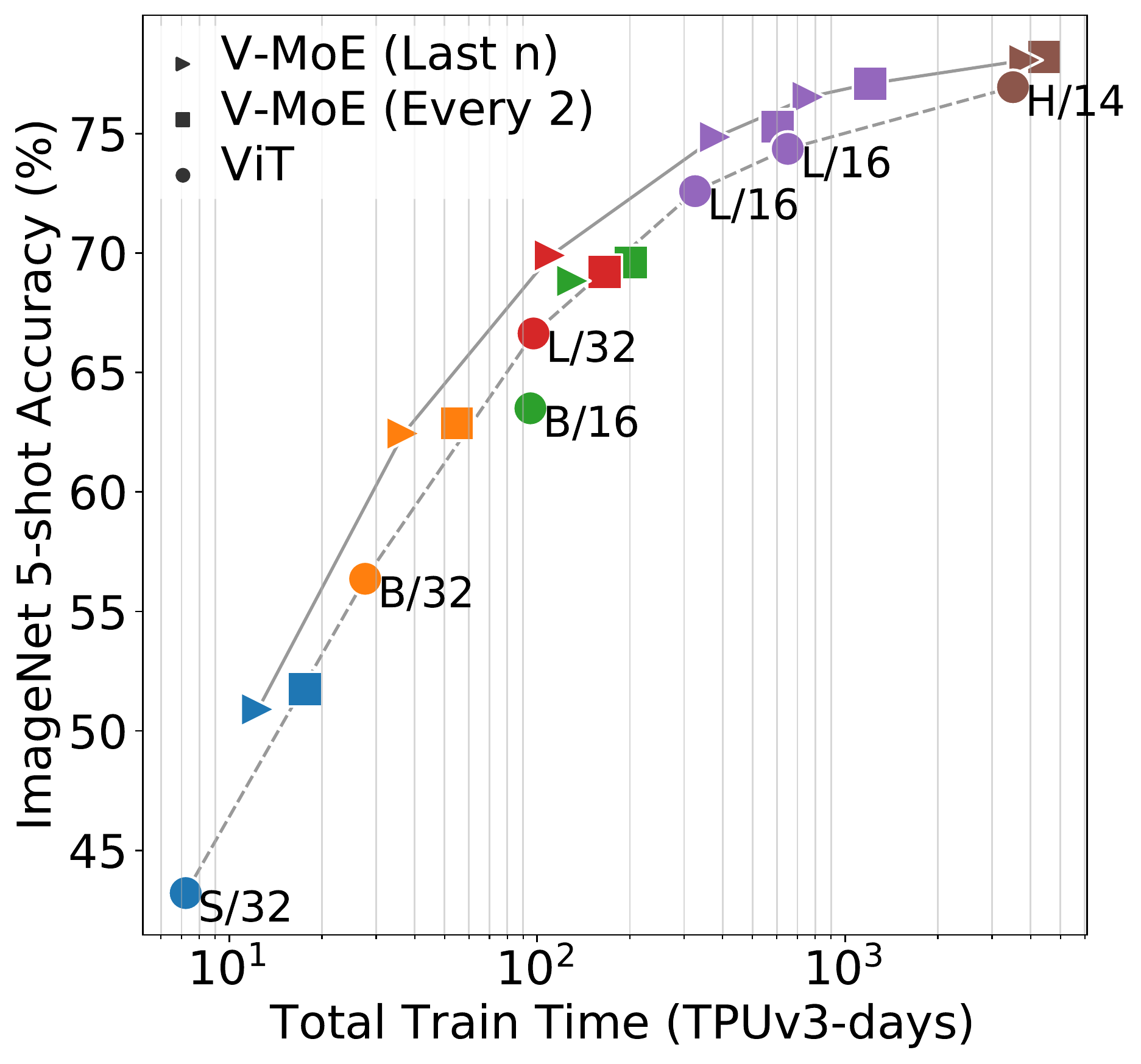}
  \caption{}
  \label{im:imagenet5shot_vs_days}
\end{subfigure}
\begin{subfigure}{.3\textwidth}
  \centering
  \includegraphics[width=\linewidth]{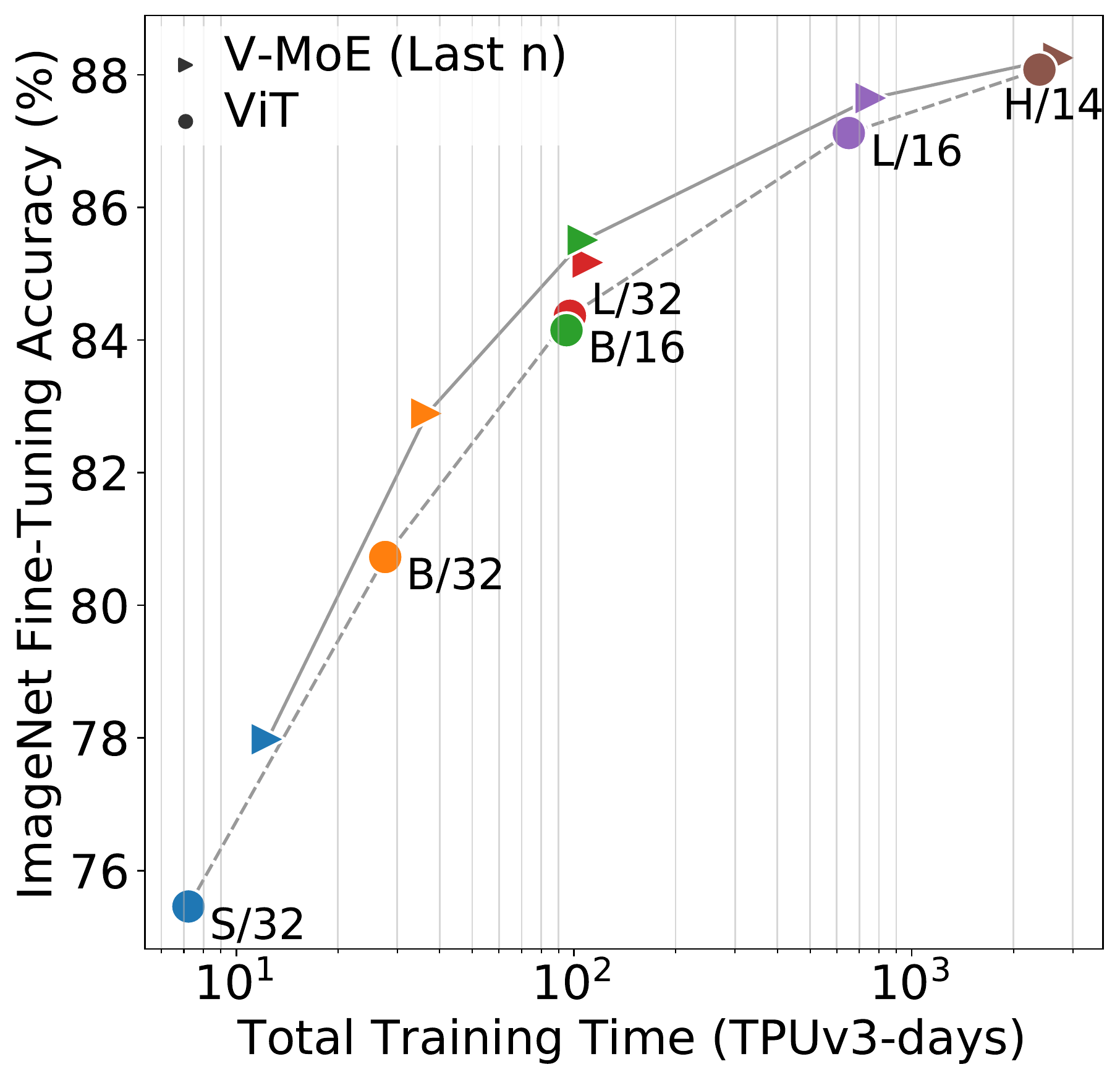}
  \caption{}
  \label{im:imagenet_finetune_vs_days}
\end{subfigure}
\caption{%
Performance on (a) JFT-300M, (b) ImageNet 5-shot and (c) fine-tuning on full ImageNet achieved by different models as a function of the total training time (TPUv3-core-days).
Colors represent different VIT variants, markers represent either standard $\densesym{}$ViT or $\lastsym{}$V-MoEs on the last $n$ even blocks.
The lines represent the Pareto frontier of VIT (dashed) and V-MoE (solid) variants.}
\label{im:performance_vs_days}
\end{figure}

\begin{figure}
\centering
\begin{subfigure}{.50\textwidth}
  \centering
  \includegraphics[width=1.\linewidth]{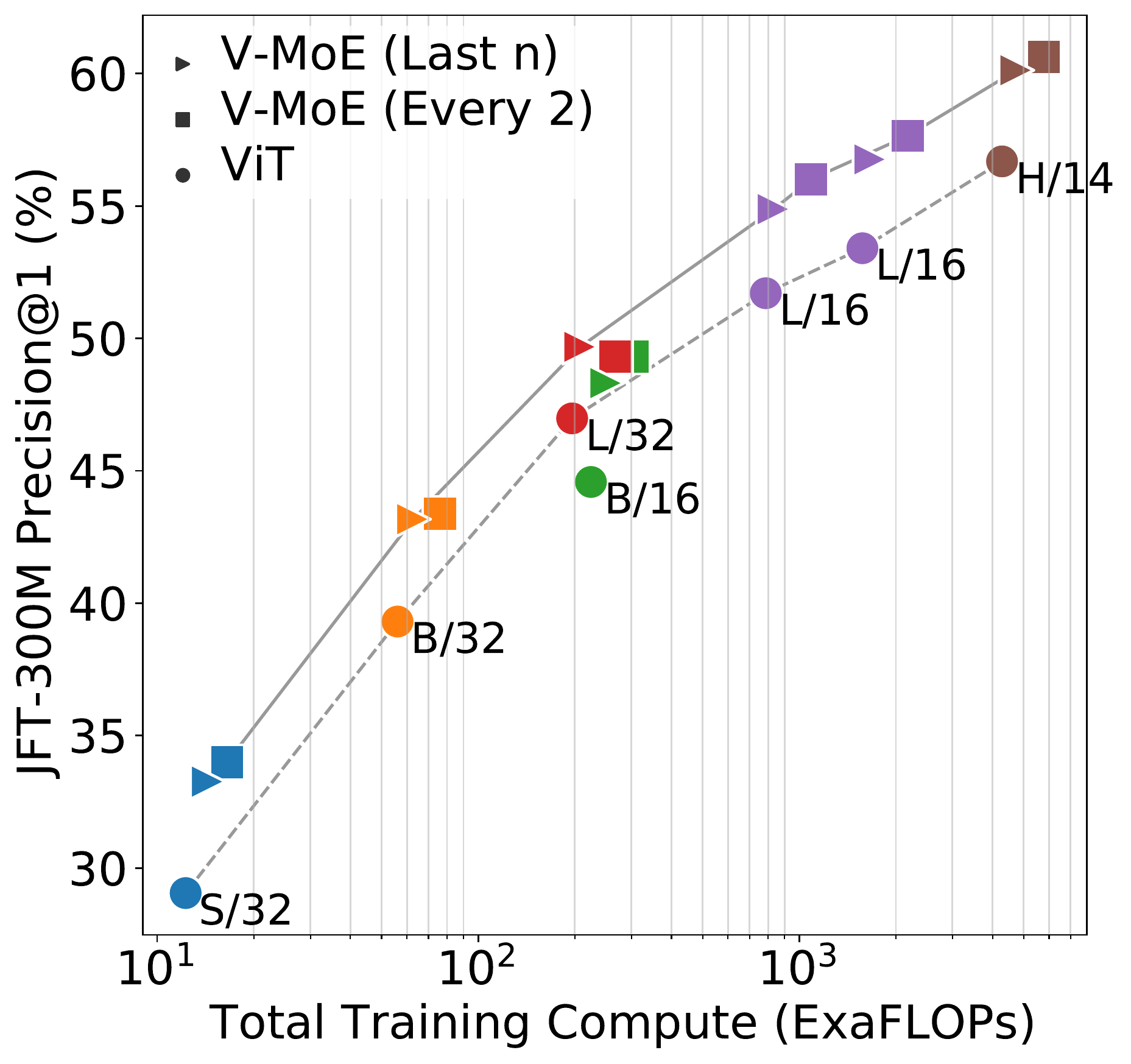}
  \caption{Total training FLOPs.}
  \label{im:upstream_vs_model_flops}
\end{subfigure}%
\begin{subfigure}{.50\textwidth}
  \centering
  \includegraphics[width=1.\linewidth]{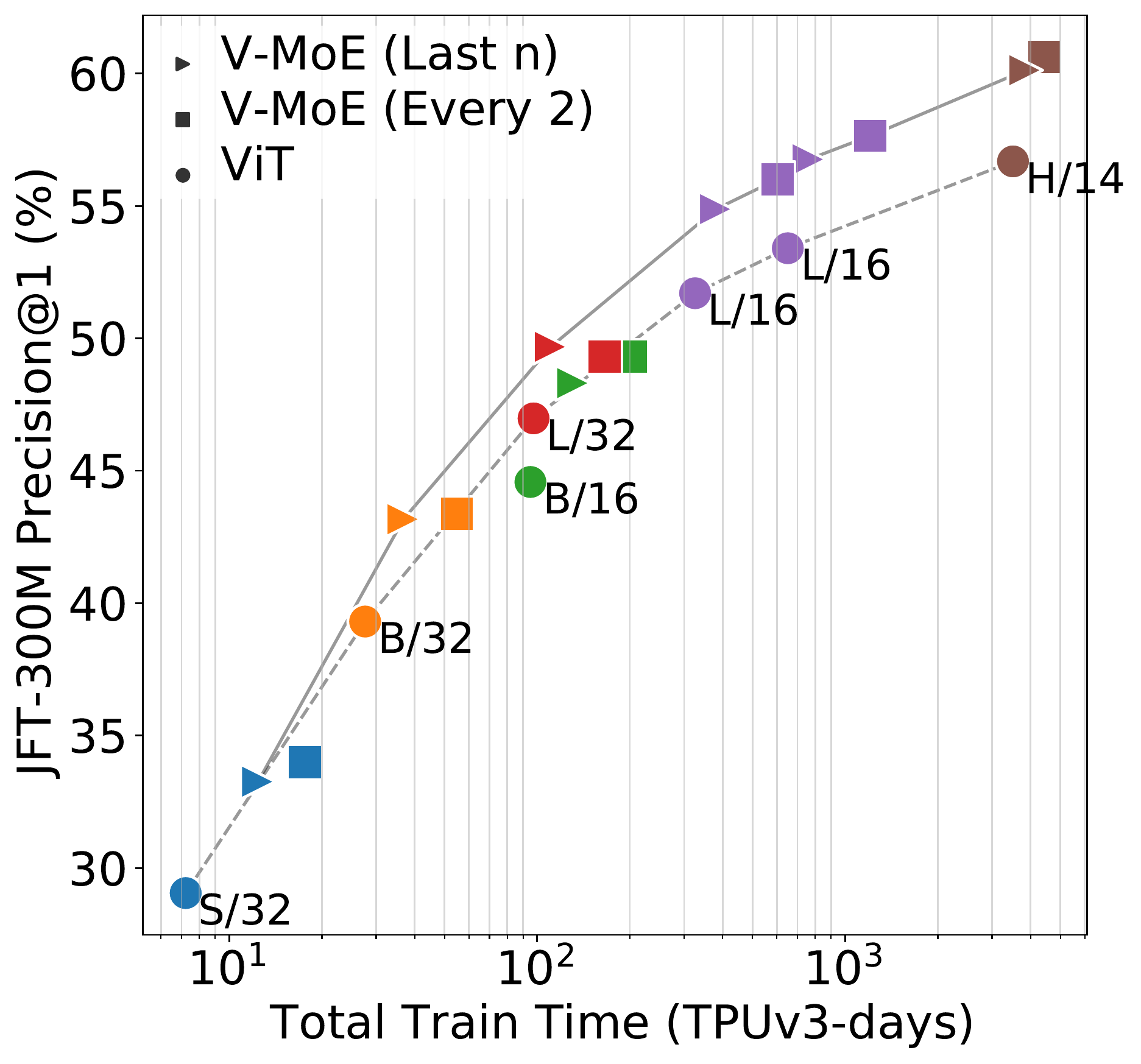}
  \caption{Total training runtime.}
  \label{im:upstream_vs_model_runtime_app}
\end{subfigure}
\caption{Upstream performance of sparse and dense models.
The $x$-axis in (a) shows the total FLOPs required during training, while (b) represents the total training time for identical hardware. 
}
\label{im:upstream_vs_model}
\end{figure}

\begin{figure}[tb]
\centering
\includegraphics[width=1.0\textwidth]{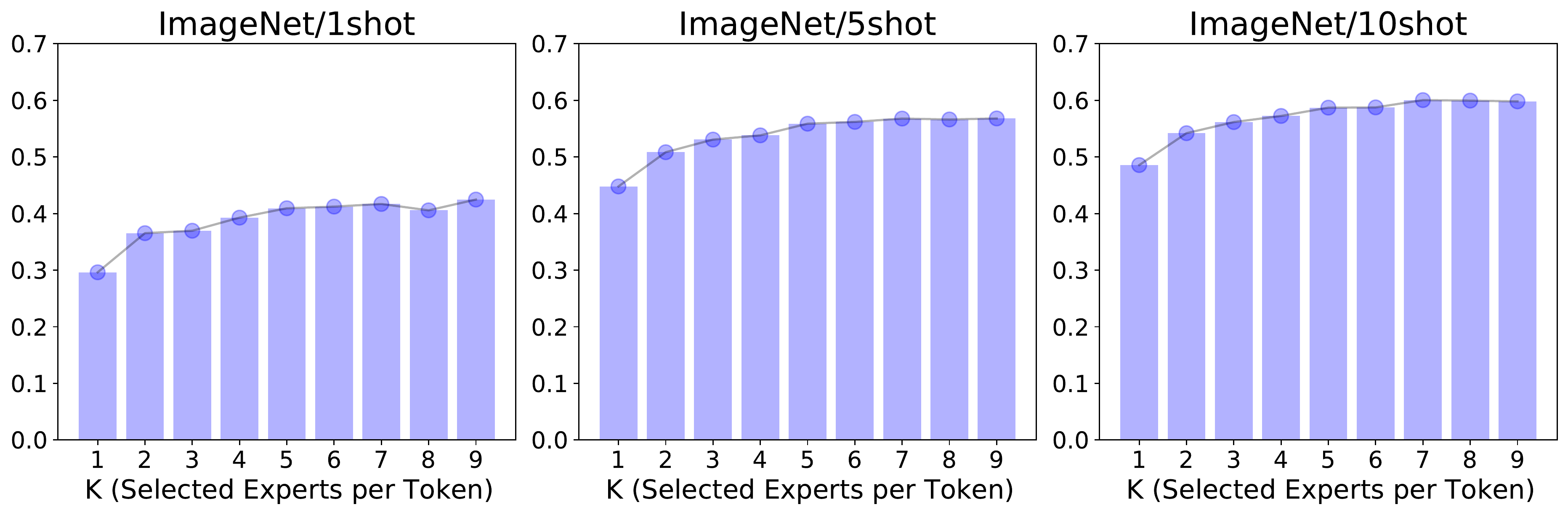}
\caption{Upstream, few-shot and training FLOPs as a function of $k$ for every-2 \abbv{}-S/32.}
\includegraphics[width=0.75\textwidth]{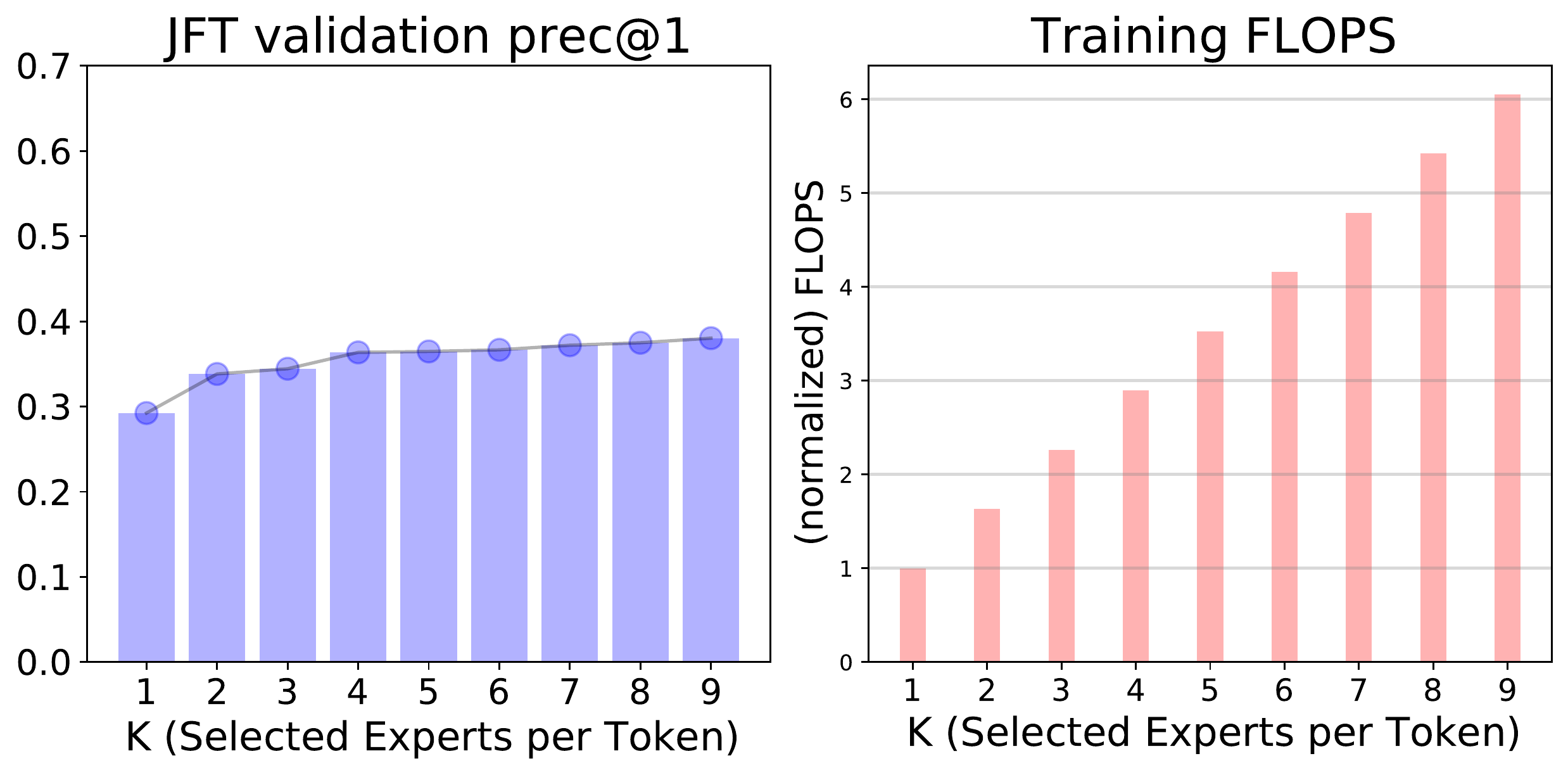}
\label{im:performance_increasing_k}
\end{figure}

\subsection{Computing Precision-at-1 on JFT}\label{sec:prec_at_1_jft}
JFT is multi-label, and it contains a hierarchy of classes.
However, for computing precision at one, we ignore this hierarchy: given predictions on an image, we just look at whether the class with highest predicted probability is indeed one of the true labels for the image.

\begin{figure}
\centering
\begin{subfigure}{.50\textwidth}
  \centering
  \includegraphics[width=1.\linewidth]{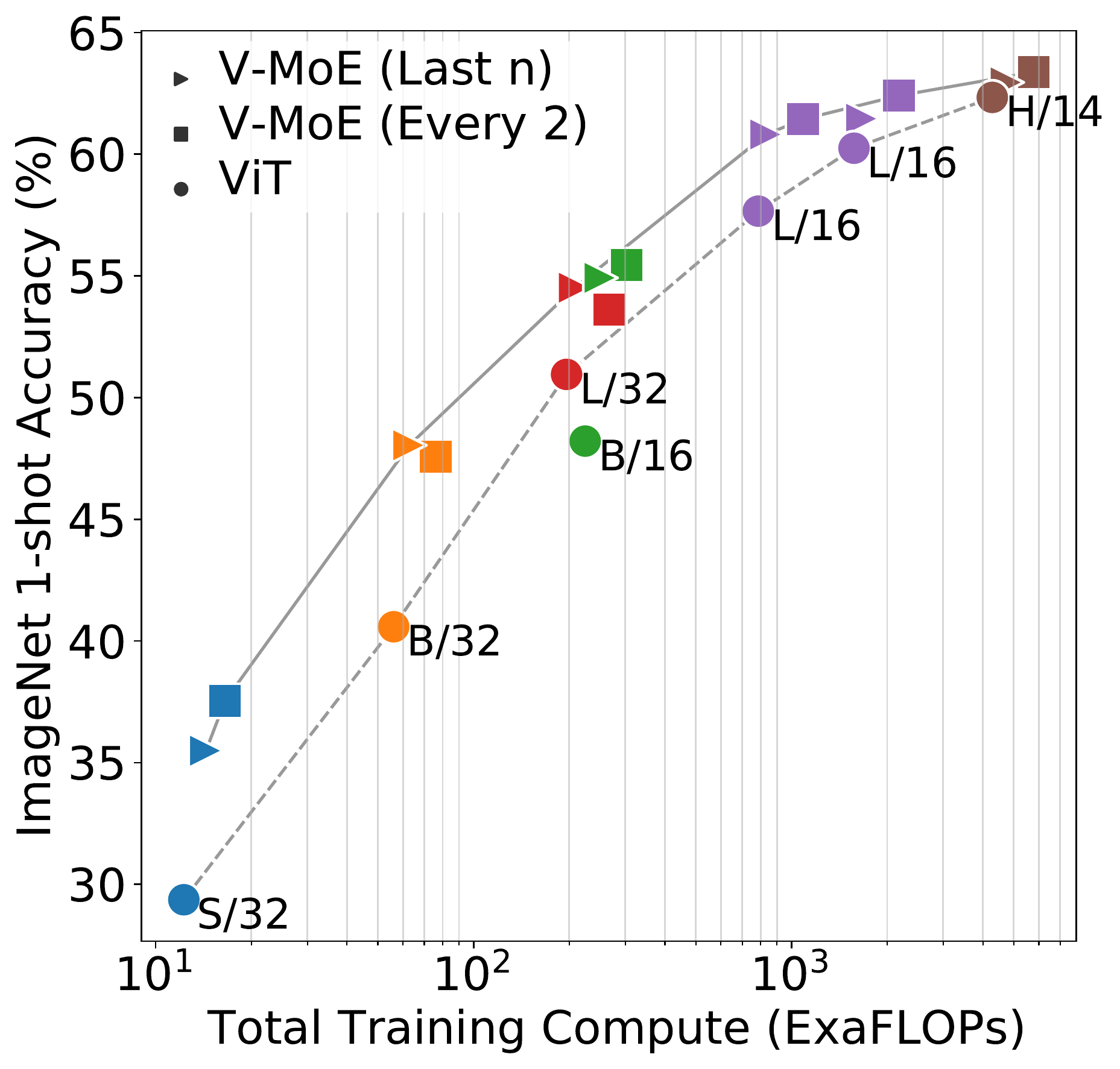}
  \caption{ImageNet 1-shot (total training FLOPs).}
  \label{im:fewshot_1_flops}
\end{subfigure}%
\begin{subfigure}{.50\textwidth}
  \centering
  \includegraphics[width=1.\linewidth]{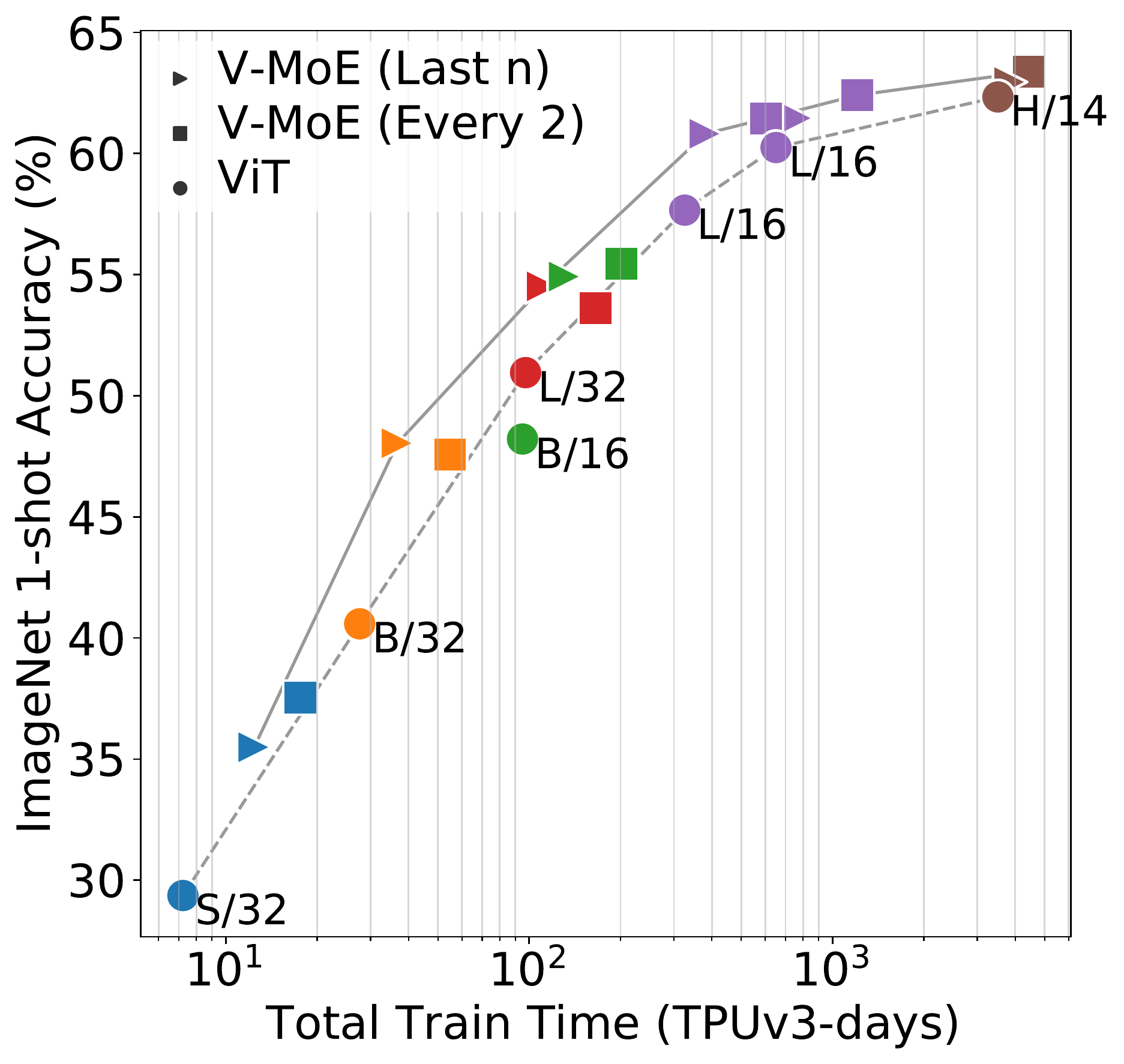}
  \caption{ImageNet 1-shot (total training runtime).}
  \label{im:fewshot_1_runtime}
\end{subfigure}
\caption{
ImageNet/1shot performance of sparse and dense models.
The $x$-axis in (a) shows the total FLOPs required during training, while (b) represents the total training time for identical hardware.
}
\label{im:fewshot_1_flops_and_runtime}
\end{figure}

\begin{figure}
\centering
\begin{subfigure}{.50\textwidth}
  \centering
  \includegraphics[width=1.\linewidth]{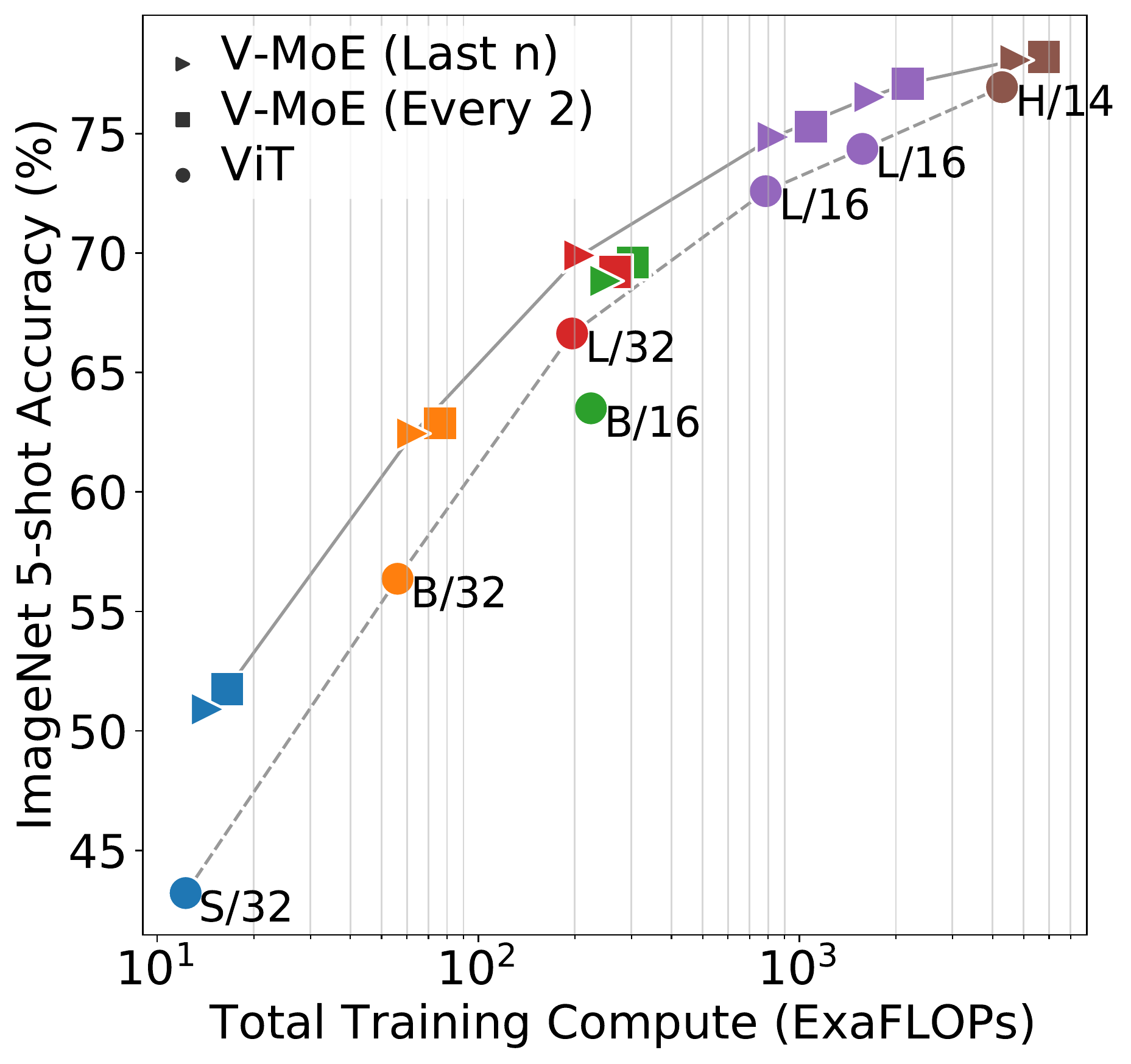}
  \caption{ImageNet 5-shot (total training FLOPs).}
  \label{im:fewshot_5_flops}
\end{subfigure}%
\begin{subfigure}{.50\textwidth}
  \centering
  \includegraphics[width=1.\linewidth]{images/aa_paper/imagenet5shot_train_total_days.pdf}
  \caption{ImageNet 5-shot (total training runtime).}
  \label{im:fewshot_5_runtime}
\end{subfigure}
\caption{
ImageNet/5shot performance of sparse and dense models.
The $x$-axis in (a) shows the total FLOPs required during training, while (b) represents the total training time for identical hardware.
}
\label{im:fewshot_5_flops_and_runtime}
\end{figure}

\begin{figure}
\centering
\begin{subfigure}{.50\textwidth}
  \centering
  \includegraphics[width=1.\linewidth]{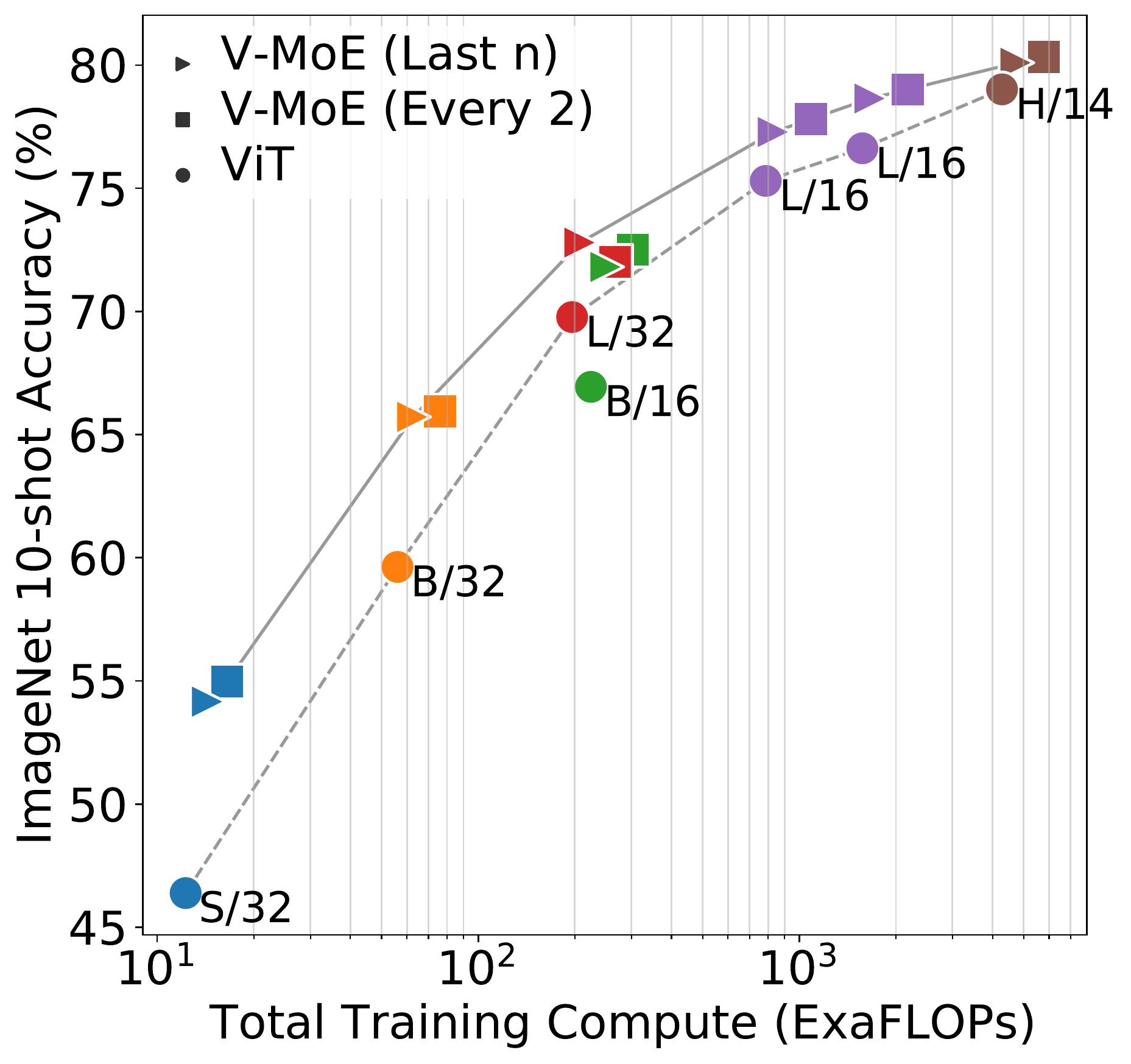}
  \caption{ImageNet 10-shot (total training FLOPs).}
  \label{im:fewshot_10_flops}
\end{subfigure}%
\begin{subfigure}{.50\textwidth}
  \centering
  \includegraphics[width=1.\linewidth]{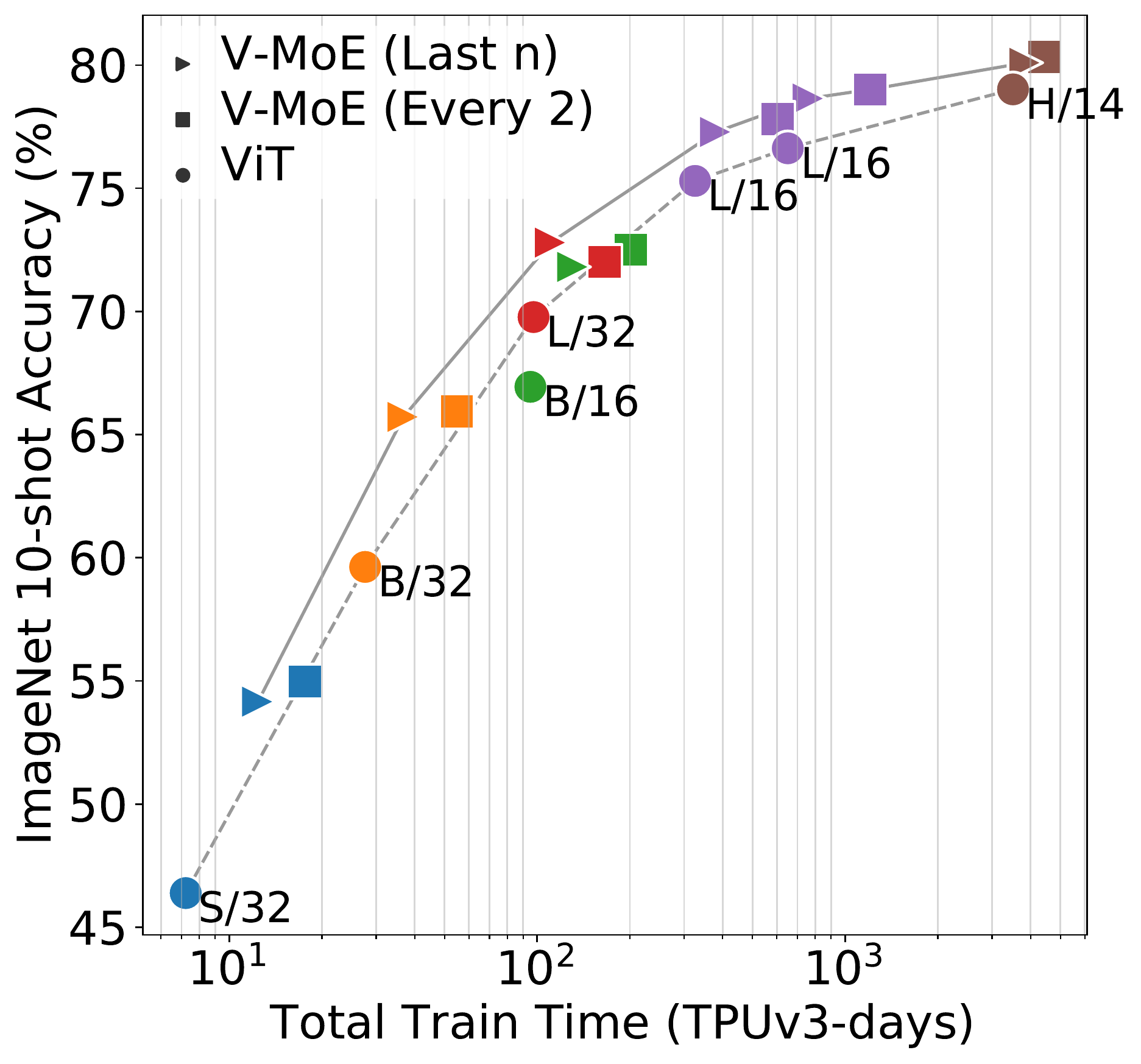}
  \caption{ImageNet 10-shot (total training runtime).}
  \label{im:fewshot_10_runtime}
\end{subfigure}
\caption{
ImageNet/10shot performance of sparse and dense models.
The $x$-axis in (a) shows the total FLOPs required during training, while (b) represents the total training time for identical hardware.
}
\label{im:fewshot_10_flops_and_runtime}
\end{figure}

\begin{figure}
\centering
\includegraphics[width=\textwidth]{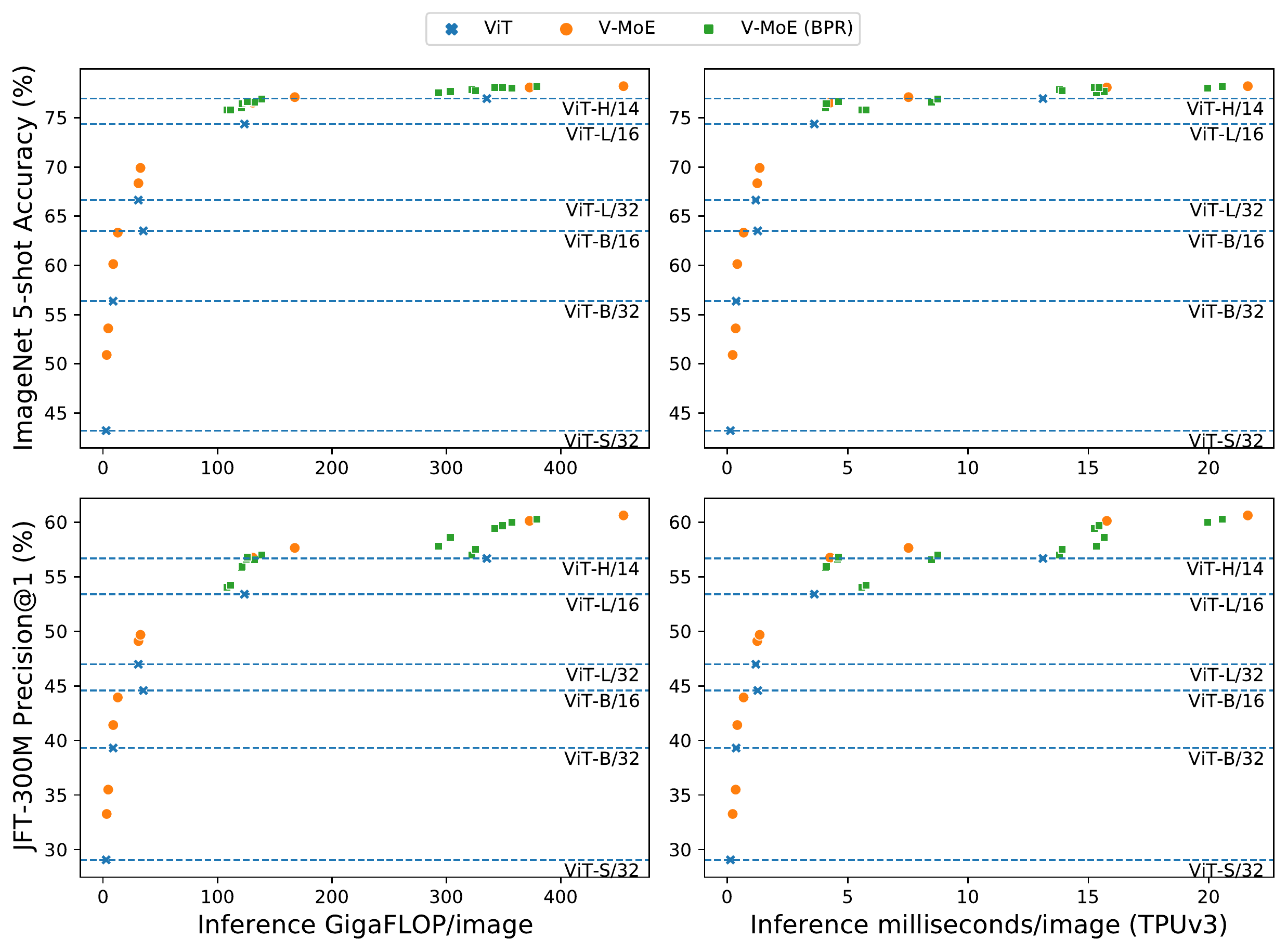}
\caption{%
\textbf{Reducing compute with priority routing.} Performance vs.\ \emph{inference} FLOPs and runtime for all models.
\abbv{}s with the original vanilla routing are represented by $\bullet$, while $\blacksquare$ shows \abbv{}s where BPR and a mix of ${C \in \{0.6, 0.7, 0.8\}}$ and ${k \in \{1, 2\}}$ are used to reduce compute. 
ViT models shown as $\mathbf{x}$.
See \cref{im:upstream_5shot_at_inference} for a zoomed-in version on the largest models (versus inference FLOPs).%
\label{im:upstream_5shot_at_inference_all}%
}%
\label{im:inference_vs_runtime}
\end{figure}

\subsection{Training data deduplication}\label{sec:dedup_models}

\cref{tab:dedup_models} shows the effect of Imagenet deduplication on the training data for fewshot with \abbv{}-S/32.
Overall, we do not observe a consistent and significant effect after de-duplicating the data.
The variance across seeds is notable and---except in the case of IN/1shot---de-duplicated models can outperform (and underperform) the original ones on few-shot evaluation.

\begin{table}[btp]
\caption{\label{tab:dedup_models} \textbf{Effect of ImageNet deduplication on the training data for fewshot with \abbv{}-S/32.}
In order to test the effect of removing some images in the training set that are ``close'' to some ImageNet ones, we trained three \abbv{}-S/32 models ---with different seeds--- on the de-duplicated dataset, and compare their few-shot performance as shown below.
The variance in the results is considerable.
The original model dominates on 1-shot, while two out of the three seeds outperform the original model on 5-, 10-, and 25-shot.
The de-duplicated dataset contained more images overall, but we limited the training set to the original size (around 305M) and trained for the same epochs.
}
\resizebox{\textwidth}{!}{\addtolength{\tabcolsep}{-2pt}
\begin{tabular}{@{}lrrrrrrr@{}} 
\toprule
 Model & Dedup & Seed & IN/1shot & IN/5shot & IN/10shot & IN/25shot \\
\midrule
\abbv{}-S/32 & No & 0 & \textbf{37.53} & 51.75 & 54.97 & 57.44 \\
\midrule
\abbv{}-S/32 & Yes & 0 & 34.07 & 49.34 & 52.21 & 55.11 \\
\abbv{}-S/32 & Yes & 1 & 35.63 & 51.95 & 55.79 & 58.19 \\
\abbv{}-S/32 & Yes & 2 & 36.72 & \textbf{53.09} & \textbf{56.50} & \textbf{58.84} \\
\bottomrule
\end{tabular}%
}
\end{table} 
\clearpage
\section{\maxrouting{}}
\label{app_skip_patch_training}

\newcommand{\po}{\phantom{1}}
\begin{table}
\centering
\begin{tabular}{@{}ccccccc@{}} 
\toprule
 Model & Experts & Routing & JFT prec@1 & INet/5shot & Time[\%] & FLOPs[\%] \\
 \midrule
VIT-H/14 & - & -             & 56.68 & 76.95 & 100.00 & 100.00 \\
VIT-L/16 & - & -             & 53.40 & 74.36 & \po27.58 & \po36.83 \\
\abbv{}-L/16 & Last-2 & Vanilla & 56.76 & 76.53 & \po32.56 & \po39.02 \\
\abbv{}-L/16 & Every-2 & Vanilla & 57.64 & 77.10 & \po57.40 & \po49.95 \\
\abbv{}-H/14 & Last-5 & Vanilla & 60.12 & 78.08 & 120.22 & 111.12 \\
\abbv{}-H/14 & Every-2 & Vanilla & 60.62 & 78.21 & 164.89 & 135.59 \\
\bottomrule
\end{tabular}%
\vspace*{3mm}
\caption{\label{tab:matching_performance_max_routing} Time and FLOPs \emph{unmatched} inference results for JFT prec@1 and ImageNet 5shot.}
\end{table}

\begin{table}
\resizebox{\textwidth}{!}{%
\begin{tabular}{@{}cccccccc@{}} 
\toprule
 Model & Experts & At Inference & C & JFT prec@1 & INet/5shot &Time[\%] & FLOPs[\%] \\
 \midrule
VIT-H/14 & - & - & - & 56.68 & 76.95 & 100.00 & 100.00 \\
\abbv{}-H/14 & Last-5 & k=2 $\to$ k=1 & 1.05 & 58.60 & 77.87 & 111.57 & 100.26 \\
\abbv{}-H/14 & Last-5 & k=2 $\to$ k=1 & 1.25 & 59.21 & 77.59 & 113.67 & 102.53 \\
\abbv{}-H/14 & Last-5 & k=2 & 0.5\po & 58.61 & 77.92 & 118.14 & 100.02 \\
\abbv{}-H/14 & Last-5 & k=2 & 0.6\po & 59.42 & 78.05 & 121.68 & 102.30 \\
\abbv{}-H/14 & Every-2 & k=2 $\to$ k=1 & 1.05 & 59.46 & 77.82 & 134.87 & 100.07 \\
\abbv{}-H/14 & Every-2 & k=2 & 0.5\po & 59.44 & 77.70 & 155.83 & 100.03 \\
\bottomrule
\end{tabular}%
} %
\vspace*{3mm}
\caption{\label{tab:matching_cost_max_routing} FLOPs \emph{matched} inference results with \maxrouting{}, lower C, and  reduced $k$.}
\end{table}

\subsection{The Routing Algorithms}

\begin{algorithm}[H]
\SetAlgoLined
\KwResult{complete assignment of patches to experts (with some potential dropping)}
 initialize empty buffers with capacity $B_e$ for all experts $e$ (see \cref{sec:model})\;
 \For{$i = 1, \dots, k$}{
  \For{patch $p = 1, \dots, N$}{
      $e, w = \text{Router}(\text{TOP}-i \text{ position}, \text{patch } p)$\;
      \eIf{$e$ is not full}{
       add patch $p$ to processing buffer of expert $e$ with weight $w$\;
       }{
       skip $i$-th expert assignment for patch $p$\;
      }
    }
 }
 \caption{Vanilla Routing Allocation}
 \label{algo:default_patch_assignment}
\end{algorithm}

\begin{algorithm}[H]
\SetAlgoLined
\KwResult{complete assignment of patches to experts (with some potential dropping)}
 initialize empty buffers with capacity $B_e$ for all experts $e$ (see \cref{sec:model})\;
 \For{patch $p = 1, \dots, N$}{
    $s(p) = \mathrm{ComputeScore}(\mathrm{patch} \ p, \ \mathrm{Router}(\cdot))$\;
 }
 $ \mathrm{patch} \ \mathrm{ordering} \ \bar{p} = \mathrm{SortPatches}(\mathrm{scores} \ s, \mathrm{decreasing = True}) $\;
 \For{$i = 1, \dots, k$}{
  \For{patch $p = (1), \dots, (N)$ according to $\bar{p}$}{
      $e, w = \mathrm{Router}(\mathrm{TOP}-i \ \mathrm{ position}, \ \mathrm{patch} \ p)$\;
      \eIf{$e$ is not full}{
       add patch $p$ to processing buffer of expert $e$ with weight $w$\;
       }{
       skip $i$-th expert assignment for patch $p$\;
      }
    }
 }
 \caption{\maxrouting{} Allocation}
 \label{algo:max_weight_patch_assignment}
\end{algorithm}
We explored a few scoring functions, and concluded that sorting according to the maximum routing weight for each patch $p$ works really well---formally, $s(p) = \max_e w_{e, p}$, where $w_{e, p}$ is the output of the routing function $g$ for patch $p$ and expert $e$ (see \Cref{sec:max_routing}).
We experimented with the sum of all the TOP-$k$ weights too (rather than just the TOP-1), leading to similar results.
Moreover, we tried to directly learn a scoring function.
In this case, the router would output $E$ weights per patch (one per expert, jointly normalized by a softmax function) together with the score $s(p)$ ---one per patch.
We explored a couple of scoring functions (linear + sigmoid, etc), to conclude that the maximum routing weight is quite a good baseline and hard to beat.

A natural extension of this algorithm consists in sorting at the patch-expert assignment level, rather than at the global patch level.
The main difference with \cref{algo:max_weight_patch_assignment} is that the sorting then looks at (patch $p$, TOP$-i$ expert for $p$) scores for $1 \le i \le k$.
For example, assume $k=2$ and we have two patches, $p_1$ and $p_2$.
Suppose $p_1$ selects experts $(e_{11}, e_{12})$ with routing weights $(0.7, 0.2)$, while $p_2$ selects $(e_{21}, e_{22})$ with weights $(0.5, 0.4)$.
Under \cref{algo:max_weight_patch_assignment} the order in which patch-expert assignments would be attempted is: $(p_1, e_{11}), (p_2, e_{21}), (p_1, e_{12}), (p_2, e_{22})$.
If we use sorting at the patch-expert level, however, we would end up with: $(p_1, e_{11}), (p_2, e_{21}), (p_2, e_{22}), (p_1, e_{12})$.
The latter could make more sense as the second assignment for $p_2$ could be more relevant than the second assignment for $p_1$ given their weights.
We have not empirically tried this approach, however.

For completeness, we also report another related algorithm we did actually experiment with.
We call it \emph{skip-patch}.
In this case, we first set a hyper-parameter $S \in (0, 1)$.
We will process a fraction $S$ of the patches, and directly \textbf{skip} the remaining $1-S$ fraction.
As before, we rank the $N$ patches according to some scoring function $s(\cdot)$.
Then, we directly discard the bottom $(1-S)$\% of the patches, and proceed like in \cref{algo:max_weight_patch_assignment} over the selected $M=SN$ patches.
\Cref{algo:skip_patch_assignment} formally describes the idea.
Going back to our previous example with two patches, if we set $S=0.5$ there, we will discard $p_2$ altogether, and just process: $(p_1, e_{11}), (p_1, e_{12})$.
Note that $S$ and $C$ are two different parameters, and it makes sense to adjust $C$ given $S$ to avoid an excessive FLOPs waste.

\begin{algorithm}[H]
\SetAlgoLined
\KwResult{complete assignment of patches to experts (with some \textbf{enforced} dropping)}
 let $S \in (0, 1)$\;
 initialize empty buffers with capacity $B_e$ for all experts $e$ (see \cref{sec:model})\;
 \For{patch $p = 1, \dots, N$}{
    $s(p) = \mathrm{ComputeScore}(\mathrm{patch} \ p, \ \mathrm{Router}(\cdot))$\;
 }
 $ \mathrm{patch} \ \mathrm{ordering} \ \bar{p} = \mathrm{SortPatches}(\mathrm{scores} \ s, \mathrm{decreasing = True}) $\;
 $ \mathrm{patch} \ \mathrm{ordering} \ \hat{p} = \mathrm{KeepPatches}(\mathrm{TOP-M}, M = SN, \bar{p}) $\;
 \For{$i = 1, \dots, k$}{
  \For{patch $p = (1), \dots, (M)$ according to $\hat{p}$}{
      $e, w = \mathrm{Router}(\mathrm{TOP}-i \ \mathrm{ position}, \ \mathrm{patch} \ p)$\;
      \eIf{$e$ is not full}{
       add patch $p$ to processing buffer of expert $e$ with weight $w$\;
       }{
       skip $i$-th expert assignment for patch $p$\;
      }
    }
 }
 \caption{Skip-Patch Routing Allocation}
 \label{algo:skip_patch_assignment}
\end{algorithm}

\subsection{Applied during Inference}

An appealing property of the algorithms introduced in the previous section is that they are agnostic to how the model was originally trained.
Indeed, we first show the effect of reducing compute at inference time by using \maxrouting{}, \cref{algo:max_weight_patch_assignment}, on models trained using \cref{algo:default_patch_assignment}.
Note the model parameters are identical in both cases, including the router parameters --we are only applying the model at inference, no further learning is involved--, but we apply different routing strategies.
Overall, we observe that discarding patches at random (as \cref{algo:default_patch_assignment} effectively does) leads to a steep loss of performance when we only keep a small percentage of the patches, as one could expect.
On the other hand, if we process the ``right'' patches ---via \cref{algo:max_weight_patch_assignment}--- the performance is surprisingly robust as long as we keep up to around $20$\% of the patches.

\begin{figure}[h]
\centering
\includegraphics[width=1.0\textwidth]{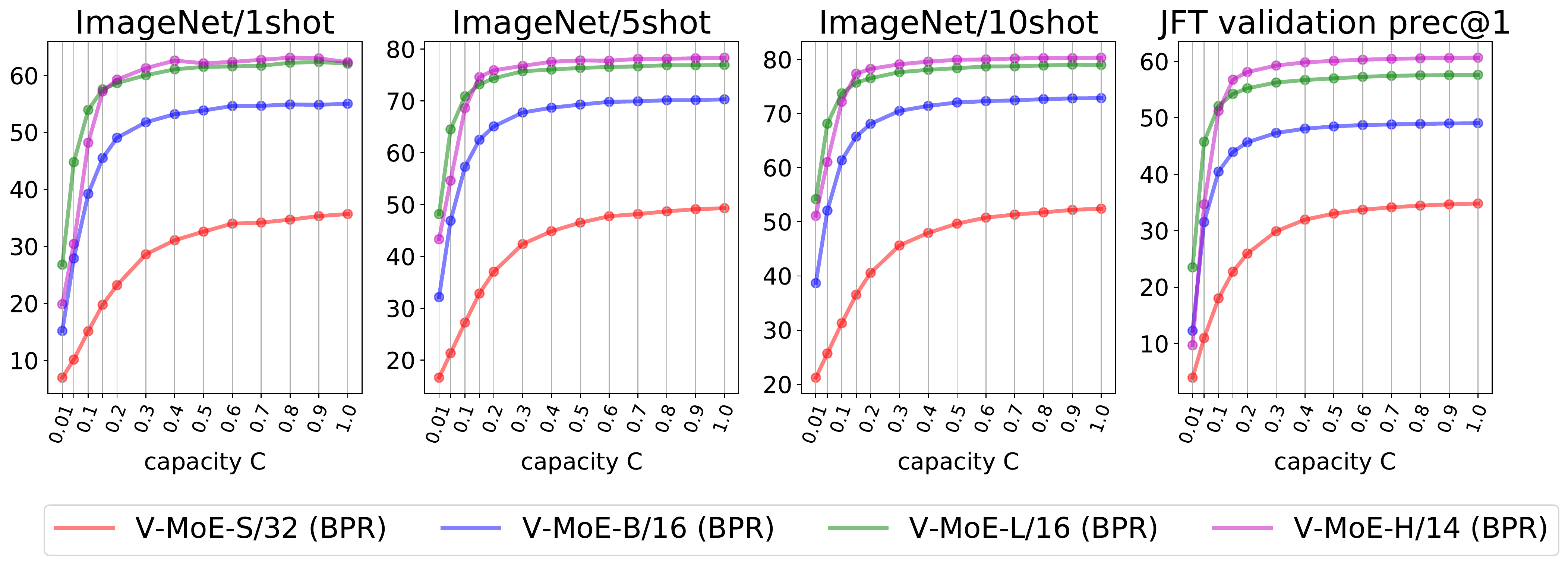}
\caption{Inference performance for various every-2 V-MoE models with $k=2$ for different capacities.
We show \maxrouting{}.}
\label{im:inference_c_vit_all}
\end{figure}

\Cref{im:inference_c_vit_all} shows the inference performance as a function of $C$ for the main every-2 expert models with $k=2$, under \cref{algo:max_weight_patch_assignment}.
We observe performance decreases slowly and smoothly as we constrain more and more the amount of patches experts can process.

\begin{figure}[h]
\centering
\includegraphics[width=1.0\textwidth]{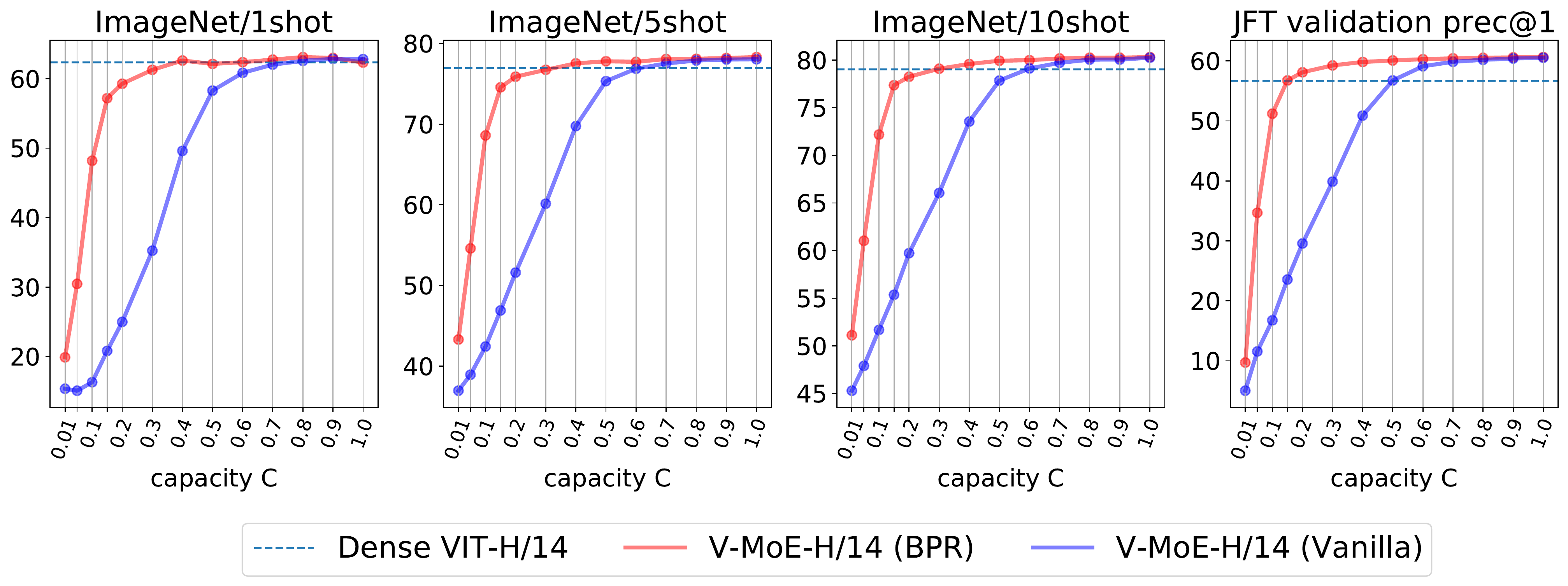}
\caption{Inference performance for every-2 V-MoE-H/14 model with $k=2$ for different capacities. \\
We show \maxrouting{} versus vanilla routing.}
\label{im:inference_c_vit_h}
\end{figure}

\begin{figure}[h]
\centering
\includegraphics[width=1.0\textwidth]{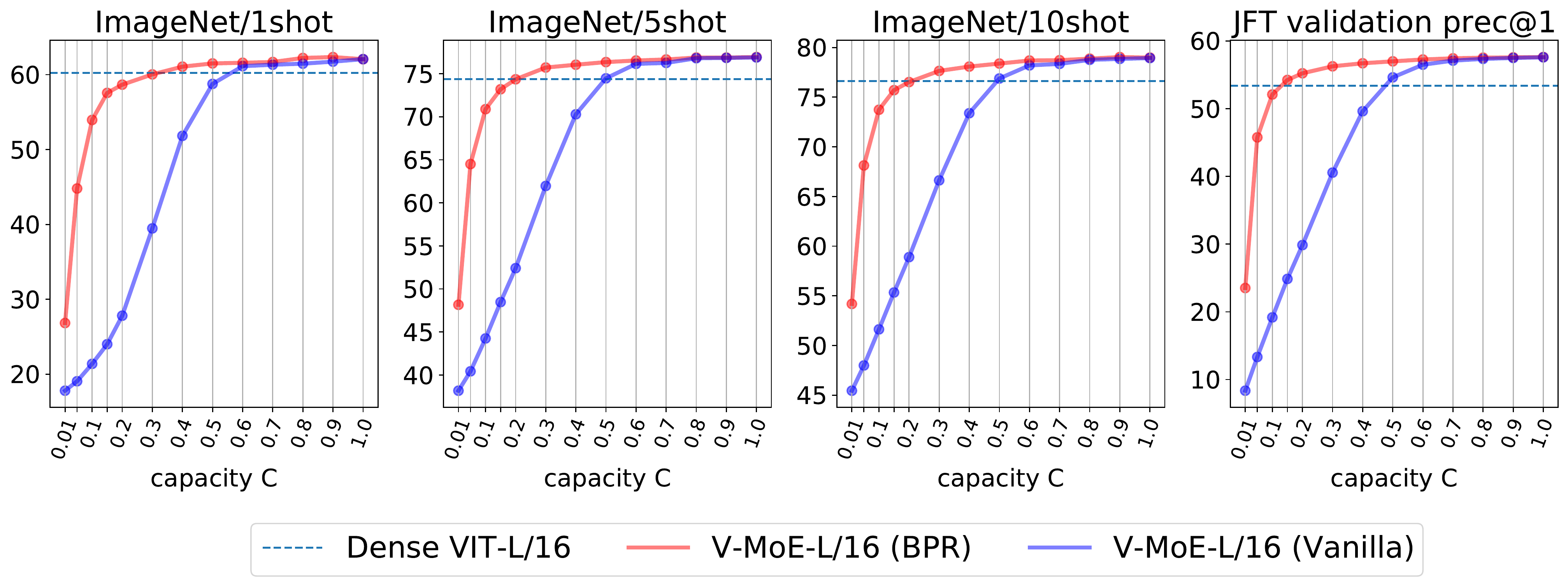}
\caption{Inference performance for every-2 V-MoE-L/16 model with $k=2$ for different capacities. \\
We show \maxrouting{} versus vanilla routing.}
\label{im:inference_c_vit_l}
\end{figure}

\begin{figure}[h]
\centering
\includegraphics[width=1.0\textwidth]{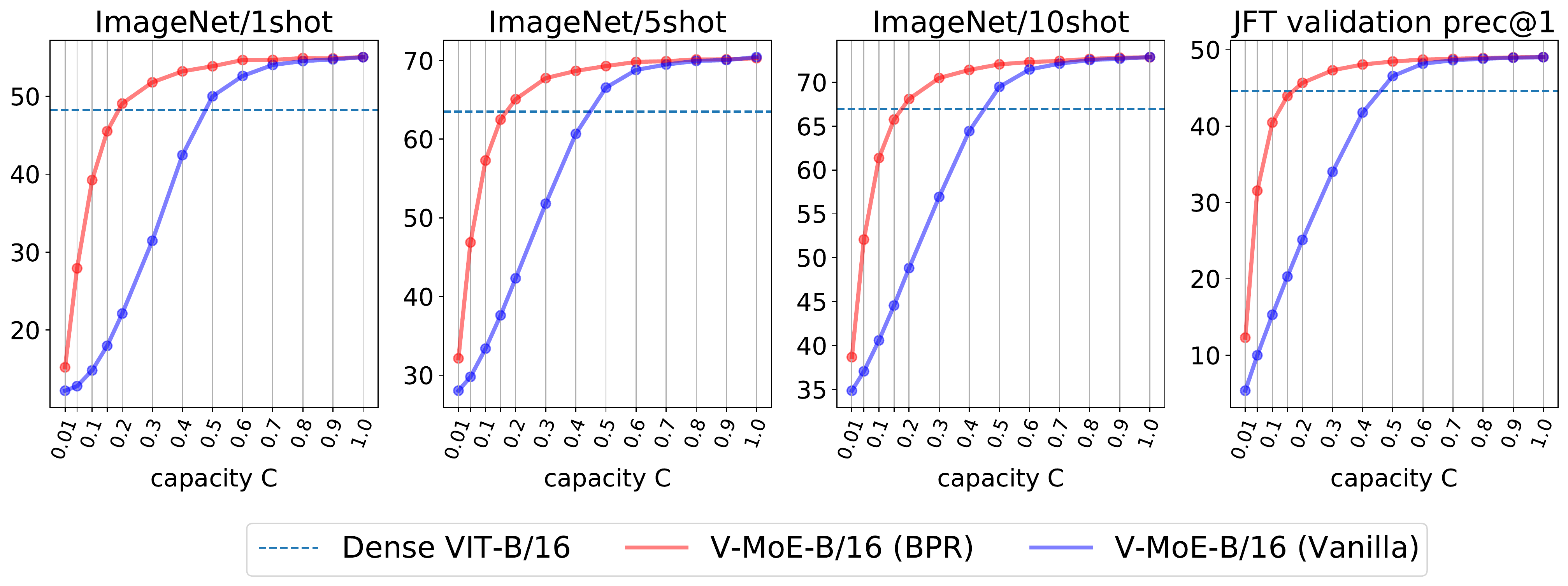}
\caption{Inference performance for every-2 V-MoE-B/16 model with $k=2$ for different capacities. \\
We show \maxrouting{} versus vanilla routing.}
\label{im:inference_c_vit_b}
\end{figure}

\begin{figure}[h]
\centering
\includegraphics[width=1.0\textwidth]{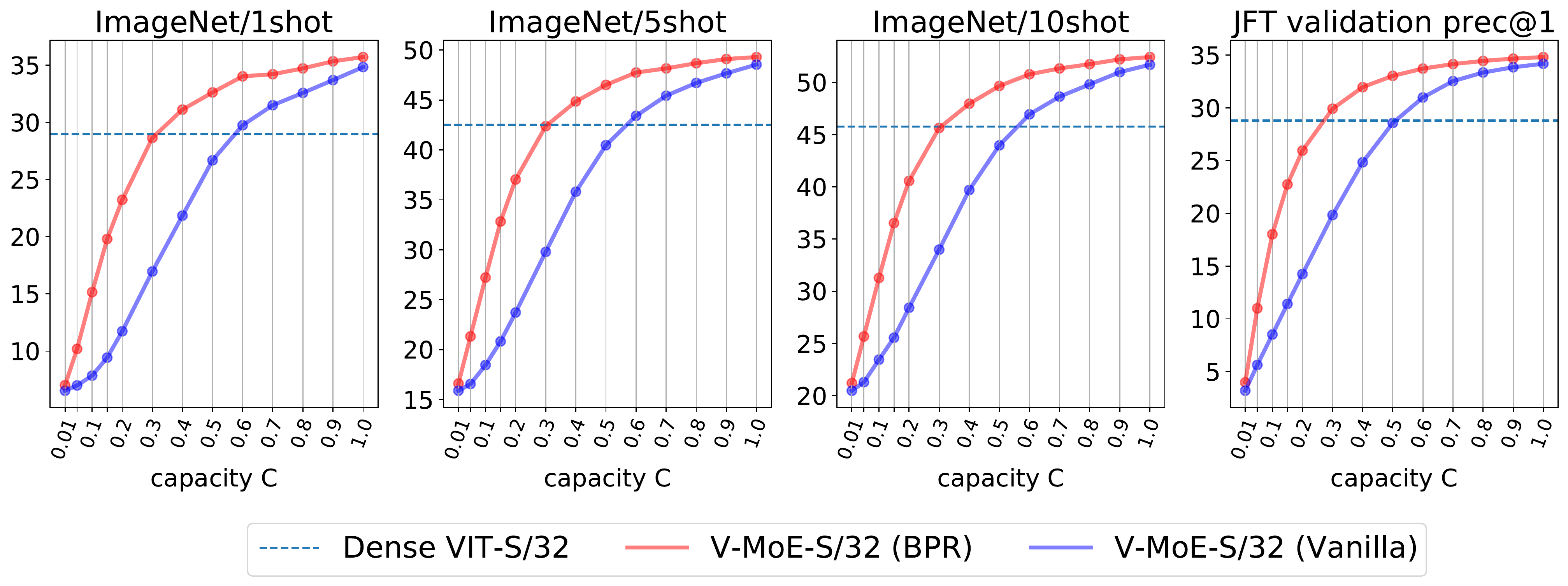}
\caption{Inference performance for every-2 V-MoE-S/32 model with $k=2$ for different capacities. \\
We show \maxrouting{} versus vanilla routing.}
\label{im:inference_c_vit_s}
\end{figure}

Next we compare the inference performance of \cref{algo:default_patch_assignment,algo:max_weight_patch_assignment}.
Results for \abbv{}-H/14 are presented in \cref{im:inference_c_vit_h}, \abbv{}-L/16 in \cref{im:inference_c_vit_l}, \abbv{}-B/16 in \cref{im:inference_c_vit_b}, and \abbv{}-S/32 in \cref{im:inference_c_vit_s}.
In all cases we see the same clear trend.
By definition of \cref{algo:default_patch_assignment,algo:max_weight_patch_assignment}, when $k=2$, if $C \ge 0.5$, then every patch has a decent change of getting its TOP-1 expert processed if routing is balanced.
Therefore, the most interesting regime here is $C < 0.5$.
In that case, we see an enormous gap in performance between \cref{algo:max_weight_patch_assignment,algo:default_patch_assignment}, showing that choosing the right patches really pays off.
Moreover, in most cases, using 15\% of the patches ($C = 0.15$) is enough to match the upstream performance of the dense model.
For the few-shot representations, between 20\% and 30\% of the patches is usually enough.

Overall, we consider the flexibility provided by \cref{algo:max_weight_patch_assignment} to be quite a remarkable property of expert models.
Once trained, they allow for a smooth trade-off between performance and compute, with no further training or adjustment needed.
This can be certainly useful in a practical setting where the use-case may determine the available resources and constraints at hand.

\clearpage
\subsection{Applied during Training}

The previous subsection explored applying priority routing during inference to a pre-trained model.
A natural extension consist in directly training a model with \cref{algo:max_weight_patch_assignment} from scratch.
By forcing experts to work with a small buffer or capacity ratio (i.e. $C << 1$), we can save substantial training FLOPs while hopefully still get decent performance improvements with respect to dense models.

We show results for three models: \abbv{}-S/32, \abbv{}-B/32, and \abbv{}-L/32.
For completeness, we compare \cref{algo:default_patch_assignment,algo:max_weight_patch_assignment}.
In all cases we see strong improvements when training with \cref{algo:max_weight_patch_assignment}.
When we use full capacity ($C \ge 1.0$), however, we expect both algorithms to behave in a fairly similar fashion, as no dropping is needed as long as routing is reasonably balanced.

\begin{figure}[h]
\centering
\includegraphics[width=1.0\textwidth]{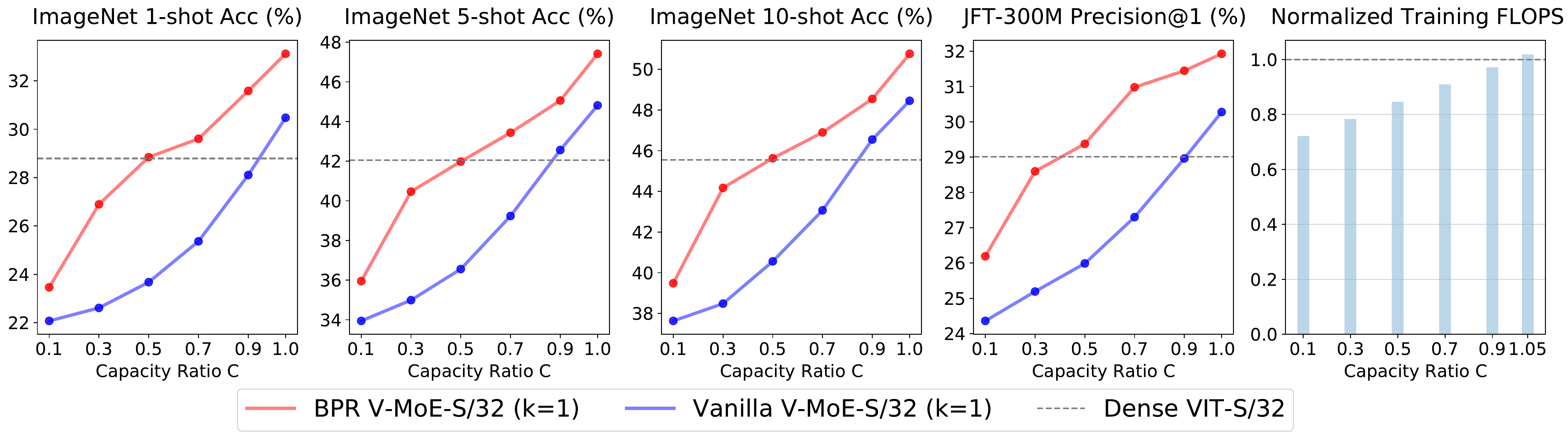}
\caption{Training with \maxrouting{}. Model: \abbv{}-S/32, $k = 1$. Mean over 4 seeds.}
\label{im:training_max_routing_vit_s_k1}
\end{figure}

\begin{figure}[h]
\centering
\includegraphics[width=1.0\textwidth]{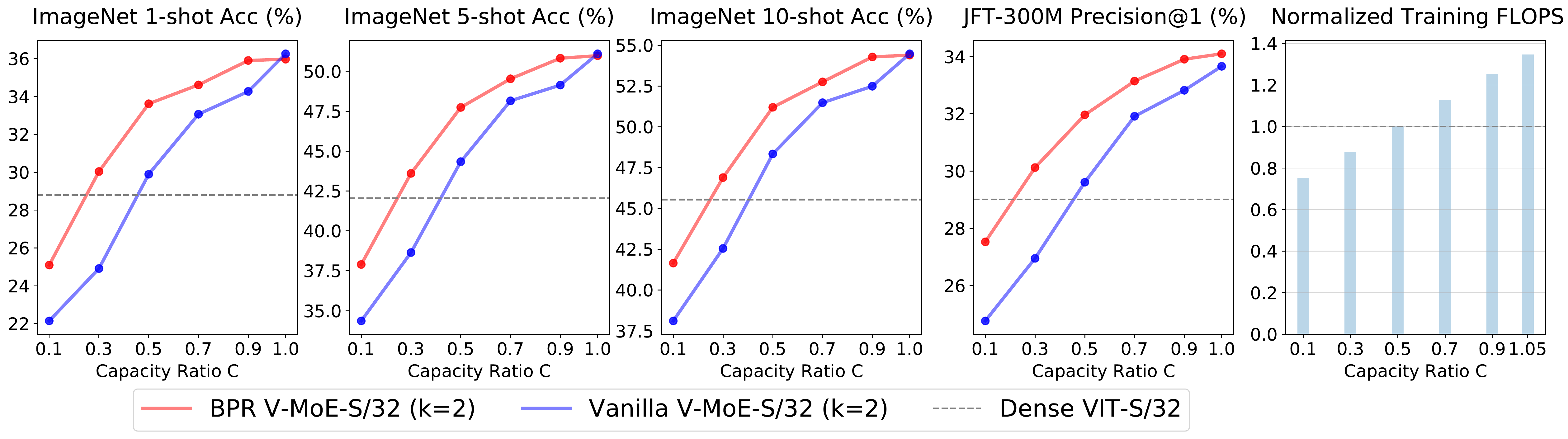}
\caption{Training with \maxrouting{}. Model: \abbv{}-S/32, $k = 2$. Mean over 4 seeds.}
\label{im:training_max_routing_vit_s}
\end{figure}

\Cref{im:training_max_routing_vit_s_k1,im:training_max_routing_vit_s} show \abbv{}-S/32 with $k=1$ and $k=2$ respectively.
We are able to match the dense upstream performance with around 80\% of the training FLOPs in both cases.
Also, around 85 and 80\% of the training FLOPs are enough to match the few-shot evaluation performance in each case.
Overall, we can save 20\% of the FLOPs while training a small model like \abbv{}-S/32.

\begin{figure}[h]
\centering
\includegraphics[width=1.0\textwidth]{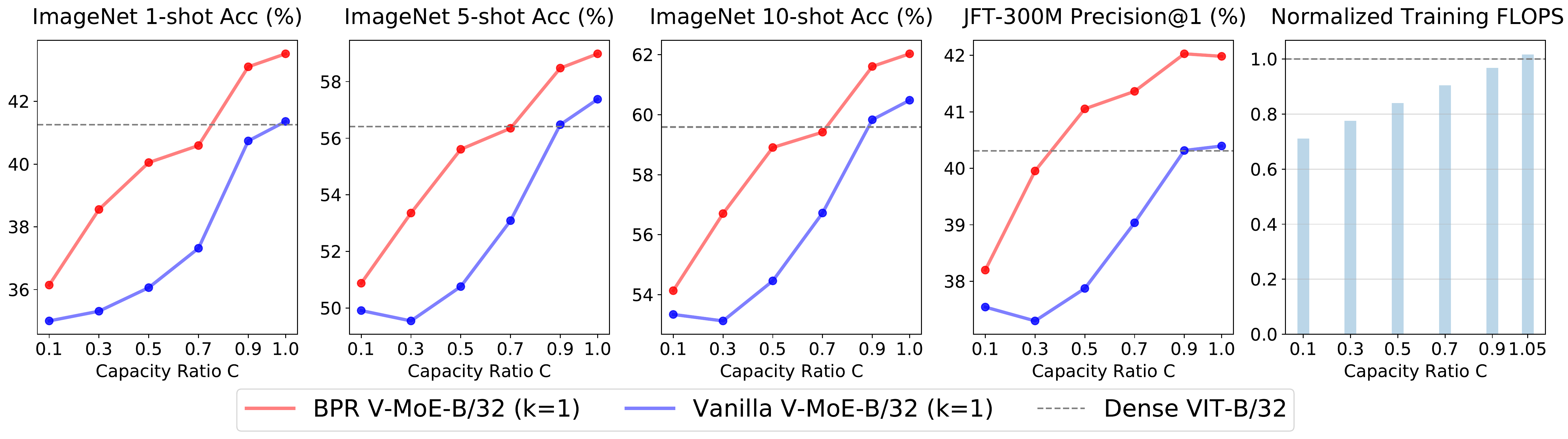}
\caption{Training with \maxrouting{}. Model: \abbv{}-B/32, $k = 1$. Mean over 4 seeds.}
\label{im:training_max_routing_vit_b_k1}
\end{figure}

\begin{figure}[h]
\centering
\includegraphics[width=1.0\textwidth]{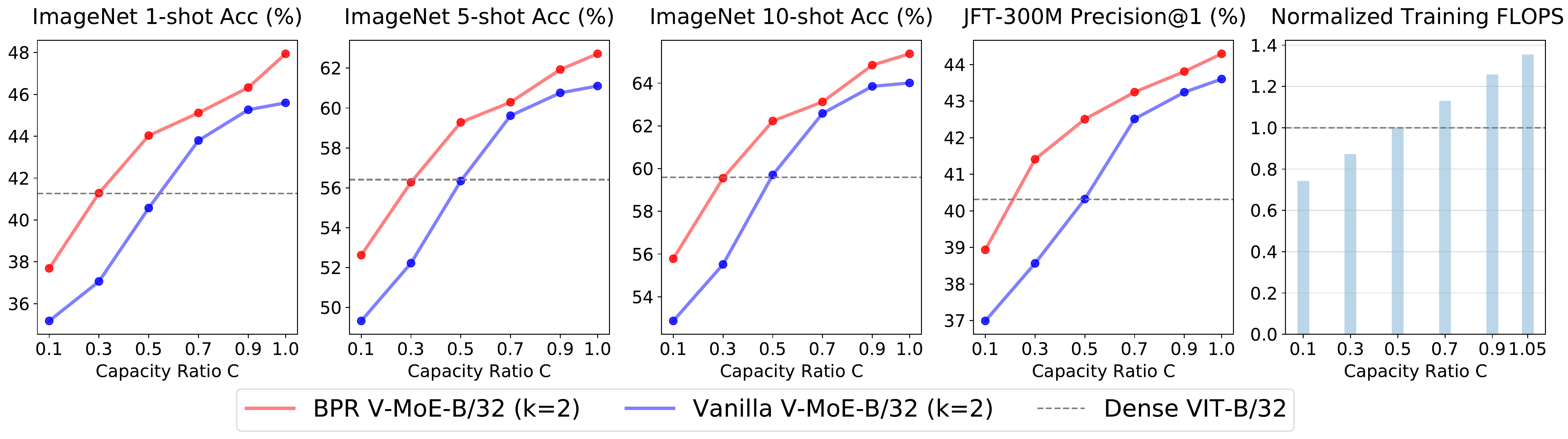}
\caption{Training with \maxrouting{}. Model: \abbv{}-B/32, $k = 2$. Mean over 4 seeds.}
\label{im:training_max_routing_vit_b}
\end{figure}

\Cref{im:training_max_routing_vit_b_k1,im:training_max_routing_vit_b} show \abbv{}-B/32 with $k=1$ and $k=2$ respectively.
Again, with at most 80\% of the training FLOPs the expert models match the upstream performance of its dense counterpart.
Also, we can save around 10\% of the training FLOPs while keeping or improving the few-shot representation quality.

\begin{figure}[h]
\centering
\includegraphics[width=1.0\textwidth]{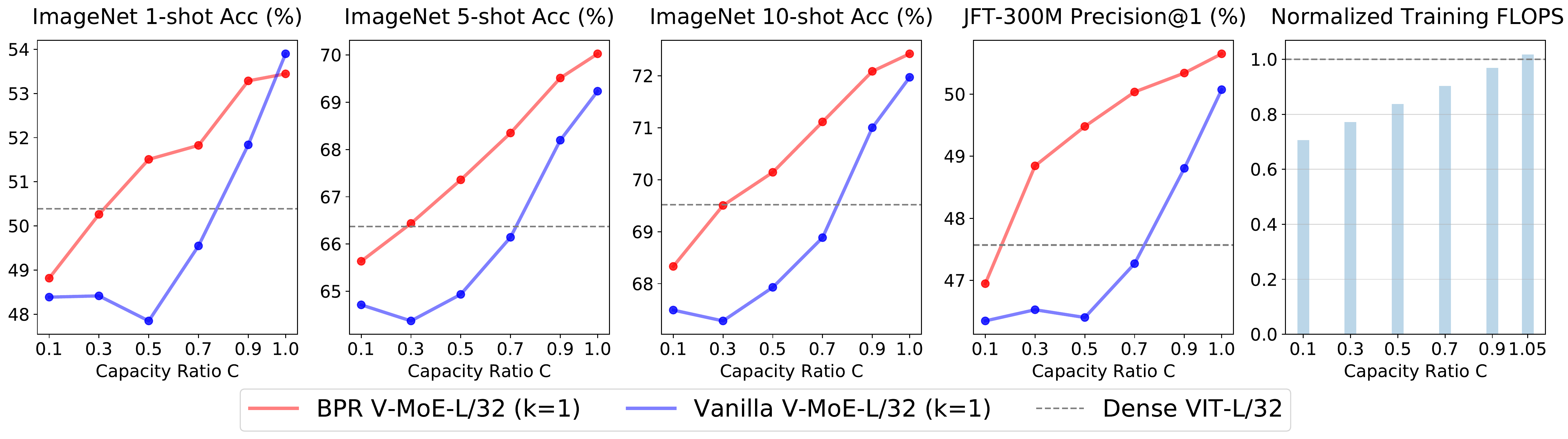}
\caption{Training with \maxrouting{}. Model: \abbv{}-L/32, $k = 1$. Mean over 4 seeds.}
\label{im:training_max_routing_vit_l_k1}
\end{figure}

\begin{figure}[h]
\centering
\includegraphics[width=1.0\textwidth]{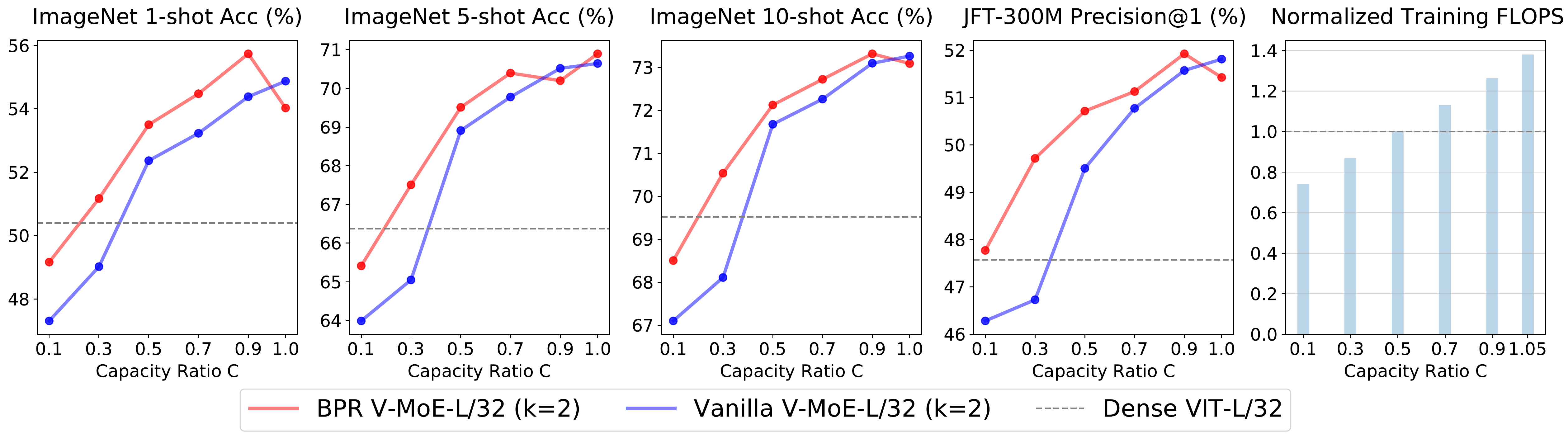}
\caption{Training with \maxrouting{}. Model: \abbv{}-L/32, $k = 2$. Mean over 4 seeds.}
\label{im:training_max_routing_vit_l_k2}
\end{figure}

Finally, \Cref{im:training_max_routing_vit_l_k1,im:training_max_routing_vit_l_k2} presents the results for VIT-L/32 with $k=1$ and $k=2$.
Remarkably, between 70 and 75\% of the training FLOPs are enough to mimic the upstream dense performance.
Note that, when $k=2$, the lowest capacity ($C=0.1$) already outperforms the dense upstream precision.
The expert model is also able to deliver identical few-shot performance while saving more than 20\% of the training FLOPs.

\clearpage

\subsection{Applied during Fine-tuning}
\begin{figure}[h]
\centering
\includegraphics[width=1.0\textwidth]{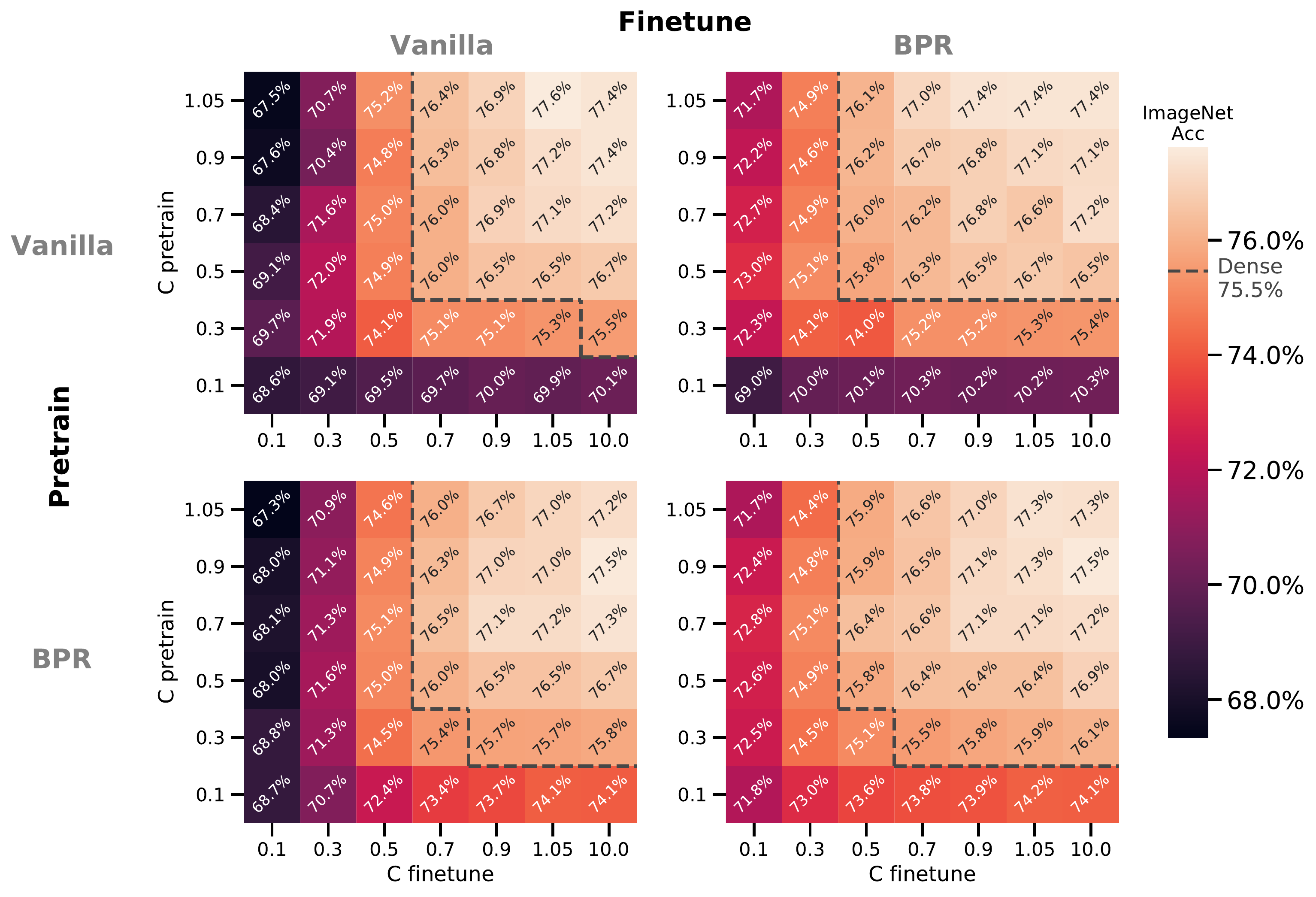}
\caption{Fine-tuning with \maxrouting{}. Model: \abbv{}-S/32, $k = 2$.}
\label{im:finetuning_max_routing}
\end{figure}

We also investigate the effect of using the max-routing algorithm in fine-tuning. We consider \abbv{}-S/32 models pre-trained at various capacities both with and without \maxrouting{}. We fine tune them on ImageNet to see the effect of priority routing during downstream fine-tuning and inference. This is shown in \cref{im:finetuning_max_routing}.

There are a few conclusions that can be garnered:
\begin{itemize}
    \item Downstream fine-tuning results are significantly impacted by capacity, with accuracy reducing from 77.4\% to 68.6\% by reducing capacity to 0.1.
    \item \maxrouting{} can recover some of this performance drop; if it is applied during pre-training and fine-tuning, accuracy increases to 71.8\% at the same capacity.
    \item It is more important to retain high capacity during fine-tuning than while pre-training. For example, with priority routing applied both at downstream and upstream, $C=1.05$ during pre-training with $C=0.1$ during fine-tuning has accuracy 71.7\%, but the inverse is significantly better with accuracy 74.1\%. In both cases, priority routing is key to ameliorating the effect of low capacity during fine-tuning and pre-training.
\end{itemize}

\clearpage
\section{Examples of patch dropping}
\label{app_skip_patch_images}

\begin{figure}[h]
\centering
\includegraphics[width=1.0\textwidth]{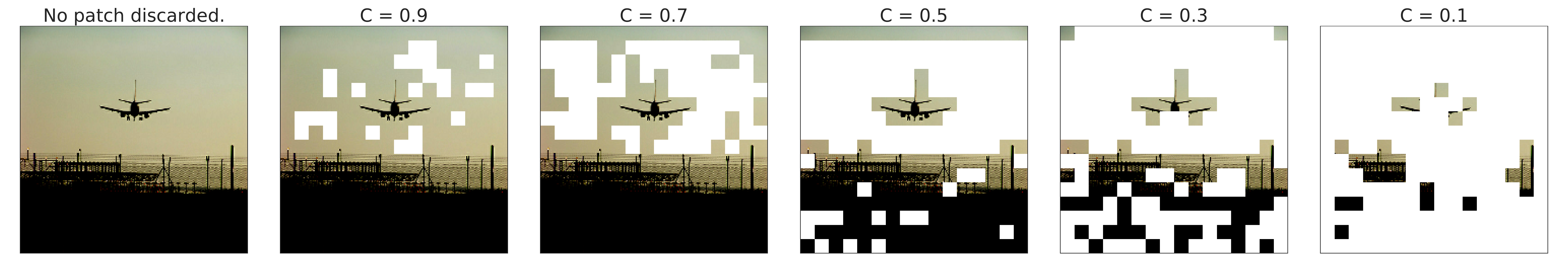}

\centering
\includegraphics[width=1.0\textwidth]{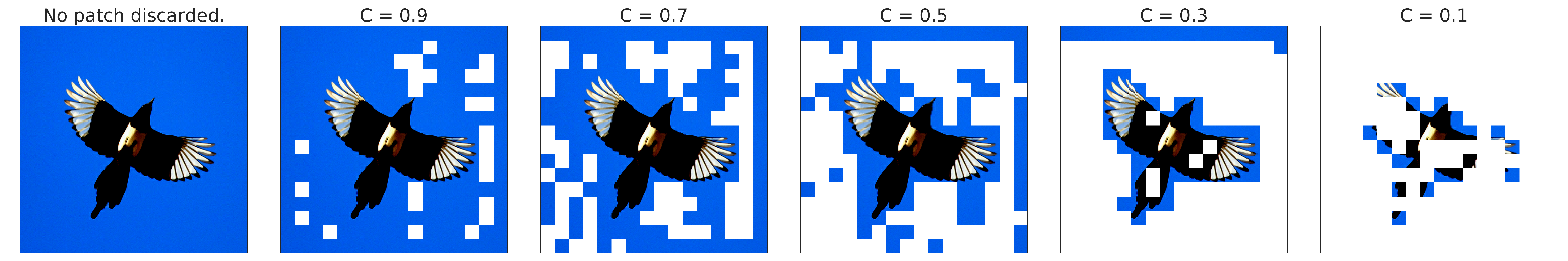}

\centering
\includegraphics[width=1.0\textwidth]{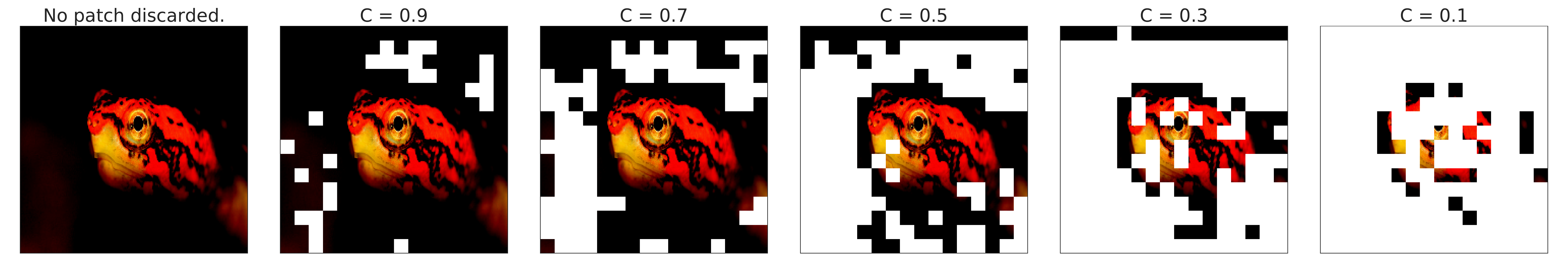}

\centering
\includegraphics[width=1.0\textwidth]{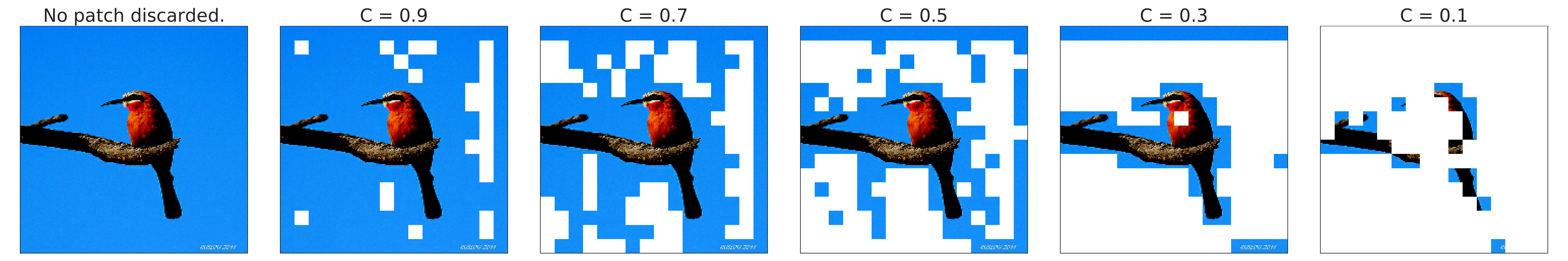}

\centering
\includegraphics[width=1.0\textwidth]{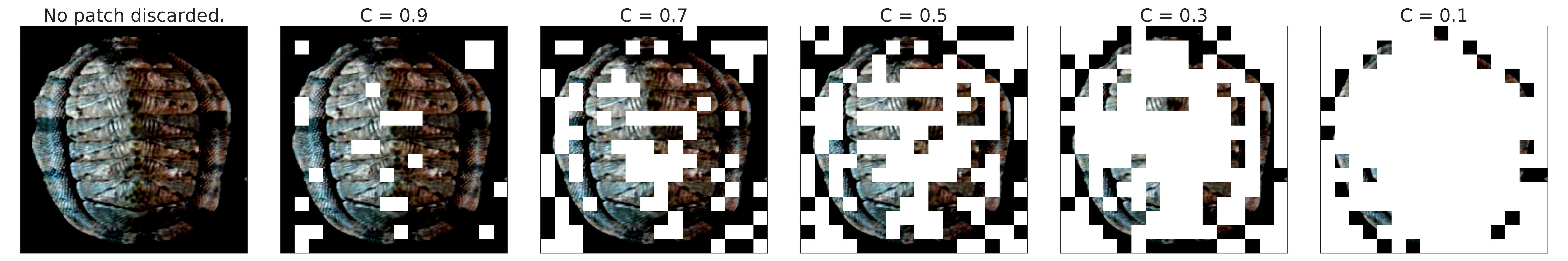}

\centering
\includegraphics[width=1.0\textwidth]{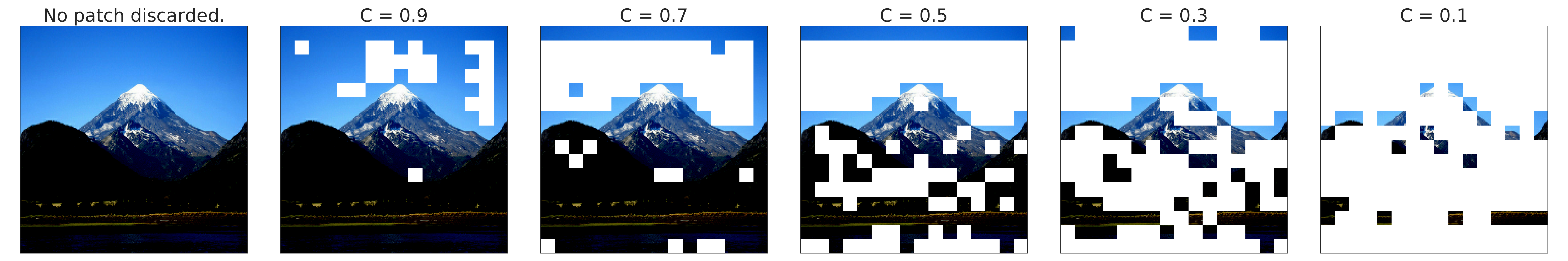}

\centering
\includegraphics[width=1.0\textwidth]{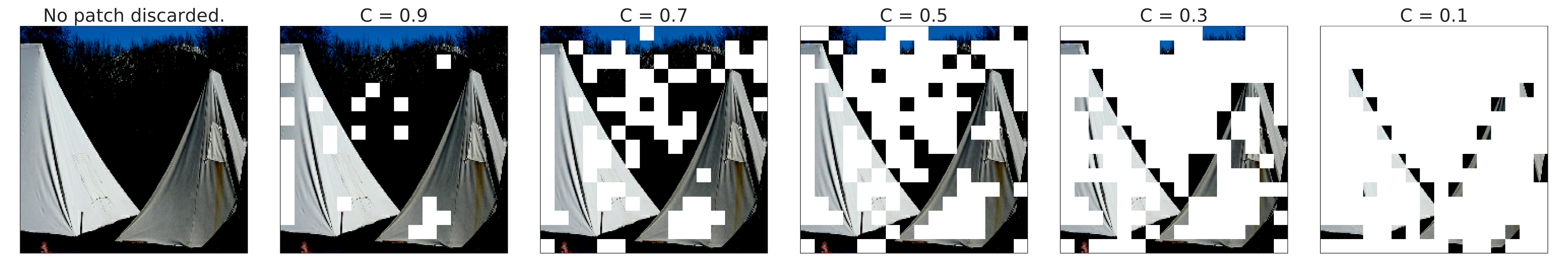}

\end{figure}

\begin{figure}[h]
\centering
\includegraphics[width=1.0\textwidth]{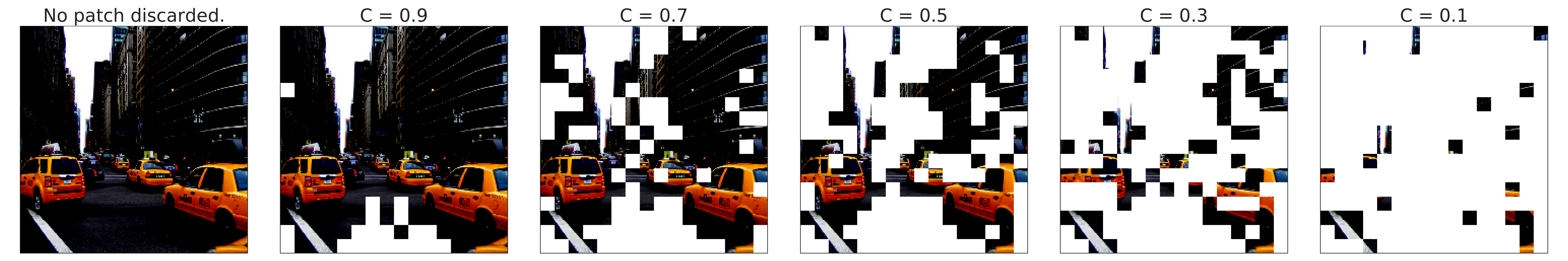}

\centering
\includegraphics[width=1.0\textwidth]{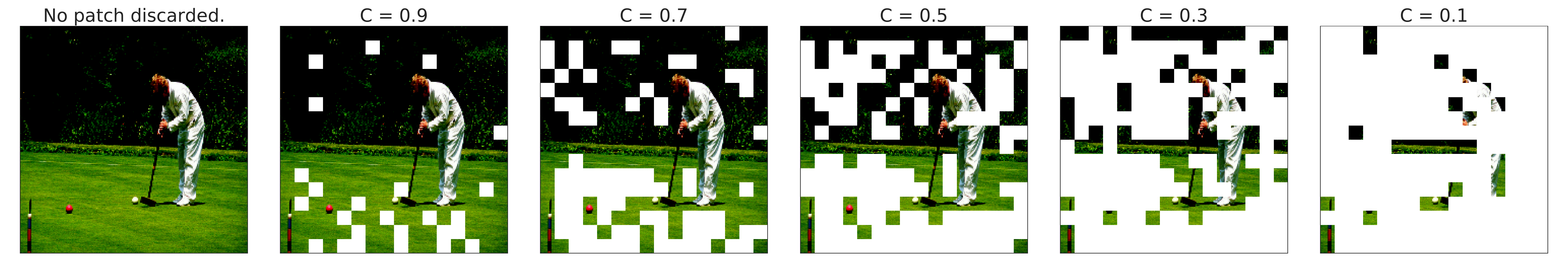}

\centering
\includegraphics[width=1.0\textwidth]{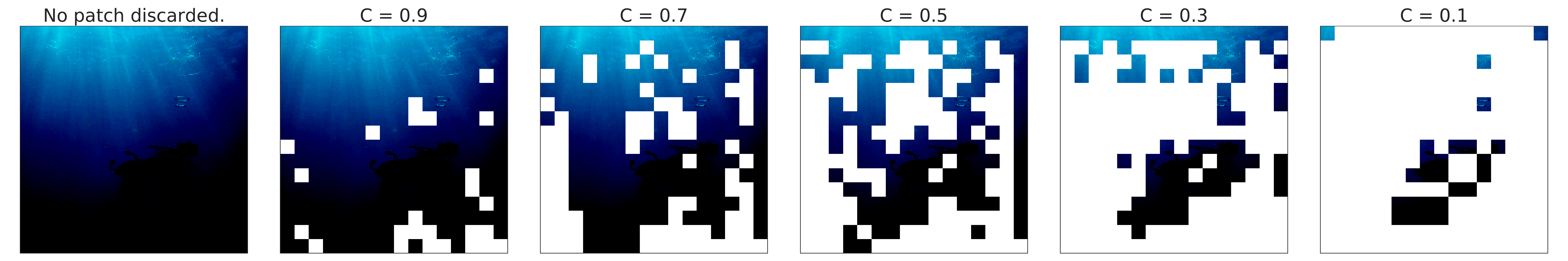}

\centering
\includegraphics[width=1.0\textwidth]{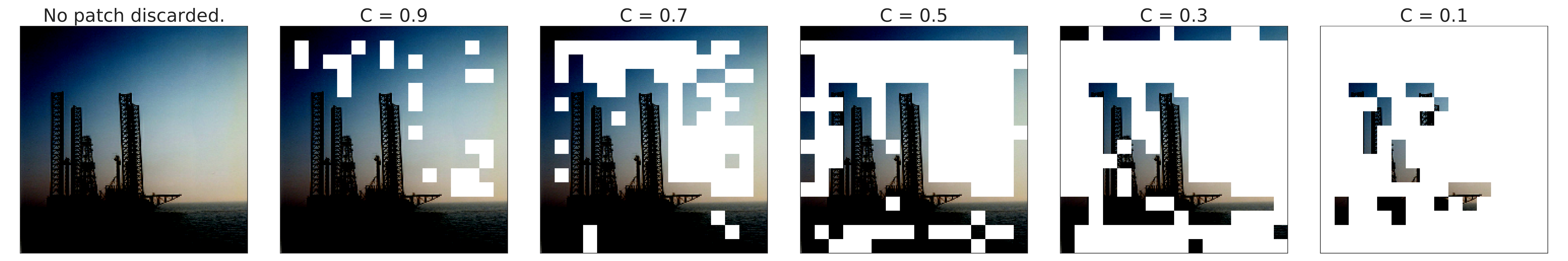}

\centering
\includegraphics[width=1.0\textwidth]{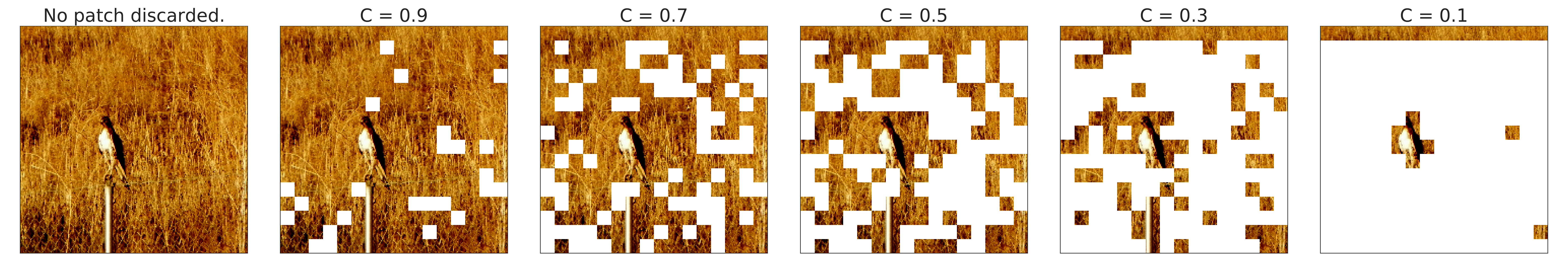}

\centering
\includegraphics[width=1.0\textwidth]{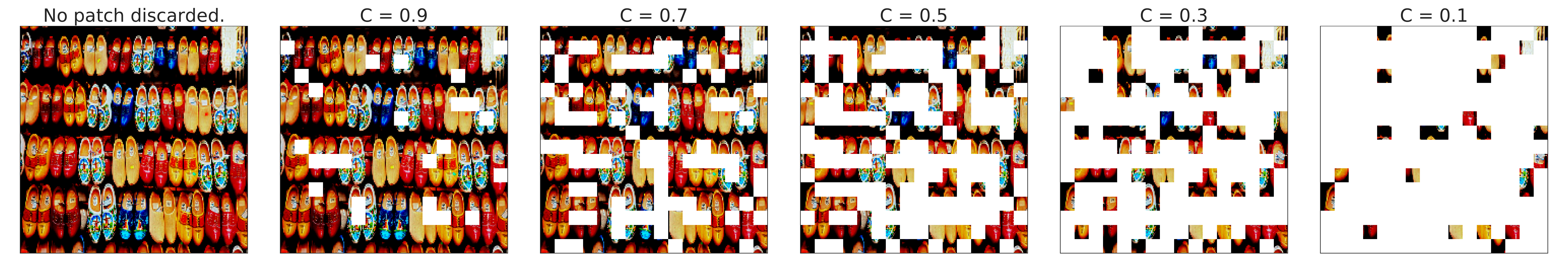}

\centering
\includegraphics[width=1.0\textwidth]{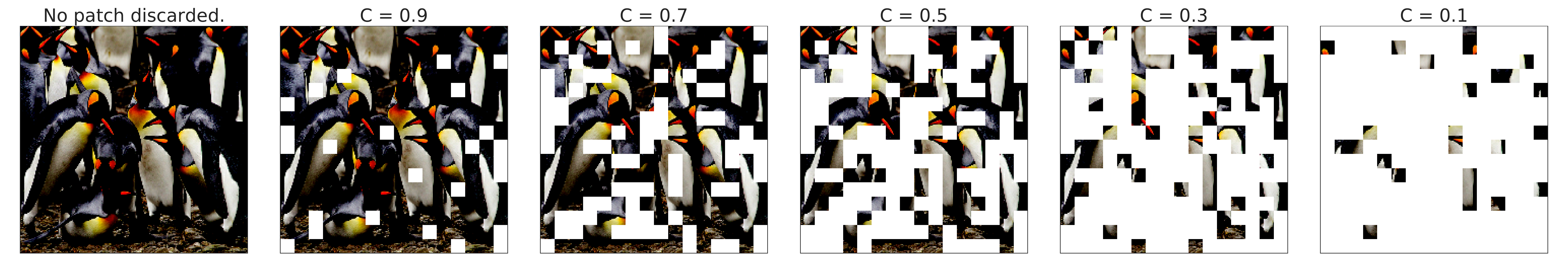}

\end{figure}
\clearpage
\section{Model Analysis}
\label{app_analysis}

Several previous works have proposed deep models based on mixture of experts; most of them have also presented promising results.
Unfortunately, despite the current excitement regarding this set of techniques, little is indeed known about how these complex models internally work.
Exploratory experiments that shed light into the mechanics of routers and expert specialization could inform new algorithms.
We try to provide the first such analysis here, which we actually found useful to develop some of the algorithms presented in the paper.

\subsection{The value of routers}
\label{app_analysis_value_routers}
\begin{figure}[h]
\centering
\includegraphics[width=1.0\textwidth]{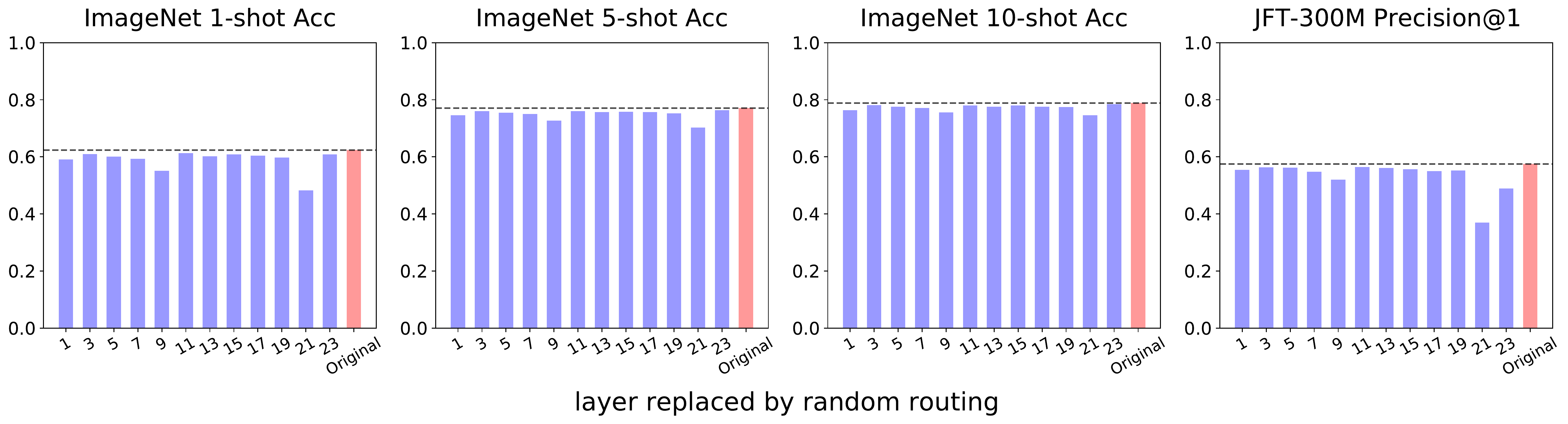}
\caption{Replace one layer at a time by a random router for \abbv{}-L/16.}
\label{im:random_router_one}
\end{figure}

The most natural question to ask after training a sparse model is whether the learned routers are doing something useful.
There are several potential ways things could go wrong.
For example, the router could just become a load balancer if experts end up implementing very similar functions.
Alternatively, the router may simply choose sub-optimal assignments.
As a first test, we replace one router at a time with a uniformly random router.
For this, we take a pre-trained model --in particular, a \abbv{}-L/16 with $k=2$--, and re-evaluate its upstream and few-shot performance when perturbing the routers.
\Cref{im:random_router_one} contains the results.
In red, we show the original performance for the pre-trained model ---that is, when applying all the learned routers.
We also show the impact of replacing each router independently and in isolation with a uniformly random router --the layer ID is shown in the $x$-axis.
In particular, the new router samples the weights in a white Gaussian fashion, so every pair of experts is equally likely to be the TOP-$k$ choice for any given input. We also tried to randomly permute the output weights ---so to avoid a distributional shift in applied routing weights---and it worsened results.

Overall, we observe that the last two layers --21 and 23-- provide an essential routing for the upstream model to work well (validation precision at 1 in JFT).
We have seen a similar pattern in other models.
Interestingly, the previous to last MoE layer (21-th in this case) is the one where getting the routing right is the most important.
The model is robust to mis-routing at most intermediate layers---layer 9 is an exception here.
This observation motivated us into trying to train sparse models with MoE layers only at the very end---21 and 23, for example---with excellent results (and computational savings).

\begin{figure}[h]
\centering
\includegraphics[width=1.0\textwidth]{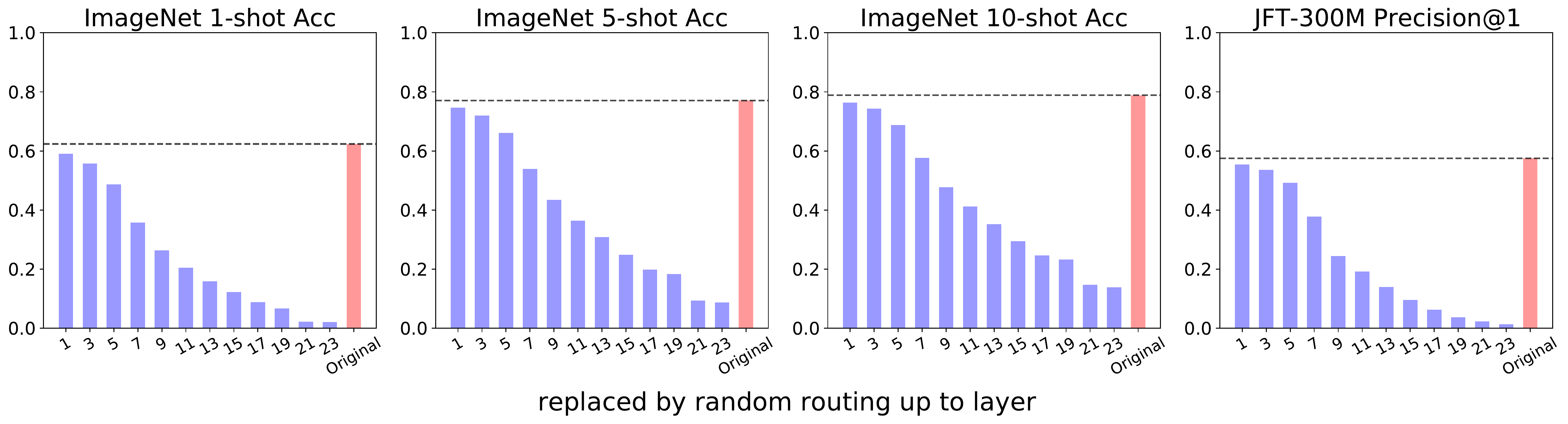}
\caption{Replace all layers up to a given one by random routers for \abbv{}-L/16.}
\label{im:random_router_up_to}
\end{figure}

After analyzing the results in \cref{im:random_router_one}, a natural follow up question is whether the model is robust to compounded mis-routing?
We answer this question by looking at what happens when we replace a number of consecutive MoE layers with uniformly random routers.
\Cref{im:random_router_up_to} shows the outcome.
We start from the bottom MoE layer, and for every MoE layer $i$ in the network, we evaluate the model where routers in 1 to $i$ layers (both included) act randomly.
Unfortunately, in this case, performance drops quickly as one would expect.
Tokens are following random walks (if we ignore capacity issues) up to some point, and then using the correct remaining routers. If the random walk is long enough, the performance is severely degraded.
We conclude the token paths in a trained model are far from random or meaningless.

\subsection{Specialized experts}
\label{app_analysis_specialized_experts}

\begin{figure}[h]
\centering
\includegraphics[width=1.0\textwidth]{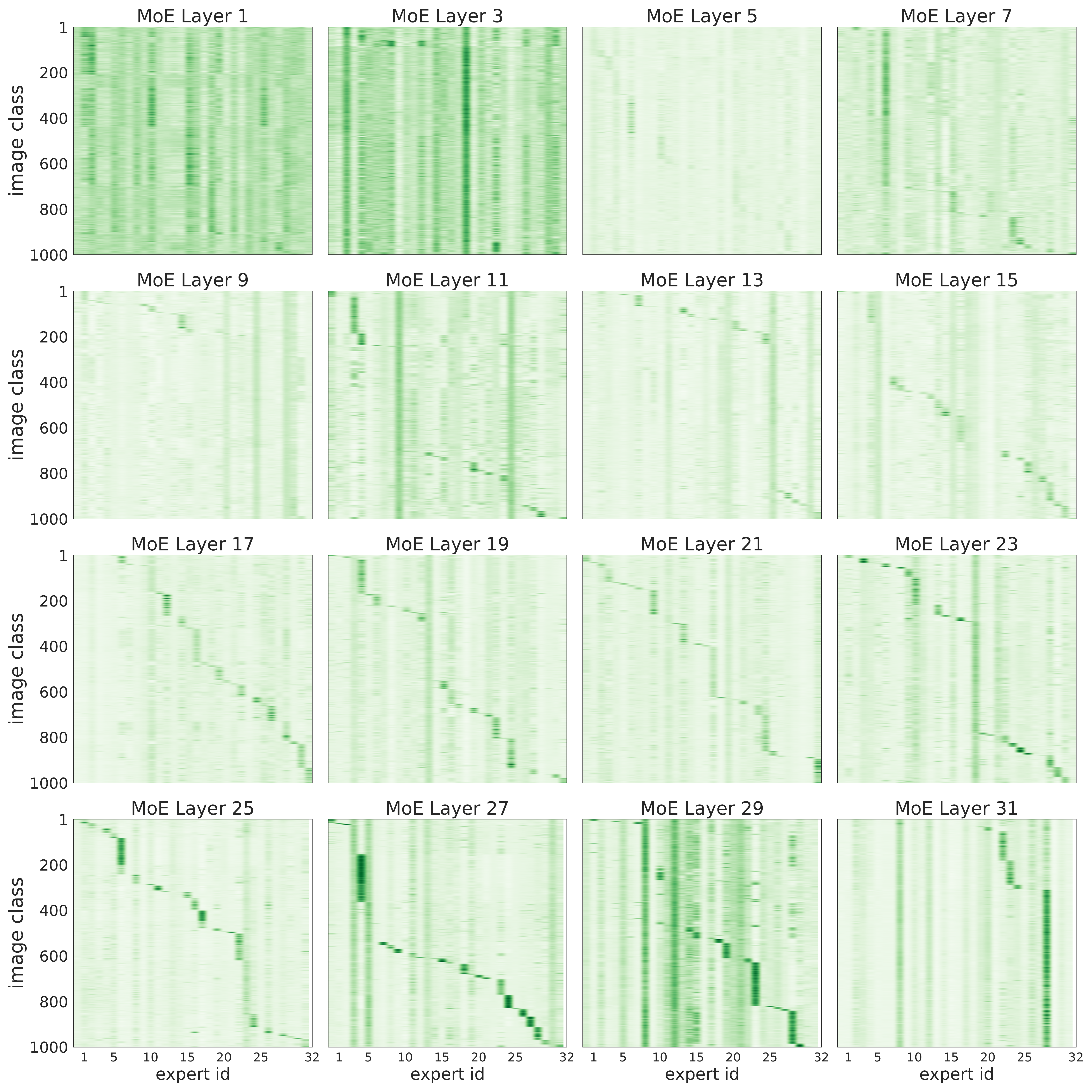}
\caption{
\textbf{Average weight for selected experts per class}.
We show the 16 MoE layers of an every-2 \abbv{}-H/14. The $x$-axis corresponds to the 32 experts in a layer. The $y$-axis are the 1000 ImageNet classes; orderings for both axes are different across plots.
For each pair (expert $e$, class $i$) we show the average routing weight for the patches corresponding to all images with class $i$ for that particular expert $e$.}
\label{im:routes_class_experts_all}
\end{figure}

In \cref{im:routes_class_experts_all} we show results for a massive model with 24 MoE layers, each of them with 32 experts.
After training the model on JFT and fine-tuning it on ImageNet, we did forward passes (up to the pre-logits) with ImageNet images.
Each plot corresponds to one MoE layer, in increasing order.
The $x$-axis corresponds to the 32 experts per layer, and the $y$-axis are the 1000 ImageNet classes (in different adjusted orders per plot; i.e., class 5 in layers $i$ and $j$ are generally different for $i \neq j$).
For each pair (expert $e$, class $i$) we show the average routing weight for the patches corresponding to all images with class $i$ for that particular expert $e$.
Intuitively, this is a proxy for how much images of class $i$ activate and use expert $e$.
\Cref{im:routes_class_experts_all} shows strong expert-class correlations for the last few layers.
In other words, it seems experts specialize in discriminating between a small set of classes (those primarily routed through the expert).
In the initial MoE layers, however, we do not observe such correlation, and we conclude the experts may focus on different aspects of the patches that may be common to all classes (background, basic shapes, etc.).

\begin{figure}[h]
\centering
\includegraphics[width=1.0\textwidth]{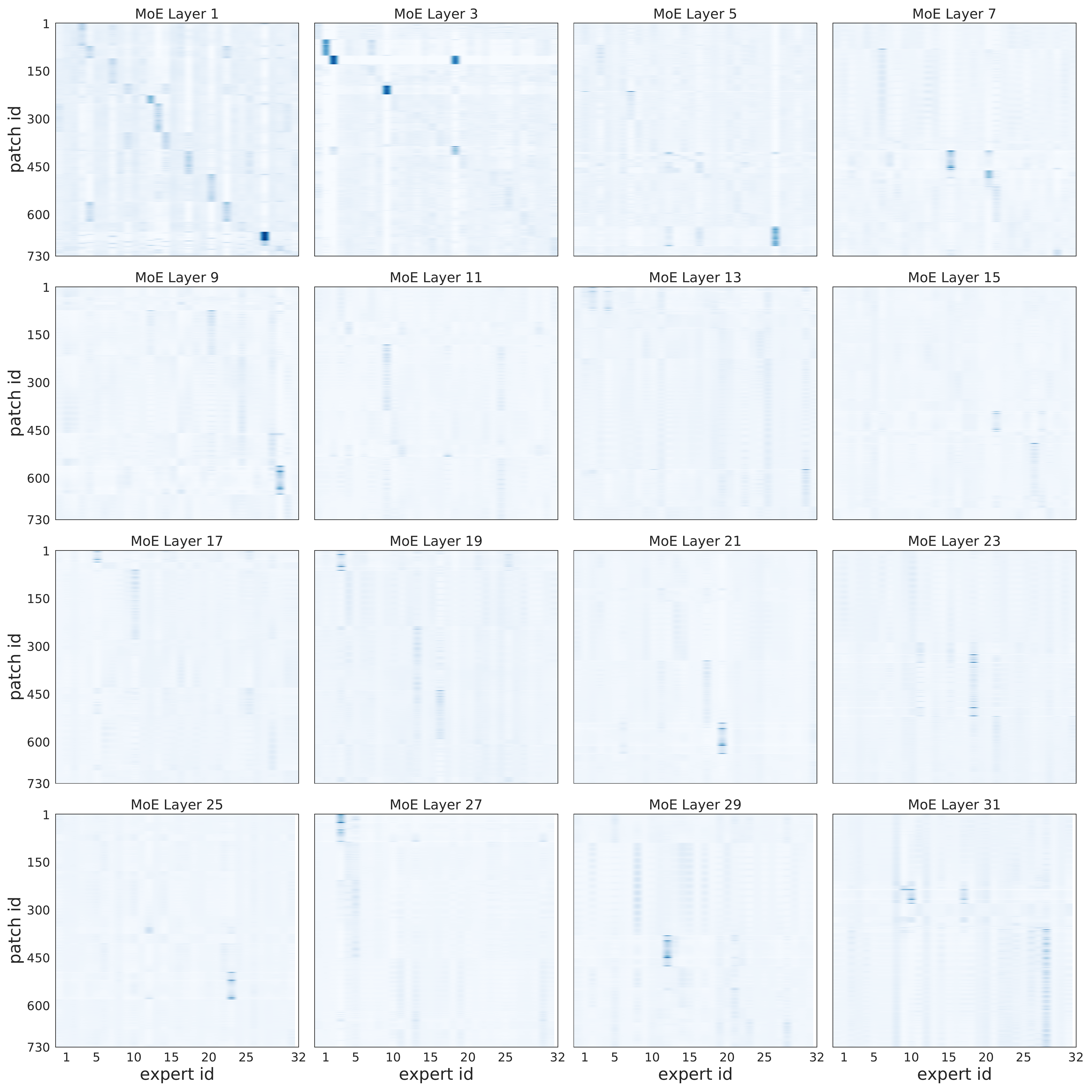}
\caption{
\textbf{Average weight for selected experts per patch position} on a every-2 \abbv{}-H/14 fine-tuned model.
The $x$-axis corresponds to the 32 experts in a layer. The $y$-axis are the 730 patches in ImageNet images with 14x14 patch size, at (384, 384, 3) resolution; orderings for the x-axis are different across plots.
For each pair (expert $e$, patch-id $i$) we show the average routing weight for all the patches with patch-id $i$ that were assigned to that particular expert $e$.}
\label{im:routes_patch_experts}
\end{figure}

To further investigate the logic behind the first layers of experts, \cref{im:routes_patch_experts} shows the correlation between selected experts and the patch id or location.
The model partitions each image in the same number of patches --say if the patch size is 14x14, and images are 224x224x3, then there are 256 patches (sometimes we add an additional learnable token).
We add a positional embedding to each patch that helps the model track the relative ordering.
In this case, we see that for the first few MoE layers, the routers tend to distribute patches to experts according to their patch id.
One simple explanation could be that patches in similar positions usually share visual characteristics that one expert learns to handle --say, image corners, backgrounds, or the center of the images with objects.

\subsection{Routing weights distribution}
\label{app_analysis_routing_weights_distribution}
\begin{figure}[h]
\centering
\includegraphics[width=1.0\textwidth]{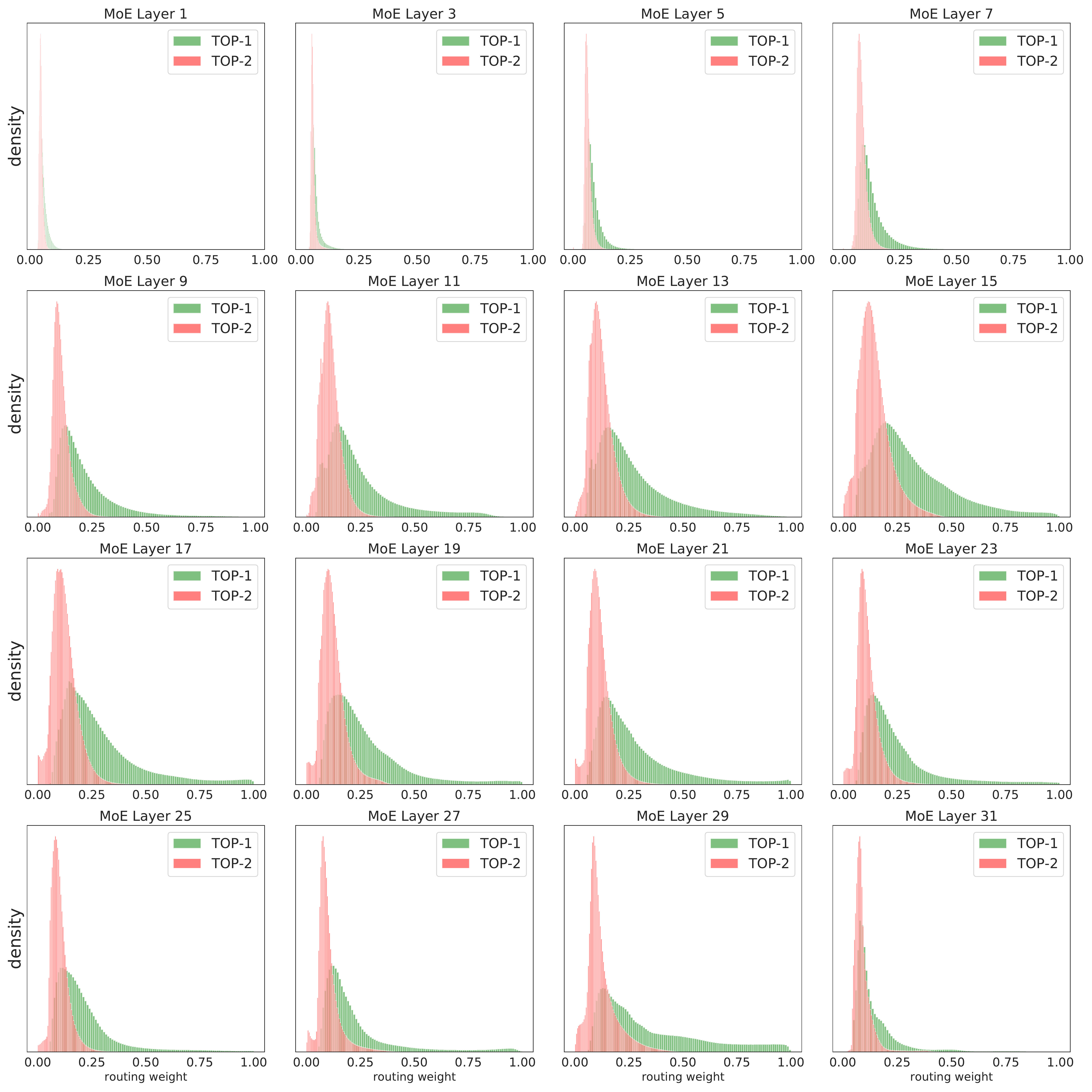}
\caption{\textbf{Routing weight distribution for TOP-1 and TOP-2 selected experts.}
We show the distribution over the TOP-1 (green) and TOP-2 (red) weights for a \abbv{}-H/14 model fine-tuned on ImageNet.
Note for any given patch these weights do not need to add to one ---and in fact they will not---, as we apply the softmax before the TOP-$k$ selection.
}
\label{im:top_k_weight_distribution}
\end{figure}

Most of the key model hyper-parameters, like the number of experts that process each patch $k$ or the expert buffer capacity ratio $C$ that controls the amount of patch dropping, can be adjusted layer-wise.
For example, if we do not see expert specialization in lower layers, we could simply set $k=1$ there to avoid wasting compute.
It may however be useful to increase $k$ in the last few layers to allow for composability of concepts, like when we try to identify several objects.
\Cref{im:top_k_weight_distribution} shows the TOP-1 and TOP-2 weight distribution for an sparse model with $k=2$.
Two main conclusions can be drawn.
First, in lower layers, both choices seem to have a similar magnitude --thus, both indeed contribute to the combined representation.
Moreover, the weights are usually low in this layers --note $1/E \approx 0.03$ is the minimum weight the top selected expert can be assigned--, which one may interpret as the router being somewhat indifferent among experts.
Second, the trend clearly changes when patches travel towards the end of the network.
In particular, the TOP-1 and TOP-2 weight distributions strongly diverge, with the former approaching 1.0 and the latter approaching 0.
This means the intrinsic $k$ at the top of the network is closer to 1 (than the actual $k=2$).
The composability that we mentioned before may not be indeed needed at the patch level, as patches are quite small for large networks (here, 14x14), and it may be difficult to identify several concepts.
Nonetheless, some tail of the distributions shown in \cref{im:top_k_weight_distribution} still uses both experts in the last layers.

\begin{figure}[h]
\centering
\includegraphics[width=1.0\textwidth]{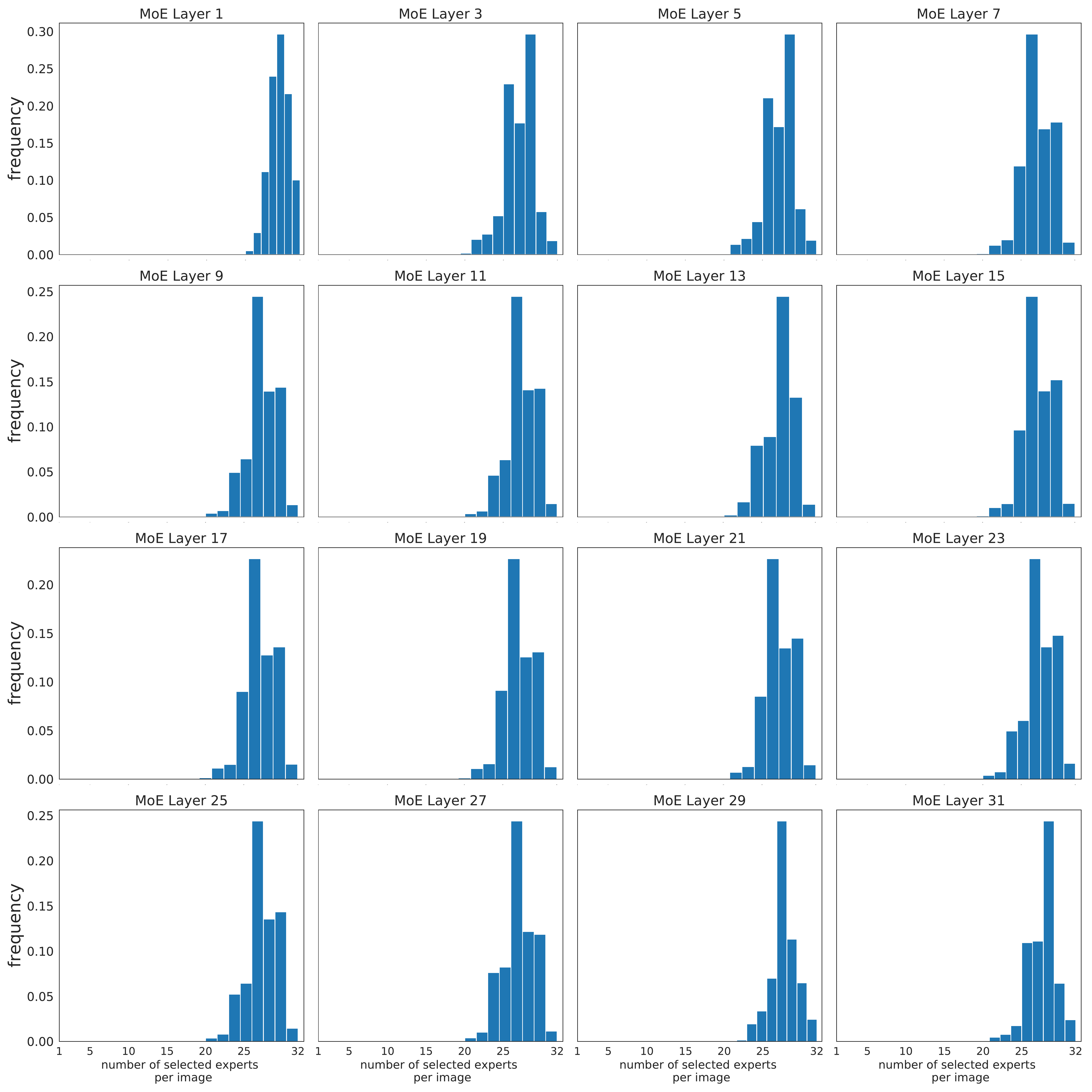}
\caption{
\textbf{Number of selected experts per image (after pooling selection from all patches).}
We show the distribution of total number of used experts per layer per image for a \abbv{}-H/14 model fine-tuned on ImageNet.
In this case, every image has 730 patches.
Even though most experts are selected at least once ---that is what we plot here---, we expect some of the experts to be selected way more often by the patches of an image, and with a higher average weight.
}
\label{im:experts_per_image}
\end{figure}

We would like to remark that each \emph{image} is subject to a large number of routing decisions through its patches.
Concretely, \cref{im:experts_per_image} shows how most images use --on aggregate by pooling over all their patches-- most of the experts in \emph{every} layer.
This motivated our efforts to try to save compute by discarding, or not processing, patches that are not useful for the final classification.
We cover this in detail in \cref{sec:skip_patch}.

\clearpage

\subsection{Changing \texorpdfstring{$k$}{k} at inference}\
\label{app_analysis_routing_changing_k}

We now explore a remarkable aspect of expert models: their flexibility.
Somewhat surprisingly, we have observed sparse models to be fairly robust to mismatches between the training and inference configurations.
In this section, we explore the effect of training with some original $k$ while applying the model at inference time with a different $k^\prime \neq k$.
This can be handy to control (decrease or increase) the amount of FLOPs per input in a particular production system.

\begin{figure}[h]
\centering
\includegraphics[width=1.0\textwidth]{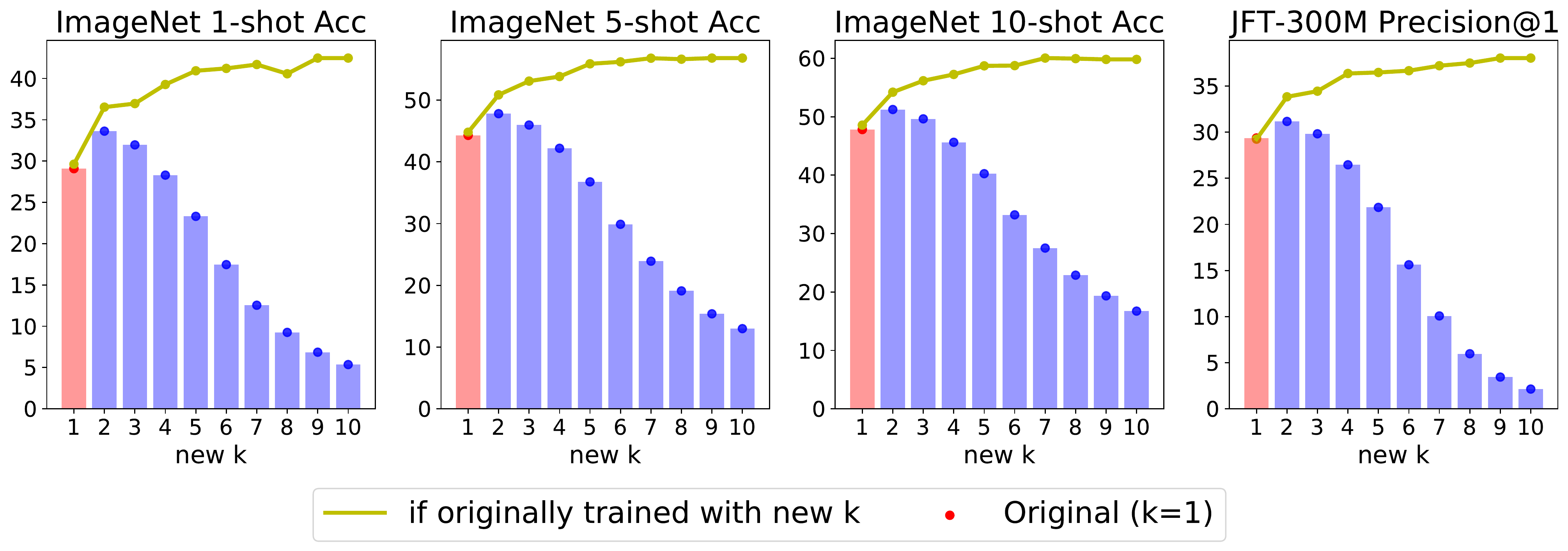}
\caption{Original V-MoE-S/32 every-2 model was trained with $k=1$.}
\label{im:from_k_to_new_k_1_full}
\end{figure}

\Cref{im:from_k_to_new_k_1_full} is based on a \abbv{}-S/32 model trained with $k=1$.
We evaluate the upstream and few-shot metrics at inference time for a range of new $k^\prime$s.
Note we do not perform any further training in any case, and the model parameters (including the router) are identical in all cases.
The only difference is the number of experts we apply to each input, the amount of the network we activate.
In red we show the original model's performance, and in blue the new ones.
Finally, for each $k^\prime$, in yellow, we show the performance of a \abbv{}-S/32 model trained \textbf{originally} with $k = k^\prime$ --which, as we expected, increases in $k^\prime$.
We see that increasing the value of $k$ from its original value ($k=1$) at inference time by one or two units actually significantly improves performance, both upstream and few-shot.
However, at some point, if the new $k^\prime$ is too large, the performance starts suffering --probably as the model is not prepared for the new distribution of total output routing weights applied in the linear combination, and sub-optimal experts for a given input start contributing to its representation.

\begin{figure}[h]
\centering
\includegraphics[width=1.0\textwidth]{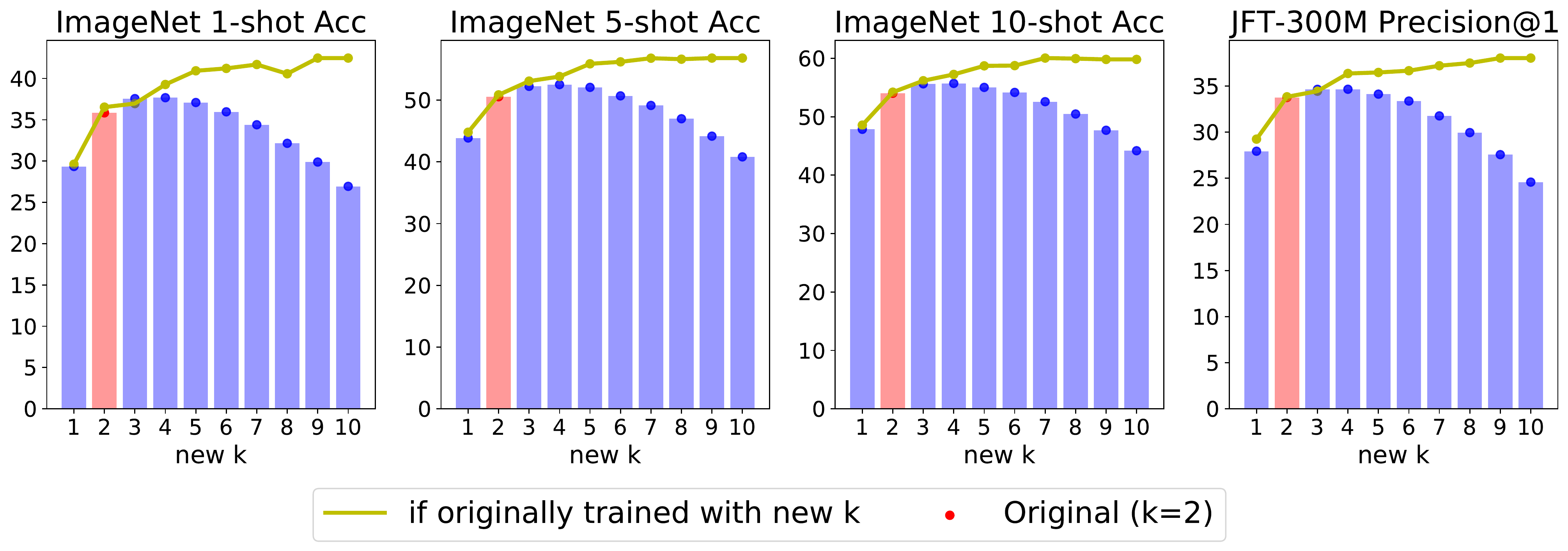}
\caption{Original V-MoE-S/32 every-2 model was trained with $k=2$.}
\label{im:from_k_to_new_k_2_full}
\end{figure}

\Cref{im:from_k_to_new_k_2_full} shows the case where the original model is a \abbv{}-S/32 with $k=2$.
The trends are somewhat similar.
By applying $k^\prime=3$ or $k^\prime=4$ we obtain modest improvements, whereas by decreasing $k$ to $k^\prime=1$ we obtain a performance very similar to the performance of a model trained \emph{directly} with $k=1$, especially for few-shot.
This is interesting, as we can devote more FLOPs for training by setting $k=2$ upfront, while deferring the choice of inference $k$ without losing potential performance.
We explored these ideas further in \cref{sec:skip_patch}.
Also, the drop in performance for large values of $k$ is less severe in this case, probably due to the fact that the trained model was used to combine several different experts (not the case for \cref{im:from_k_to_new_k_1_full}).

\begin{figure}[h]
\centering
\includegraphics[width=1.0\textwidth]{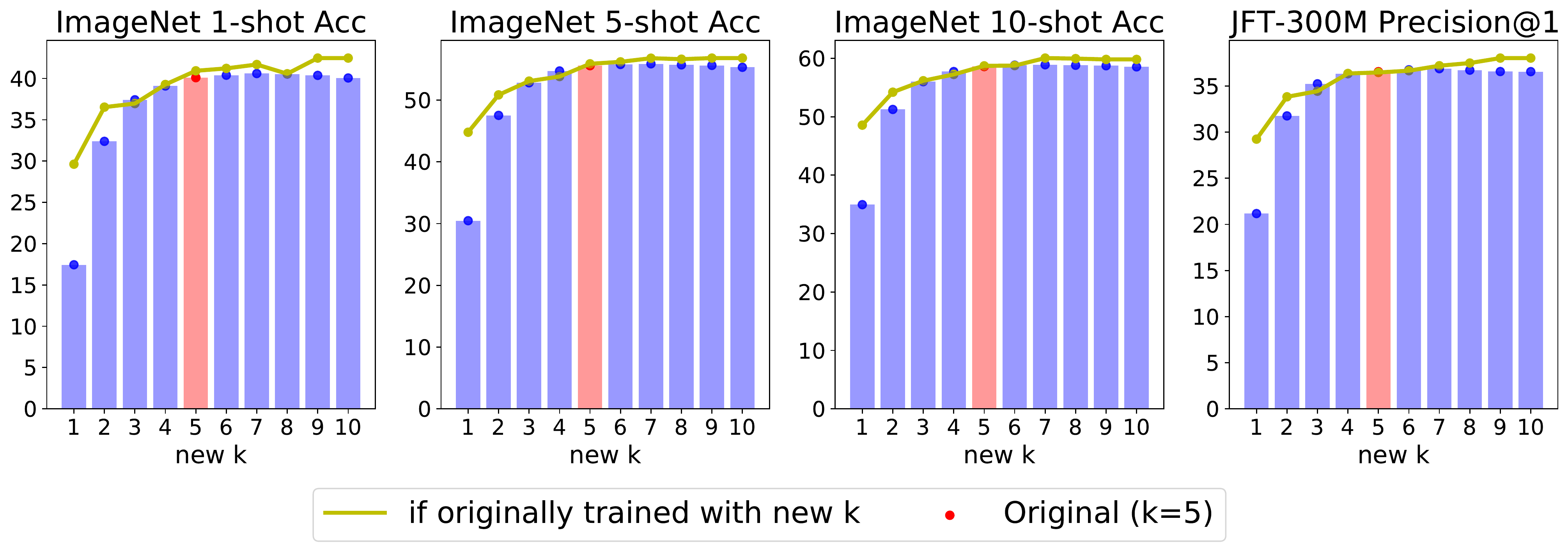}
\caption{Original V-MoE-S/32 every-2 model was trained with $k=5$.}
\label{im:from_k_to_new_k_5_full}
\end{figure}

Finally, in \cref{im:from_k_to_new_k_5_full} we present the case where the upstream model was trained with $k=5$.
This is an expensive model to train, and we see we can change the inference value of $k$ from $k^\prime = 3$ to $k^\prime = 7$ with results that are similar to their optimal value, if we had trained with those values in the first place. At this point the model is stable enough to deal with large values of $k$, but it suffers way more when we set $k^\prime = 1$, as the model is not used to picking a single expert and --we suspect-- the TOP-1 expert may not carry so much importance or weight for this model where five experts were selected per input while training.
Of course, it may not just be a matter of routing weight distribution.
The expert themselves may be quite different when training with $k=1$ --say, more self-contained-- and with $k=5$ --perhaps more team-players.

\subsection{Changing \texorpdfstring{$k$}{k} during fine-tuning}
\begin{figure}[h]
\centering
\includegraphics[width=0.65\textwidth]{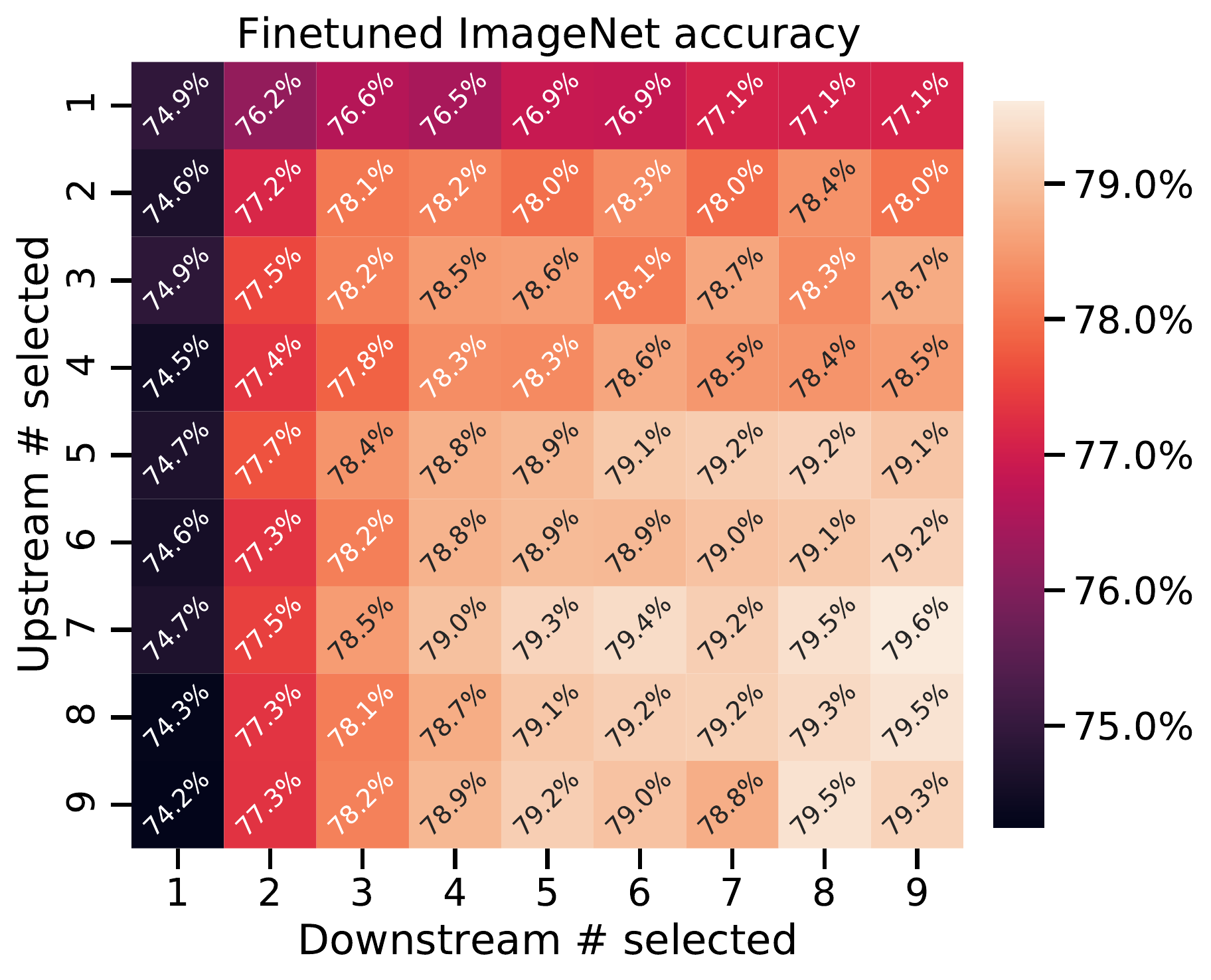}
\caption{Varying $k$ (number of selected experts) at fine-tuning/inference times for \abbv{}-S/32 models pre-trained with different values of $k$.}
\label{im:vary_k_finetuning}
\end{figure}

\
We also consider the effect of adjusting the number of selected experts during fine-tuning and inference. We consider the aforementioned \abbv{}-S/32 models, with 32 experts, pre-trained with $k = \{1, ..., 9\}$ experts. These models are then fine-tuned with varied $k$. We show the result of this in \cref{im:vary_k_finetuning}.
As one may expect given our previous results, generally increasing $k$ improves performance. Regardless of upstream $k$, generally accuracy improves from increasing $k$ during fine-tuning. Similarly, increasing $k$ during pre-training improves performance downstream.

Conversely, when $k = 1$ downstream, all models fail to improve from pre-training with higher upstream $k$. Models pre-trained with $k>1$ seemingly learn to \textit{combine} expert outputs, in that they do not generalize as well to selecting a single expert downstream, and lose the benefits of pre-training with larger $k$.

\subsection{Pre-training with less data}
\begin{figure}[h]
\centering
\includegraphics[width=\textwidth]{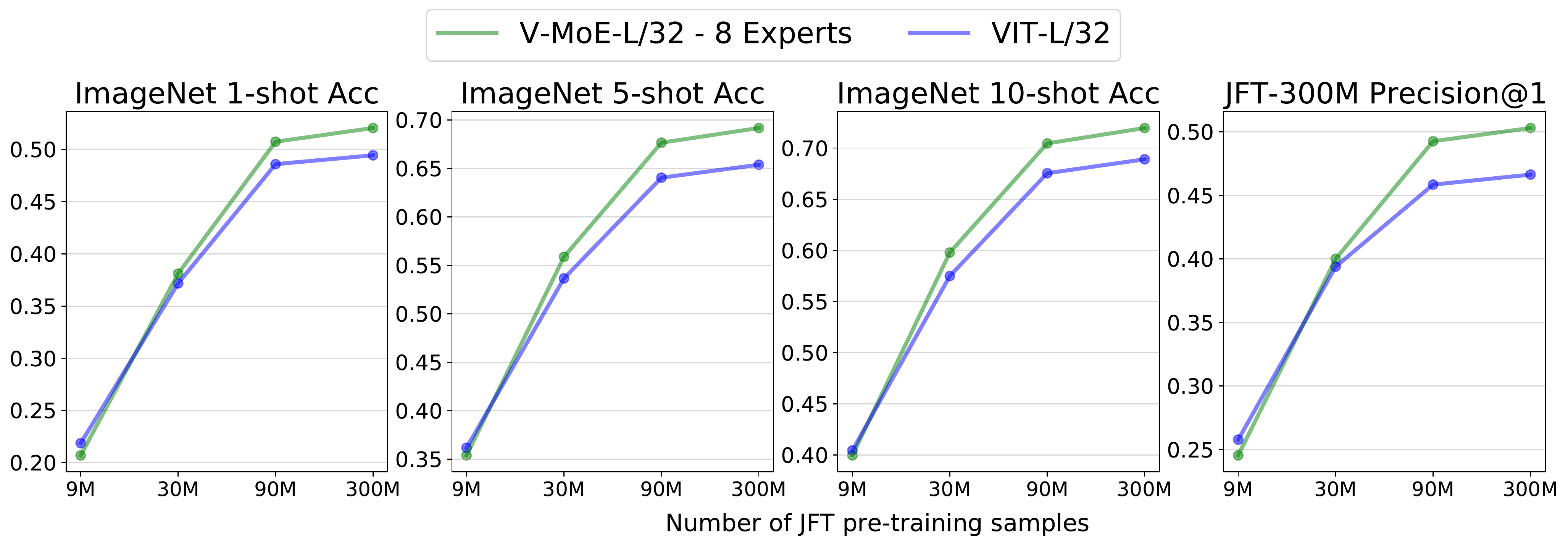}
\caption{\textbf{The effect of varying the amount of pre-training data.}
We compare the performance of \abbv{}-L/32 and VIT-L/32 for increasing data sizes.
In particular, we take subsets of JFT-300M with 9M, 30M, 90M, and 300M datapoints ---note the full dataset contains around 305M datapoints.
Given that we train with smaller datasizes, we decided to use 8 experts rather than 32 (every-2).
At the lowest data size (9M, around 3\% of the original), the MoE model is not able to leverage its extra-capacity.
For the remaining ones, starting at 30M (around 10\% of the original dataset), it does.
}
\label{im:data_efficiency}
\end{figure}

We have shown that the standard recipe of pre-training with large datasets allows use of powerful sparse models on downstream vision tasks where less data is available. The question naturally arises: do these models require large amounts of data upstream?
We present here some initial explorations in this direction.
\paragraph{Training on JFT300M with less data.}
We first train a \abbv{}-L/32 on subsets of JFT300M. This was also done for dense models in \cite{dosovitskiy2020image}, and in \cref{im:data_efficiency} we compare directly to their results. \abbv{} seems initially fairly robust to reduced data, but after reducing to 9M pre-training samples (3\% of the dataset), it becomes slightly preferable to instead train a dense model.

\begin{figure}
\centering
\begin{subfigure}{\textwidth}
  \centering
  \includegraphics[width=1.\linewidth]{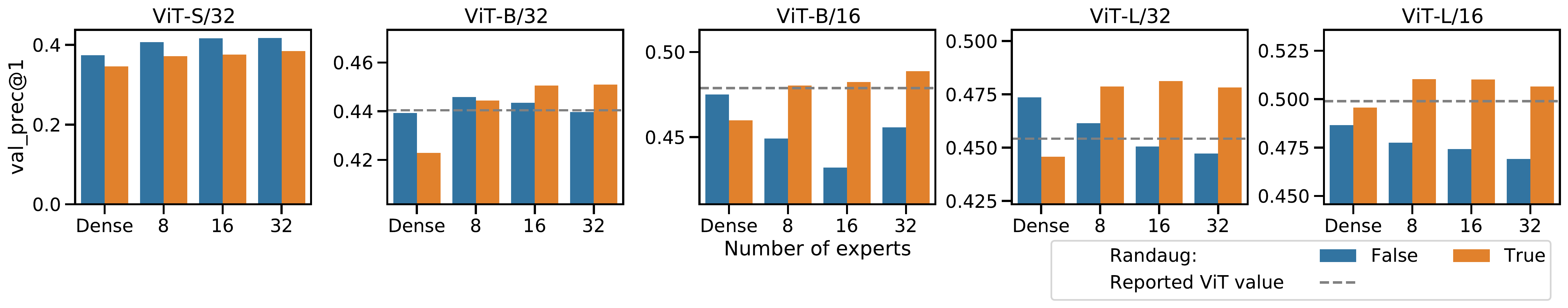}
  \caption{Precision@1 on the ImageNet-21k validation set}
  \label{im:imagenet21k_prec}
\end{subfigure}
\begin{subfigure}{\textwidth}
  \centering
  \includegraphics[width=1.\linewidth]{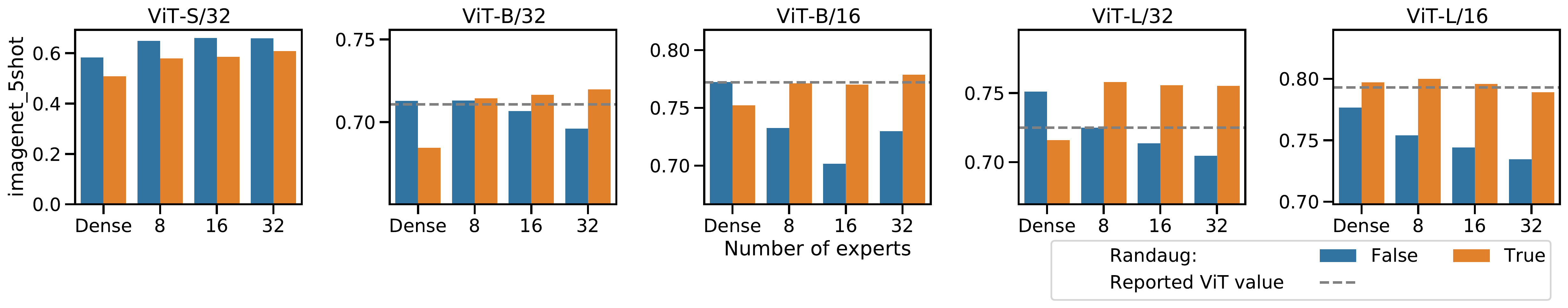}
  \caption{5-shot linear ImageNet performance}
  \label{im:imagenet21k_5shot}
\end{subfigure}
\caption{\textbf{Performance of ImageNet-21k pre-trained models.}}
\label{im:imagenet21k}
\end{figure}

\paragraph{Training on ImageNet21k.}
ImageNet21k~\cite{deng2009imagenet} is a large public dataset with approximately 14M images and 21k classes. Previous works~\cite{dosovitskiy2020image,kolesnikov2019big} have successfully pre-trained on it to achieve strong results in downstream tasks. In particular, dense ViT models trained on ImageNet21k perform reasonably well.
With the exception of ViT-S, where \abbv{} immediately outperforms the dense counterpart, applying sparse scaling generally harmed performance. We observed overfitting, both in the sense of reducing validation accuracy on the pre-training dataset, but also in reduced transfer performance as training continued. As an initial attempt at tackling this, we used RandAugment~\cite{randaug2020cubuk} with $N=2$ transformations of magnitude $M=10$. This is shown in \cref{im:imagenet21k}. Interestingly, RandAug typically helps expert models while harming dense models. With this applied, for each architecture, there is an expert model which outperforms the dense baseline.

This is far from a complete exploration; it indicates that these models can work with smaller data sources, and the key to their efficacy likely lies in more careful considerations of data augmentation and regularisation. We expect recent bodies of work exploring this for dense transformers~\cite{jiang2021token,touvron2021cait} to be useful here, and that works in data efficient vision transformers~\cite{touvron2020deit,yuan2021tokens} to also further unlock the potential of \abbv{} with less pre-training data.

\end{document}